# The Past and Future of AI Scientists

Ross D. King
[1]Department of Chemical Engineering and Biotechnology, University of Cambridge, Cambridge, UK.
[2]Department of Computer Science and Engineering, Chalmers University of Technology, Gothenburg, Sweden.

## Abstract

We present a survey of the past and future of AI Scientists: machines capable of automating science. AI Scientists can originate hypotheses, deduce their consequences, design and execute experiments, interpret their results, and revise their beliefs. Such systems are not enlarged prediction programs or conversational assistants. They are integrated scientific agents, connected to the literature, formal knowledge, mathematical models, simulations, data-analysis systems and physical laboratories.

Adam was the first machine to make novel scientific discoveries through cycles of hypothesis formation and physical experimentation. Eve established the architecture of the modern self-driving laboratory. Foundation models, autonomous agents and laboratory robotics now make it possible to build systems far more general than either Adam or Eve.

The central problem is no longer whether individual components of science can be automated. They can. The problem is integration. AI Scientists must combine neural learning with logic, probability, mathematics, causal reasoning, simulation, experimental design, robotics and formal scientific records.

AI Scientists have the potential to transform science: to make science faster, cheaper, more systematic and more reproducible. AI Scientists could investigate systems too complicated for unaided human science, and enable thousands of AI scientists to work together on single problems. They could also concentrate scientific power, automate error and connect scientific reasoning directly to dangerous physical actions. Their development must therefore include rigorous evaluation, complete provenance, accountability, safety and governance.

The Nobel Turing Challenge sets the goal of developing by 2050 AI systems capable of automating Nobel-quality discoveries. Progress is ahead of schedule. When we succeed it will create a new form of science and transform the world.

## 1. Introduction

The scientific method originated only once in history, in seventeenth-century Europe. It transformed the world. Thanks to technologies based on science, billions of people now live better lives than kings did in the past: they have access to better food, medical care, sanitation, communication, education and opportunities for travel. However, to meet the formidable challenges facing the twenty-first-century world—including climate change, food insecurity, antimicrobial resistance, cancer and future pandemics—science and technology need to become substantially more productive. The only feasible way to achieve the required increase in scientific productivity is through the application of Artificial Intelligence (AI).

The idea of automating reasoning is ancient (Boden, 2006). Its ultimate form is AI. The birth of modern AI is generally associated with Alan Turing's seminal paper 'Computing Machinery and Intelligence' (Turing, 1950) and the Dartmouth Summer Research Project on Artificial Intelligence in 1956 (McCarthy et al., 1955; Boden, 2006). AI may be defined as the study and construction of intelligent agents that perceive their environments and take actions designed to maximise the probability of achieving their goals (Russell & Norvig, 2021). Applying this definition to science, an AI Scientist is an intelligent agent that observes the real world through experiments and forms models that predict and explain what is observed.

Science is particularly well suited to the application of AI (Kitano, 2021). Science is abstract, like the games of chess and Go, in which AI systems can now defeat the best human players. Scientific problems are also restricted in scope. This matters because AI systems traditionally struggled to understand the wider context of a problem and to determine what knowledge is relevant: the 'Cabbages and Kings' problem (Boden, 1977). An AI system working on a particular scientific problem needs principally to understand the direct scientific context; it does not need comprehensive knowledge of vegetables, politics and every other aspect of the world.

Another significant advantage of AI for science is that nature is honest. If a human or an AI Scientist performs a scientific experiment, the physical world is not strategically attempting to deceive the scientist. This differs fundamentally from many applications of AI in business, politics, cybersecurity and warfare, where intelligent agents may actively seek to mislead one another. Nature may be difficult to understand, experimental observations may be noisy, and scientific models may be incomplete, but the underlying world is not an adversarial agent.

AI systems already possess superhuman scientific powers in specific components of science. They can search and process quantities of information beyond human capacity, execute formally verified deductions when connected to proof systems, maintain probability distributions over large hypothesis spaces, analyse enormous datasets, run simulations repeatedly and operate without fatigue. Their capabilities are complementary to those of human scientists, who remain superior in physical grounding, flexibility, tacit experimental understanding, social intelligence, moral responsibility and the identification of questions that are important rather than merely tractable.

Recent successes have established AI as a general-purpose technology for science. Deep learning systems can predict protein structures and molecular interactions, identify patterns in astronomical and physical data, generate candidate molecules and materials, emulate computationally expensive simulations, analyse biomedical images and extract structured information from the scientific literature (LeCun et al., 2015; Jumper et al., 2021; Merchant et al., 2023; Lam et al., 2023; Abramson et al., 2024; Dagdelen et al., 2024). Large Language Models (LLMs) can discuss scientific concepts, write and debug software, invoke external tools, search documents, analyse data and generate candidate hypotheses. Agentic systems can combine such capabilities into multi-step workflows that plan actions, use tools, evaluate intermediate results and revise their plans (Boiko et al., 2023; Bran et al., 2024; Lu et al., 2024; Gottweis et al., 2026; Ghareeb et al., 2026).

These advances make it necessary to distinguish several related classes of scientific AI system.

An AI scientific assistant supports one or more scientific activities under close human direction. Examples include systems for literature search, data analysis, code generation, prediction, image interpretation and manuscript preparation.

An agentic scientific system plans and executes multi-step computational research workflows. Such a system may generate ideas, search the literature, write and run code, analyse results and prepare scientific reports, but it need not interact physically with the world.

A self-driving laboratory autonomously selects, executes and analyses physical experiments within a defined experimental domain. Its principal emphasis is the closed-loop optimisation or exploration of experimental space.

An AI Scientist integrates scientific knowledge, hypothesis generation, experimental design, the physical or computational execution of experiments, interpretation of results and iterative revision of scientific hypotheses or theories, with a specified degree of autonomy.

These categories overlap, but they are not synonymous. A predictive model is not an AI Scientist merely because its predictions are scientifically valuable. An LLM that writes a research paper is not necessarily an AI Scientist. Similarly, laboratory automation becomes a self-driving laboratory only when experimental selection and interpretation are integrated into a closed feedback loop.

The central thesis of this article is that the components required to construct increasingly capable AI Scientists are now converging. These components include formal scientific knowledge, foundation models, automated reasoning, probabilistic belief states, hypothesis generation, simulation, experimental-design algorithms, agentic software, laboratory robotics, machine-readable protocols, data analysis and provenance. Their integration will enable the progressive automation of the scientific method itself.

## 2. The History of Discovery Science

The application of AI to science has traditionally been termed 'discovery science' or 'machine discovery'. Its founding masterwork was the DENDRAL/Meta-DENDRAL project, developed at Stanford University during the 1960s and 1970s (Buchanan et al., 1969; Lindsay et al., 1980). The project was conceived and led by Joshua Lederberg, a Nobel laureate in Physiology or Medicine, and involved a remarkable interdisciplinary team that included Edward Feigenbaum, later a recipient of the ACM A. M. Turing Award; Carl Djerassi, one of the inventors of the oral contraceptive pill; and Bruce Buchanan.

The application domain of DENDRAL was analytical chemistry, specifically the inference of molecular structures from mass-spectrometric and related chemical data. DENDRAL was the world's first expert system. It demonstrated that a computer could achieve expert-level performance in a specialised scientific domain by combining formalised domain knowledge with heuristic search. Its success established the central expert-system insight that intelligence depends not only on general inference mechanisms, but also on extensive, explicit and carefully structured knowledge about a particular field.

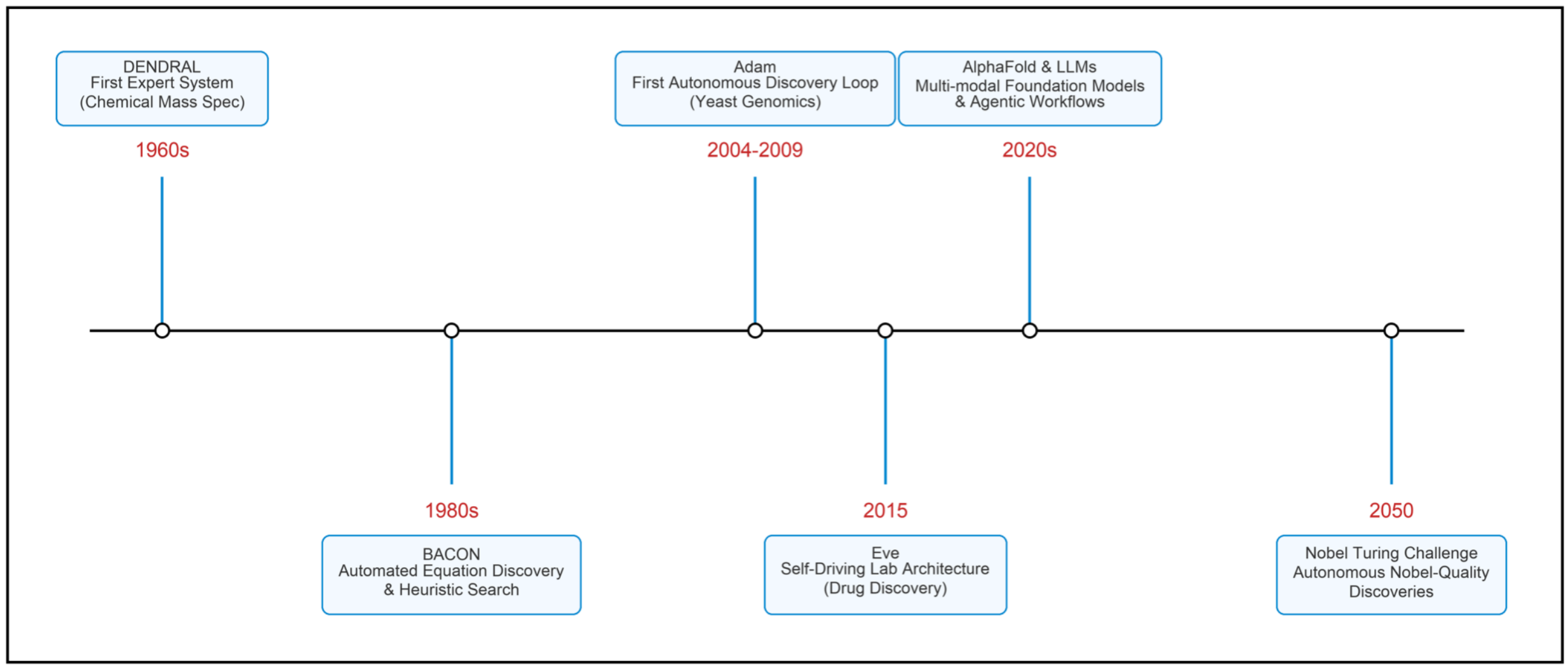

Figure 1 Discovery Science Timeline.

Meta-DENDRAL extended this work by learning rules for interpreting mass spectra from examples. It was therefore also one of the first important machine-learning systems and one of the earliest demonstrations that a machine could infer scientifically meaningful rules from empirical data. The DENDRAL programme established several themes that remain central to AI for science: the formal representation of scientific knowledge, the use of search to explore large spaces of hypotheses, the integration of data with background knowledge, and the production of interpretable scientific conclusions.

Another key early achievement in discovery science was the BACON project developed by Patrick Langley, Gary Bradshaw, Jan Żytkow and Herbert Simon (Langley et al., 1987). BACON used machine learning to generate equations that fitted numerical data and could rediscover forms of well-known scientific laws. This task would now be termed equation discovery or symbolic regression. The BACON systems demonstrated that apparently creative aspects of scientific reasoning could be represented as computational search guided by heuristics such as identifying constant ratios, products and differences among measured quantities.

The importance of BACON was not that it rediscovered scientific laws, but that it provided a computational account of part of the scientific discovery process. It showed that the formation of mathematical laws could, at least in restricted domains, be decomposed into explicit operations that a machine could perform. Contemporary systems for symbolic regression, equation discovery and automated model identification continue this research tradition, although they now make use of substantially greater computing power, improved search methods, deep learning and richer representations of physical knowledge.

Other important early discovery systems included AM and EURISKO, developed by Douglas Lenat (Lenat, 1977, 1983), and systems such as CYRANO for conceptual scientific discovery. These projects explored the generation of concepts, heuristics and conjectures rather than only the fitting of predictive models. They addressed a central challenge that remains unresolved: how a machine can transform its own representational language and invent concepts that were not explicitly specified by its designers.

A related historical tradition arose in automated mathematics. The Logic Theorist demonstrated that computers could prove mathematical theorems (Newell et al., 1957), while subsequent automated theorem provers and proof assistants increasingly mechanised deductive reasoning. Mathematics differs from empirical science because mathematical claims are established by proof rather than by physical experiment. Yet automated mathematics provided essential methods for formal representation, search, deduction and verification that are directly relevant to AI Scientists.

Literature-based discovery constituted another important route to machine-assisted scientific hypothesis generation. Don Swanson showed that novel hypotheses could be generated by connecting previously separate bodies of scientific literature. His classic work linked evidence concerning fish oil, blood viscosity and Raynaud's syndrome, despite the absence of publications directly connecting the two endpoint concepts (Swanson, 1986). Literature-based discovery established that new scientific hypotheses may already be implicit in the distributed structure of the published literature, but remain invisible to individual scientists because no human can read and integrate all relevant papers.

The transition from the rigid, top-down symbolic logic of early expert systems to the multi-layered neural architectures of modern deep learning was mediated by a crucial mid-history shift towards statistical bioinformatics and cheminformatics during the 1990s and 2000s. In chemistry, moving beyond the manually engineered heuristics of DENDRAL, the maturation of Quantitative Structure-Activity Relationship (QSAR) models introduced rigorous statistical regression and classical machine learning, including support-vector machines and random forests, to map high-dimensional molecular descriptors to chemical and biological activity (Kubinyi, 1993; Cherkasov et al., 2014).

Machine learning also became established in computational biology and drug design. Early work used inductive learning to predict protein secondary structure from amino-acid sequence and later developed comprehensible machine-induced rules for the same problem (King, 1987; King & Sternberg, 1990). Inductive logic programming was then used to learn relational structure–activity models for trimethoprim analogues binding to dihydrofolate reductase, illustrating how explicit chemical background knowledge could constrain drug-design hypotheses (King et al., 1992).

Concurrently, the deluge of genomic data generated by the high-throughput sequencing revolution necessitated probabilistic frameworks capable of parsing stochastic biological sequences. Hidden Markov Models (HMMs) emerged as the definitive mathematical architecture for this era, driving breakthroughs in structural gene finding, the classification of protein families via profile-HMMs (Krogh et al., 1994; Eddy, 1998), and the statistical validation of sequence alignment algorithms like BLAST (Altschul et al., 1990). Rather than relying on explicit logical predicates to define biological systems, these fields transitioned to leveraging continuous probability distributions over large empirical datasets to model evolutionary space (Durbin et al., 1998). By shifting the computational paradigm from hard-coded deductive rules to data-driven empirical induction, this statistical revolution established the repositories, mathematical frameworks, and multiple sequence alignment workflows that directly enabled contemporary structural biology breakthroughs like AlphaFold.

The first use of an AI system to directly control laboratory equipment was made by Jan Żytkow and colleagues (Żytkow et al., 1990). This was an important conceptual transition. Earlier discovery systems operated principally on existing data. Once an AI system could control scientific instruments, it became possible for the machine not only to analyse observations, but also to intervene in the physical world and acquire new observations selected to discriminate among competing hypotheses.

During the 2010s and 2020s, autonomous experimentation expanded rapidly across chemistry, materials science, biology and engineering. Advances in robotics, cloud computing, active learning, standardised laboratory interfaces and machine-readable protocols made it increasingly practical to connect AI systems to physical experiments. LLMs and foundation models subsequently added natural-language interfaces, software generation, tool use and agentic planning. These developments are reviewed in Section 3. Their historical significance lies in the convergence of two traditions: hypothesis-led machine discovery, established by Adam, and data-driven closed-loop experimentation, established by Eve. Genuine AI Scientists must ultimately connect scientific knowledge and hypotheses to interventions in the world, observations generated by those interventions and rational changes in belief.

## 2.1. Adam & Eve

The Robot Scientist project proposed the core idea of automating the scientific method. The first Robot Scientist, Adam, was designed to automate functional-genomics research in the yeast Saccharomyces cerevisiae (King et al., 2004; King et al., 2009). Adam represented background biological knowledge and competing hypotheses in logic, selected experiments to discriminate among those hypotheses, physically executed the experiments using laboratory robotics, interpreted the resulting data and revised the probabilities assigned to the hypotheses.

Adam was the first machine to autonomously discover novel scientific knowledge through cycles of hypothesis formation, experiment selection, physical experimentation and interpretation. Adam therefore led directly to the modern conception of AI Scientists: machines capable of automating substantial parts of the scientific method, rather than merely performing isolated prediction or optimisation tasks.

The second Robot Scientist, Eve, applied the same general philosophy to early-stage drug discovery (Williams et al., 2015). Eve integrated compound screening, assay automation, machine learning, active learning and iterative experimental selection. Whereas Adam was organised primarily around explicit competing biological hypotheses, Eve placed greater emphasis on closed-loop experimental optimisation and the efficient exploration of chemical and biological spaces.

Eve therefore led directly to the modern field of self-driving laboratories. A self-driving laboratory combines automated experimentation with machine learning or optimisation so that the results of one set of experiments determine which experiments are conducted next. The system repeatedly executes a cycle of prediction, experiment selection, physical execution, measurement, learning and renewed selection. The terminology became widespread later, particularly in chemistry and materials science, but the essential architecture was established by Eve.

This historical distinction remains important. Adam is the primary conceptual ancestor of AI Scientists because it automated hypothesis-led scientific reasoning. Eve is the primary conceptual ancestor of self-driving laboratories because it demonstrated autonomous, data-driven, closed-loop experimental search. Contemporary systems increasingly combine these two traditions: explicit or language-based scientific reasoning derived from the AI Scientist lineage, and efficient closed-loop experimentation derived from the self-driving-laboratory lineage.

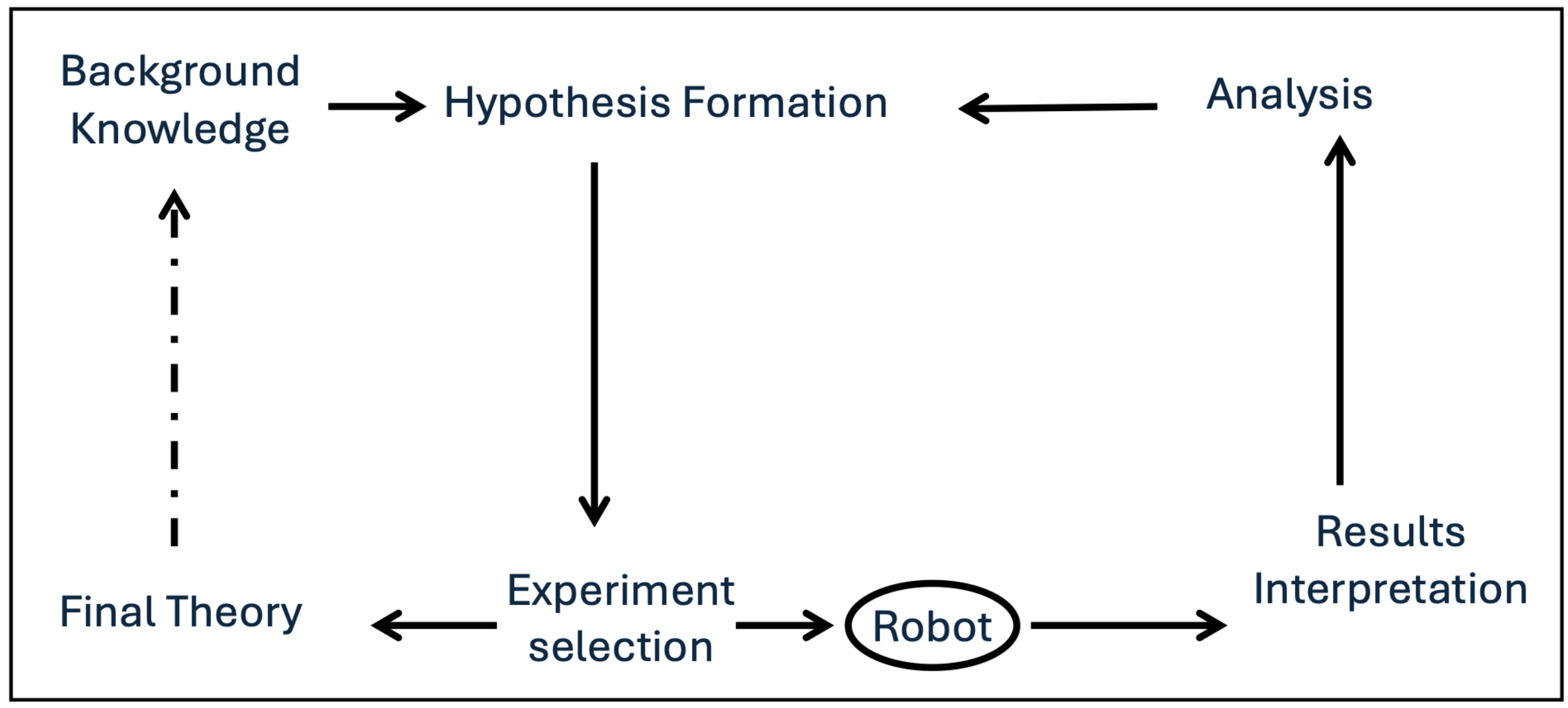

Figure 2 The Original Automation of Science diagram from (King et al., 2004).

## 3. The Current State of AI for Science

AI is already embedded in modern science. It predicts structures, proposes molecules, writes code, reads papers and selects experiments. Yet most systems automate one task. The frontier is not another predictor. It is the integration of these powers into agents that conduct science.

The commercial success of machine learning was made possible by decades of investment in infrastructure for capturing, communicating and storing data at large scale. A parallel process has occurred in science (Hey et al., 2009). Modern instruments, high-throughput experiments, electronic health records, imaging facilities, genome-sequencing programmes, astronomical observatories and physical simulations produce quantities of data that cannot feasibly be analysed by humans alone. Data, metadata, ontologies, software and executable analysis methods now form an essential foundation for scientific AI.

The present landscape can be organised into four overlapping classes: predictive scientific AI, LLM-based and agentic scientific systems, self-driving laboratories and integrated AI Scientists.

### 3.1. Predictive scientific AI

The most visible successes of AI for science involve prediction. Deep neural networks can learn complex functions from large quantities of empirical data and often make accurate predictions without using the explicit symbolic theories traditionally favoured in science.

The most prominent example is AlphaFold 2, which achieved a qualitative advance in predicting protein structure from amino-acid sequence (Jumper et al., 2021). The associated AlphaFold Protein Structure Database made predicted structures available for more than 200 million proteins. AlphaFold 3 extended the approach to the joint structures and interactions of proteins, nucleic acids, small molecules, ions and modified residues (Abramson et al., 2024).

Comparable methods are used for molecular generation, protein design, weather forecasting, materials-property prediction, medical imaging, quantum chemistry and surrogate modelling. Scientific foundation models seek to learn general representations from large and heterogeneous scientific datasets and adapt those representations to multiple downstream tasks.

Such systems possess impressive scientific capabilities, but prediction alone is not scientific autonomy. A predictive model is generally given a human-defined problem, a fixed representation of inputs and outputs, training data selected by humans and an externally specified measure of success. It does not necessarily identify which scientific question should be asked, formulate explanatory hypotheses, design decisive experiments or revise its conceptual representation of a domain.

### 3.2. Large Language Models and agentic scientific systems

LLMs have expanded the range of scientific tasks that can be automated. They can summarise literature, explain concepts, generate and debug code, translate between natural and formal languages, propose experiments, interpret tables and figures and communicate results. Their general linguistic competence allows them to act as interfaces among human scientists, scientific software, databases and laboratory equipment.

The important development is not language generation in isolation, but integration into agentic systems. An agentic scientific system can decompose a goal into subtasks, search for information, select and invoke tools, write and execute software, inspect intermediate outputs, identify errors and alter its plan. Architectures commonly divide work among specialised components for planning, literature search, coding, analysis, criticism and communication.

Coscientist combined LLMs with internet and documentation search, code execution and experimental automation to design and perform chemistry tasks (Boiko et al., 2023). ChemCrow connected an LLM to specialist chemical tools (Bran et al., 2024). The AI Scientist automated a computational workflow from idea generation through code execution, analysis, manuscript production and simulated peer review (Lu et al., 2024). Co-Scientist used multiple agents to generate, debate, rank and refine biomedical hypotheses (Gottweis et al., 2026). Robin integrated literature-search and data-analysis agents to generate hypotheses, propose experiments, interpret biological data and update its hypotheses (Ghareeb et al., 2026).

These are substantial advances. Yet most remain dependent on human-defined objectives, curated data, established tools and restricted environments. They can generate false statements, invented references, invalid reasoning and plausible but experimentally unsupported hypotheses. Computational completion of a research workflow is not by itself the complete automation of empirical science.

While such systems demonstrate impressive engineering integration by automating the end-to-end computational research workflow—from execution to manuscript generation—they also highlight a critical boundary between process automation and genuine conceptual discovery. By utilising large language models to iterate on existing machine learning pipelines, the framework successfully streamlines the execution and reporting of research tasks. However, the insights generated remain largely incremental and tightly bounded by the initial code environment. This underscores a vital distinction: automating the delivery of structured academic artifacts is not synonymous with true scientific insight, which ultimately demands ontological innovation, the identification of fundamental anomalies, and paradigm shifts that look beyond the existing literature.

### 3.3. Self-driving laboratories

Self-driving laboratories integrate automated experimental equipment with algorithms that use accumulated results to select subsequent experiments. Their typical objective is to search an experimental space efficiently, optimise one or more objectives or learn a predictive model relating experimental conditions to outcomes.

Their basic cycle is:

- Represent the current state of knowledge;
- Select the next experiment or experiments;
- Translate the selection into an executable protocol;
- Execute the protocol using laboratory automation;
- Collect and analyse the resulting data;
- Update the model or hypothesis distribution;
- Repeat the cycle.

Eve established this architecture for drug discovery. Subsequent self-driving laboratories have applied related principles to chemical synthesis, catalyst optimisation, materials discovery, formulation, protein engineering and process development. Bayesian optimisation, active learning, evolutionary search and multi-objective optimisation are commonly used to choose experiments.

A-Lab combined computational predictions, knowledge extracted from the literature, active learning, robotics and automated characterisation for autonomous inorganic-materials synthesis (Szymanski et al., 2023). SAMPLE performed repeated closed-loop cycles of protein design, construction and testing (Rapp et al., 2024). Fast-Cat mapped multi-objective Pareto fronts for catalysts (Bennett et al., 2024), while automated flow platforms have jointly optimised reaction yield, process intensification and scale-up (Slattery et al., 2024).

Modern self-driving laboratories increasingly use LLM agents as interfaces to laboratory software, documentation and protocols. This reduces the amount of task-specific software required, but raises the standard of reliability: a false textual statement is undesirable, whereas an incorrect command to a physical laboratory can damage equipment, waste scarce materials or create a safety hazard.

### 3.4. Integrated AI Scientists

AI Scientists are more general than self-driving laboratories. A self-driving laboratory may efficiently optimise an objective without generating or evaluating explicit scientific hypotheses. An AI Scientist should use background knowledge to explain observations, originate hypotheses, derive experimentally testable consequences, select experiments that discriminate among alternatives, execute those experiments, interpret their results and revise its scientific beliefs.

Adam was the first system to instantiate this conception in a physically embodied form. Current advances in LLMs, knowledge graphs, neuro-symbolic AI, causal inference, probabilistic programming and automated experimentation create the prospect of systems with much broader scientific competence than Adam.

No existing system yet matches the breadth, depth and open-ended autonomy of a first-rate human scientist. Present systems usually operate within domains chosen by humans. Their instruments, materials, hypothesis languages, objective functions and permissible actions are largely predetermined. They rarely invent new experimental technologies, reconstruct their conceptual foundations or decide independently which scientific problems are important.

The major components of autonomous science have nevertheless each demonstrated substantial capability. Foundation models acquire broad scientific representations; LLMs use tools and coordinate workflows; formal languages and knowledge graphs represent explicit information; reasoners derive consequences and check consistency; learning systems generate models and hypotheses; active learning selects informative experiments; robots execute protocols; and agents analyse results and revise plans. The challenge is increasingly one of integration.

| System | Paradigm / domain | Main contribution | Automation / revision | Primary limitation |
|---|---|---|---|---|
| Adam<br>(King et al., 2009) | Integrated AI Scientist (Pioneer)<br>Functional genomics | First fully autonomous, hypothesis-led discovery loop | Full (Direct robotic execution)<br>Yes (Logic-based revision) | Restricted to a narrow, highly structured biological domain. |
| Eve<br>(Williams et al., 2015) | Self-Driving Lab (Pioneer)<br>Early-stage drug discovery | Pioneer of active-learning-driven experimental optimization | Full (Direct robotic execution)<br>Yes (Data-driven optimization) | Limited to pre-configured screening assays and fixed target objectives. |
| AlphaFold 2 & 3<br>(Jumper et al., 2021; Abramson et al., 2024) | Predictive Scientific AI<br>Structural biology | Atomic-accuracy prediction of biomolecular structures & interactions | None (Purely computational)<br>No (Static inference) | Requires human-defined problems, fixed input tokens, and static training datasets. |
| The AI Scientist<br>(Lu et al., 2024) | LLM-Based / Agentic System<br>Machine-learning research | End-to-end automated computational workflow & paper writing | None (Purely computational)<br>Partial (In-silico idea generation & tuning) | Confined entirely to software-based, simulation-only domains. |
| Co-Scientist / Robin<br>(Gottweis et al., 2026; Ghareeb et al., 2026) | LLM-Based / Agentic System<br>Biomedicine & Biology | Multi-agent literature synthesis, cross-paper hypothesis generation, & ranking | Indirect (Human-in-the-loop or semi-automated)<br>Partial (Iterative textual updating) | Lacks a fully automated physical execution loop; susceptible to LLM hallucination. |
| Coscientist / ChemCrow<br>(Boiko et al., 2023; Bran et al., 2024) | LLM-Based / Agentic System<br>Chemistry & Synthesis | LLM tool integration, agentic planning, and direct API hardware control | Full (API-driven lab hardware)<br>Bounded (Execution & validation focused) | High dependency on prompt reliability; high risk of hardware damage if plans fail. |
| A-Lab<br>(Szymanski et al., 2023) | Self-Driving Laboratory<br>Materials science | Autonomous inorganic material synthesis and active-learning characterization | Full (Direct robotic synthesis)<br>Yes (Model-guided optimization) | Bound strictly by the specific physical capabilities of the target materials domain. |
| SAMPLE<br>(Rapp et al., 2024) | Self-Driving Laboratory<br>Protein engineering | Closed-loop automated "Design-Make-Test" cycles for proteins | Full (Direct robotic synthesis)<br>Yes (Active-learning optimization) | Confined to exploring search spaces that are explicitly predefined by human designers. |

**Table 1.** Taxonomy and capabilities of representative scientific AI systems

The distinctions in Table 1 are functional rather than promotional. We should reserve the term AI Scientist for systems that integrate substantial parts of the scientific loop. A useful predictor is not automatically a scientist.

## 4. Architecture and Roadmap for AI Scientists

An AI Scientist must close the scientific loop. It must move from knowledge to questions, from questions to hypotheses, from hypotheses to experiments, and from observations back to revised knowledge. Any break in that loop returns the system to the status of a tool.

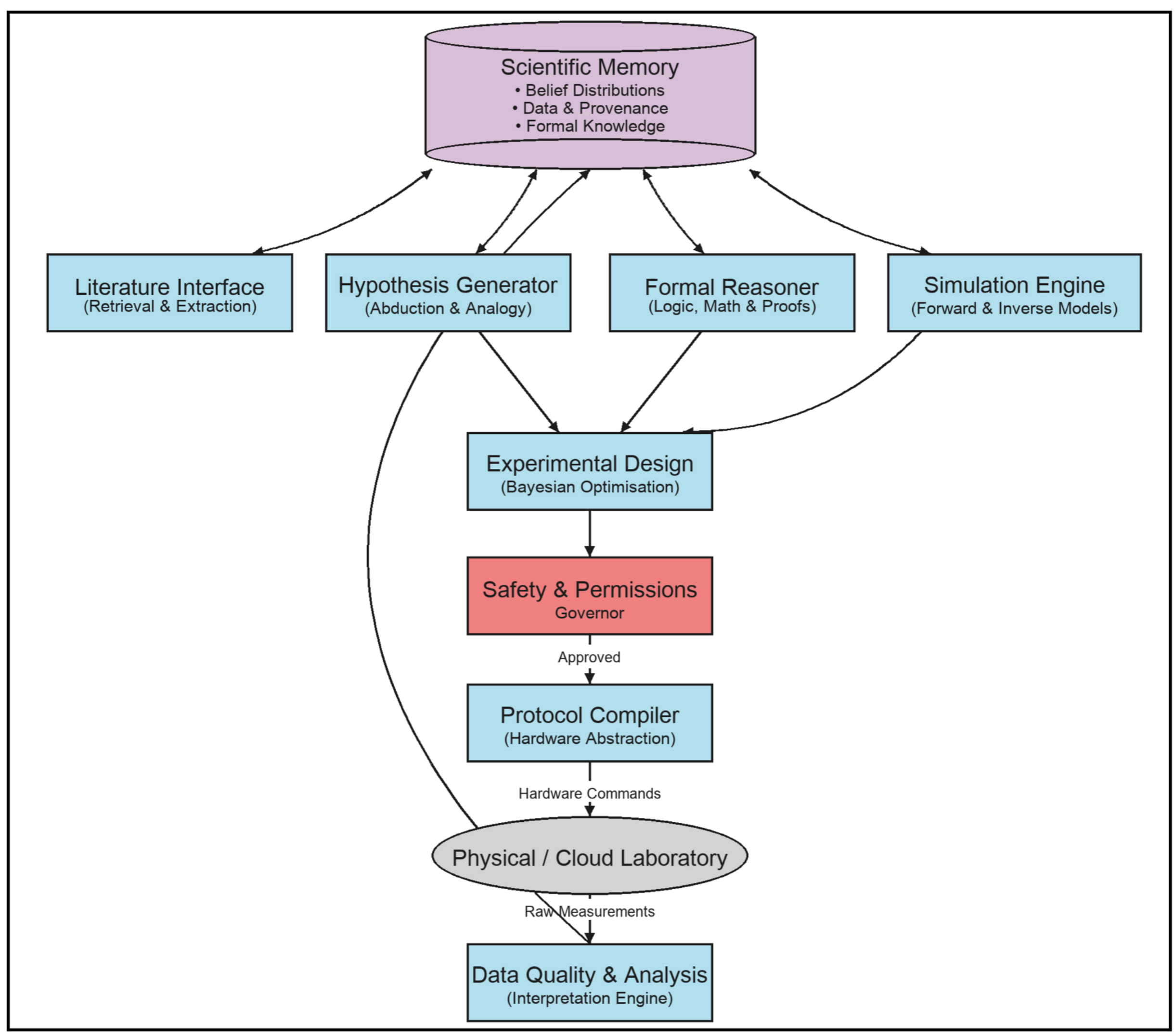

**Figure 3** Shows an AI Scientist as a federation of interacting modules rather than a single monolithic model.

The principal components are:

- A scientific memory containing formal knowledge, probabilities, models, data, protocols and provenance;
- A literature interface for retrieval, reading, extraction and citation checking;
- A question and hypothesis generator;
- Formal logical, mathematical and probabilistic reasoners;
- A simulation and model-execution environment;
- An experimental-design and portfolio-allocation system;
- A protocol compiler linked to inventories and equipment capabilities;
- Laboratory-control and observational interfaces;
- Data-quality, statistical-analysis and interpretation modules;

- Adversarial critics and independent verification agents;
- A safety and permissions governor;
- A provenance system recording every material decision and action;
- Interfaces through which human scientists set objectives, inspect reasoning and intervene.

Every arrow in this loop is a potential failure point. A hypothesis may be linguistically plausible but scientifically incoherent. A simulation may implement the wrong model. A valid experimental design may be compiled incorrectly for an instrument. A robot may issue the intended command but fail physically. A statistical analysis may be mathematically correct while answering the wrong question. A complete AI Scientist must represent and check these distinctions explicitly.

The scientific memory should not be a passive archive. It should contain a probabilistic belief state over competing hypotheses and models. Every new observation should be linked to its experimental and analytical provenance and should update the relevant beliefs. The system should preserve disagreements and negative results rather than overwrite them with a single current answer.

The architecture should be neuro-symbolic and probabilistic. Neural models provide perception, flexible language understanding, representation learning and generative search. Symbolic systems provide explicit semantics, constraints, deduction and proof. Probability theory represents uncertainty and rational belief revision. Mathematics and executable programs express quantitative models. Robotics and instruments connect these representations to nature.

No single representation is sufficient. A purely symbolic system cannot acquire all relevant scientific knowledge manually. A purely neural system cannot guarantee logical validity, calibrated uncertainty or faithful provenance. A purely statistical system may predict without explaining. A complete AI Scientist must select and combine representations according to the scientific task.

## 4.1. One autonomy framework

Progress should be described using one autonomy framework, not several overlapping stage systems. We adopt six levels, numbered 0 to 5. The levels describe the degree of human intervention in the scientific process; they do not by themselves measure intelligence or scientific merit.

| Level | Description | Human role |
|---|---|---|
| 0: No scientific autonomy | AI is used as an ordinary computational tool and performs only explicitly requested calculations, searches or transformations. | Humans formulate the question, hypotheses, experiments and conclusions. |
| 1: Task autonomy | The system performs one bounded scientific task, such as analysing data, retrieving literature or optimising a specified function. | Humans define the task, provide the inputs and interpret the result. |
| 2: Workflow autonomy | Several scientific tasks are integrated. The system may propose hypotheses or experiments but requires substantial human selection, information transfer or approval. | Humans control the research programme and intervene repeatedly. |
| 3: Closed-loop autonomy | The system selects and executes experiments, analyses results and updates a model within a predefined question and experimental space. | Humans specify the objective, available actions and operational boundaries. |
| 4: Research-programme autonomy | The system formulates subsidiary questions, generates hypotheses, designs and executes experiments, integrates results and communicates conclusions within a broad human-defined domain. | Humans specify high-level goals, resources, ethical limits and review points. |

| Level | Description | Human role |
|---|---|---|
| 5: Open-ended scientific autonomy | The system can identify important scientific problems, create new concepts and experimental methods, execute extended programmes and produce scientific knowledge without human intervention in the discovery process. | Humans provide external governance rather than scientific direction. |

**Table 2.** Levels of scientific autonomy

Autonomy must be reported separately from capability: how well the system performs; generality: the range of domains and representations in which it operates; scale: the number of hypotheses, agents, experiments and laboratories coordinated; reliability: the frequency and severity of error; scientific significance: the importance of the discoveries; and risk: the potential consequences of failure or misuse.

A Level 3 system may make an important discovery, while a Level 5 system may produce trivial or incorrect work. Conversely, a highly capable system may be operated under close supervision for reasons of safety or law.

### 4.2. Milestones

Near-term progress will come principally from making bounded scientific systems dependable. The first requirement is reliable extraction of claims, evidence and protocols from the literature, together with formal representations that connect natural language to models, data and experiments. Those representations must support calibrated uncertainty over competing hypotheses and prospective hypothesis generation whose quality is determined by experiment rather than rhetorical plausibility.

A second group of milestones concerns physical execution. Protocol generation must be formally checked, aware of inventory and equipment, and constrained by safety rules. Autonomous laboratories must diagnose and recover from experimental failures, while measurement and analysis pipelines preserve uncertainty and complete provenance. Results should be reproduced independently across laboratories and implementations, not merely repeated on the same platform.

Progress will also require benchmarks that measure novelty, correctness, autonomy and scientific significance, and interoperable networks of laboratories and computational resources. The transition to higher autonomy will demand more than longer agentic workflows: systems must be able to change their own problem representations, invent new experimental actions, recognise when an existing instrument is inadequate, and distinguish a failed theory from a failed experiment.

### 4.3. Evaluation as part of the architecture

Evaluation cannot be postponed until after deployment. Each module should expose measurable claims about its competence, uncertainty and operating limits, and the complete AI Scientist should recognise when a problem lies outside its validated domain and request human assistance or independent verification.

Every scientific claim should carry a machine-readable evidential record. That record should identify the hypothesis or model under assessment, the evidence used, the deductions and simulations performed, the experiments selected and the reasons for selecting them, the observations obtained, the analyses applied, and the resulting changes to the system's belief state. It should also distinguish human and machine contributions and state the uncertainty that remains.

This architecture provides the organising framework for the remainder of the article. Sections 5 and 6 address scientific representation and literature; Section 7 addresses hypothesis generation; Section 8 addresses deduction and simulation; Sections 9 and 10 address observation, experimentation, analysis and belief revision; Section 11 addresses evaluation; and Sections 12–14 address collaboration, governance and the future.

## 5. Representations for Scientific Knowledge and Models

A scientist can discover only what its language enables it to represent. The choice of representation determines which hypotheses can be stated, which deductions can be made and which experiments can be designed. AI Scientists therefore require languages that are expressive, executable and precise.

We distinguish formal languages and natural languages such as English. Most human science is conducted using a mixture of the two. Natural language is used to motivate problems, describe observations, communicate arguments and explain significance. Formal languages are used to express mathematical models, statistical assumptions, chemical structures, logical relations, computer programs and experimental data.

Formal languages are to be preferred in science because only they provide the semantic clarity required for the unrestricted exchange, computational integration and reproducibility of scientific knowledge. Natural-language statements can be ambiguous, incomplete, context-dependent or internally inconsistent. By contrast, a well-defined formal language specifies the syntax of valid statements and gives those statements an explicit semantics. This permits scientific claims to be checked for consistency, related deductively to other claims and, in suitable cases, verified automatically.

The best-understood formal languages for expressing scientific knowledge are formal logic, probability theory, mathematics and, most generally, computer programs or Turing machines. These languages are complementary. Logic is particularly suitable for expressing entities, relations, hypotheses and deductive consequences. Probabilities express uncertainty and degrees of belief. Mathematics expresses quantitative structures and scientific laws. Computer programs can represent arbitrarily complex executable models and procedures.

AI Scientists form hypotheses through abductive or inductive inference, supplemented by probabilistic reasoning, analogy, search and the invention of new concepts and predicates. The expressive power of the language used to represent hypotheses therefore determines the space of scientific ideas that an AI Scientist can explore.

### 5.1. Logic, Ontologies and Executable Knowledge

Formal logic was invented/discovered by Aristotle approximately 2,400 years ago and was used from the beginning to express scientific knowledge (Kneale & Kneale, 1962). Aristotle was a founder of both physics and biology, as well as of formal logic. The logical representation of knowledge subsequently became central to scholastic philosophy and to the development of modern analytical philosophy.

The use of logic to represent scientific knowledge has traditionally been favoured by philosophers of science. As Toulmin observed, most analytical philosophy of science was based on the assumption that the intellectual content of a natural science could be represented as a formal propositional system possessing a definite logical structure (Toulmin, 1953). Logic has the advantages of being comprehensible, supporting rigorous inference and providing transparent explanations of how conclusions follow from premises.

The study of logic gave birth to computer science. The modern computer emerged from attempts to formalise mathematics and determine which logical and mathematical operations could be executed mechanically (Hilbert & Ackermann, 1928; Gödel, 1931; Church, 1936; Turing, 1936). Logic remains fundamental to the computational representation of knowledge, database theory, programming-language semantics, software verification and automated reasoning.

#### 5.1.1. The formalisation of scientific knowledge

The formalisation of scientific knowledge is becoming increasingly technologically necessary. Every area of science faces a growing volume of publications, data, models, protocols and software. In many fields, this growth has become a deluge. Science therefore depends increasingly on computers to store, retrieve, integrate, analyse and communicate information.

The full power of computers—which themselves originated as a consequence of the formalisation of mathematics—can be exploited only when the knowledge upon which they operate is formalised. A computer can search natural-language text for statistically associated strings, but it cannot reliably reason over scientific knowledge unless the relevant entities, relations, assumptions and meanings have been made sufficiently explicit.

This requirement motivated the Semantic Web: the development of standards through which information on the web could be given machine-interpretable meaning rather than being represented only as documents intended for human readers (Berners-Lee et al., 2001). Technologies including the Resource Description Framework (RDF), the Web Ontology Language, rule languages, query languages and provenance models provide elements of an infrastructure for publishing, linking and reasoning over formalised knowledge.

A related objective underlies e-Science and the FAIR principles: papers, data, metadata, software, workflows and experimental methods should be findable, accessible, interoperable and reusable (Wilkinson et al., 2016). For AI Scientists, interoperability requires more than common file formats. It requires explicit agreement about what the data denote, how measurements were obtained, what assumptions were made and how conclusions are related to evidence.

#### 5.1.2. Ontologies

A key step in formalising scientific knowledge is to define an explicit ontology: an account of what exists in the domain and of the categories and relations through which it is understood. Logic is generally used to specify ontologies.

We follow the widely accepted definition of an ontology as an "explicit specification of a conceptualisation" (Gruber, 1993). More compositionally, an ontology consists of concepts, their hierarchical organisation, relations among them and axioms that formalise their definitions and permitted relations (Mizoguchi, 2004).

An ontology does not merely provide a controlled vocabulary. It defines the intended meanings of terms and constrains how they may be combined. For example, an ontology may distinguish an experimental process from its physical inputs and outputs, a hypothesis from the proposition it expresses, and an observation from the conclusion inferred from it. Such distinctions are essential if scientific records are to support reliable machine reasoning.

Ontologies are increasingly used to organise and exchange scientific knowledge. They have been developed to describe genes, proteins, diseases, chemicals, phenotypes, scientific hypotheses, experimental designs, laboratory actions, machine-learning studies and research provenance. The EXPO ontology provided a general formal representation of scientific experiments (Soldatova & King, 2006), while EXACT formalised the actions involved in experimental protocols (Soldatova et al., 2008). These ontologies underpinned the Robot Scientist systems by making the aims, structure, execution and interpretation of experiments explicit.

Later standards broadened this semantic infrastructure. The Ontology for Biomedical Investigations represents the planning, execution and reporting of biomedical investigations across disciplines (Bandrowski et al., 2016). W3C PROV-O provides a general ontology for entities, activities, agents and provenance relations (Lebo et al., 2013), while RO-Crate packages data, software, workflows and their metadata as connected, machine-readable research objects (Soiland-Reyes et al., 2022). LabOP applies related principles to executable biological protocols (Bartley et al., 2023). Together, these resources show a movement from isolated domain vocabularies towards interoperable representations of complete scientific processes.

Ontologies will be essential to future AI Scientists because scientific terms cannot be interpreted reliably in isolation. Their meanings depend on relations to other concepts, constraints on their use and the context in which a claim was produced. Ontologies provide the semantic scaffolding required to relate natural-language expressions, database identifiers, formal hypotheses, experimental actions and observations.

#### 5.1.3. Expert systems and the knowledge-acquisition bottleneck

Early AI systems were based principally on logical inference (Boden, 2006). This changed with the advent of expert systems, which combined general inference mechanisms with formalised declarative and operational knowledge about specialised domains.

Expert systems dominated AI during the 1980s. Their great strengths were that their knowledge was explicit and formalised, their reasoning could be inspected, and their conclusions could be queried and checked. Provided that the encoded facts and inference rules were correct and the inference engine was sound, expert systems did not hallucinate. Their conclusions followed from an identifiable body of knowledge through explicit inference steps. (Buchanan & Shortliffe, 1984).

The expert-system programme was limited by the knowledge-acquisition bottleneck. Knowledge engineering—the extraction and formal representation of knowledge possessed by human experts—proved difficult, expensive and slow. Human experts frequently could not articulate the tacit knowledge underlying their judgements. Even when the knowledge was articulated, converting it manually into formal rules and ontologies required extensive collaboration between domain experts and knowledge engineers. (Feigenbaum, 1977).

Most explicit human knowledge is contained in written natural-language texts. Until recently, extracting the knowledge expressed in such texts and translating it into formal representations was computationally intractable at scale. This prevented the construction and maintenance of sufficiently comprehensive expert systems.

#### 5.1.4. Knowledge graphs

The most commonly used contemporary structures for representing large bodies of formalised knowledge are knowledge graphs. A knowledge graph represents entities as nodes and relations as edges, together with attributes, types, identifiers, provenance and schema-level constraints. It can integrate heterogeneous datasets by connecting records that refer to the same entities and by organising them according to a shared ontology.

The meaning of the information in a knowledge graph is defined by its ontology or schema. This specifies the types of entities that may occur, the relations that may connect them and constraints on those relations. A reasoner may then derive additional knowledge, identify inconsistencies or classify entities under more general concepts.

Knowledge graphs are widely used in search engines, question-answering systems, recommender systems, biomedicine and autonomous systems. Large commercial graphs contain billions of entities connected by still larger numbers of relations (Noy et al., 2019). Scientific knowledge graphs include resources describing genes, proteins, diseases, chemicals, drugs, materials, publications and experimental evidence.

The graph structure provides practical efficiency, scalability and ease of integration, but it also limits the semantic detail and forms of inference that can be expressed directly. Most large knowledge graphs represent knowledge as binary relations or subject–predicate–object triples. Scientific claims, however, commonly involve quantification, negation, modality, uncertainty, temporal conditions, experimental context and relations among relations. Richer logical representations are therefore required when the semantics of a scientific hypothesis cannot be reduced adequately to a collection of triples.

Knowledge graphs and LLMs have complementary strengths. Knowledge graphs contain explicit, inspectable and addressable statements, but are incomplete and expensive to construct. LLMs contain broad implicit knowledge and can interpret diverse linguistic expressions, but their knowledge is distributed through numerical parameters and cannot be inspected or trusted directly. Future AI Scientists will combine the flexibility of LLMs with the explicit semantics and verifiability of ontologies and knowledge graphs.

#### 5.1.5. Logic programming and programs as knowledge

The most general possible formal representation of knowledge is a computer program or Turing machine. A program can represent an executable scientific model, a simulation, an experimental protocol, a data-analysis procedure or a theory that generates predictions.

Logic programming is a programming paradigm in which logic serves both as a declarative representation of knowledge and as an executable language. Kowalski's formulation separated the logical meaning of a set of clauses from the control strategy used to execute them, establishing the principle that an algorithm can be understood as logic plus control (Kowalski, 1974). The semantics and operational foundations of Prolog and related languages were subsequently developed systematically (Lloyd, 1987).

This dual role is attractive for AI Scientists. The same representation can state background knowledge, express hypotheses, derive predictions and explain conclusions. Variables and relations permit general statements about structured objects; constraints can encode units, types and admissible experimental states; and proof traces can connect a conclusion to explicit premises. Because the declarative content is separable from execution, alternative search or inference engines can operate over the same scientific knowledge.

Inductive logic programming extends the paradigm by learning logic programs from observations and background knowledge (Muggleton, 1991). It therefore unifies knowledge representation, machine learning and executable models, and supports the learning of recursive relations, interpretable rules and invented predicates. Modern ILP systems use meta-level search, constraints and neural guidance to reduce the immense hypothesis space, but scalability, noise and predicate invention remain central research problems (Cropper et al., 2022).

#### 5.1.6. LLMs and the return of expert systems

LLMs transform the prospects for formal scientific knowledge. They perform remarkably well at extracting entities, relations, events, measurements and other structured information from natural-language scientific text. They can map varied linguistic expressions onto common schemas, propose ontology classes and relations, translate text into logical or programmatic representations and generate candidate definitions for previously unnamed concepts.

Experiments in materials science have shown that language models can extract complex structured records from scientific papers, including relations among materials, dopants, phases, morphologies, properties and applications (Dagdelen et al., 2024). LLM-based systems can also construct knowledge graphs without requiring the complete schema to be placed in the prompt and can induce and canonicalise parts of the schema automatically (Zhang & Soh, 2024). But scientific extraction remains imperfect, especially where papers contain complex tables, figures, implicit conditions or domain-specific conventions. Formalisation must therefore be followed by checking and verification.

The resulting architecture reverses the principal weakness of classical expert systems. The knowledge-acquisition bottleneck can now be overcome by using LLMs to translate the natural-language scientific literature into formal representations. The proposed representations can then be checked using ontologies, type systems, logical reasoners, databases, proof assistants and links to primary evidence.

The appropriate pipeline is:

- Retrieve the relevant scientific documents and data;
- Use an LLM to identify claims, entities, relations, quantities, conditions and provenance;
- Translate these into a formal representation;
- Align the representation with established ontologies and identifiers;
- Check its syntax, type constraints and logical consistency;
- Connect each formal claim to the evidence from which it was extracted;
- Test deductions using a sound formal reasoner;
- Identify uncertainty, contradictions and missing information;
- Submit unresolved claims to human review or experimental testing; and
- Add verified claims to a persistent formal scientific knowledge base.

Once knowledge has been formalised, it can be checked for correctness and truth. Logical consequences can be verified mechanically. Contradictions can be identified. Claims can be traced to their evidence. Experimental observations can be used to increase or decrease the probability assigned to formal hypotheses. Thus, certified scientific knowledge can now be generated at scale. LLMs provide the flexible interface required to extract and create formal representations; symbolic systems provide the semantic discipline, verification and transparency that LLMs alone lack.

The combination provides the basis for a new generation of expert systems with trusted intelligent output. Unlike classical expert systems, their knowledge bases can be constructed and updated automatically from the scientific literature. Unlike ungrounded LLMs, their conclusions can be constrained by explicit knowledge, checked by sound inference engines and traced to evidence.

The formalisation of the scientific literature will create the foundational knowledge required by future AI Scientists. It will allow them not merely to retrieve text that resembles a query, but to understand what claims have been made, what evidence supports them, how they relate logically to other claims and which experiments would most efficiently resolve the remaining uncertainty.

## 5.2. Probabilities

In an 1850 letter, James Clerk Maxwell famously and correctly stated that “the true logic for this world is the calculus of probabilities” (Maxwell, 1850; Harman, 1990). Scientific knowledge is provisional and probabilistic. There is therefore no sharp epistemic division between established scientific knowledge and new hypotheses. The distinction is one of degree: established knowledge consists of hypotheses and theories to which the scientific community assigns high probabilities because they are supported by extensive and varied evidence.

A core idea in automating science is that hypotheses may have probabilities associated with their correctness. An AI Scientist should not normally represent a hypothesis simply as true or false. It should represent its degree of belief in the hypothesis, the uncertainty associated with its parameters, the reliability of the evidence and the probabilities of relevant alternative explanations.

### 5.2.1. Bayesian updating

Bayes’ theorem provides the fundamental rule for revising the probability assigned to a hypothesis in the light of new evidence.

When a prediction made by a hypothesis is experimentally confirmed, the probability of that hypothesis increases when the observation was more probable under the hypothesis than under its alternatives. Conversely, when a prediction is not confirmed, the probability of the hypothesis decreases when the failure was less probable under the hypothesis than under competing hypotheses.

This has direct implications for AI Scientists. An AI Scientist should select experiments whose possible outcomes have very different probabilities under competing hypotheses. Experiments that are predicted equally well by every hypothesis provide little information, even if their results agree with all of them.

### 5.2.2. Popper and probabilistic confirmation

Some philosophers of science have argued against assigning probabilities to hypotheses, most notably Karl Popper (Popper, 1972). Popper argued that scientific theories cannot be confirmed inductively and should instead be judged through attempted falsification.

Popper was wrong. Scientific evidence commonly increases or decreases rational confidence in a hypothesis without either conclusively proving or conclusively falsifying it. It has also been known for centuries that the prior probabilities of hypotheses differ. A highly specific and independently motivated theory should not receive the same prior probability as an arbitrary and excessively complicated construction designed only to fit existing observations.

Falsification is itself rarely logically decisive in empirical science. An apparent conflict between a theory and an observation may result from an incorrect auxiliary assumption, an instrument failure, an inappropriate statistical model, experimental contamination or an error in data processing. Rational scientific inference therefore requires probabilistic comparison among alternative explanations of the complete body of evidence.

Popper was correct that risky predictions and severe tests are scientifically valuable. Bayesian theory explains why. An observation gives strong support to a hypothesis when it was highly probable under that hypothesis but improbable under its competitors. A successful risky prediction therefore has a large Bayes factor.

#### 5.2.3. Ockham's razor

The heuristic that simpler hypotheses are more probable is known as Ockham's razor, after the medieval scholastic philosopher William of Ockham. Bayesian theory provides several theoretical justifications for this principle (Jefferys & Berger, 1992; MacKay, 1992).

A flexible model may fit many possible datasets, including the one actually observed. But because its probability is spread across a large range of possible observations, it need not assign high probability to the particular evidence that occurred. A simpler model makes a narrower range of predictions and may therefore assign greater probability to the observed data.

Bayesian inference therefore does not merely reward a model for fitting existing data. It balances fit against the volume of parameter space and predictions that the model permits. This provides a principled explanation of why a simpler model may be preferred over a more flexible model with a marginally better fit.

Simplicity is nevertheless dependent on the language of representation. An equation that is simple in one coordinate system may be complicated in another. The prior assigned to a hypothesis therefore reflects scientific background knowledge, representational choices and assumptions about which structures are natural.

#### 5.2.4. Bayesian model averaging and Explanation Selection

Bayesian theory provides a mathematically rigorous foundation for assigning probabilities to scientific knowledge (Berger, 1985; Jaynes, 2003). When multiple mutually exclusive models $\{M_1, \dots, M_n\}$ remain plausible, the standard Bayesian optimal approach to prediction is Bayesian Model Averaging (BMA), where predictions are weighted by each model's posterior probability (Hoeting et al., 1999):

$$P(y \mid E) = \sum_{i=1}^{n} P\,(y \mid M_i, E) P(M_i \mid E)$$

Bayesian Model Averaging preserves uncertainty and avoids the overconfidence associated with selecting a single "best" model, but its interpretation depends on the hypothesis space. In an M-closed analysis one of the candidate models is assumed to be the data-generating model; in an M-open analysis every candidate may be misspecified. Posterior probability can then concentrate on the model closest to the data-generating process without showing that the model is literally true. For purely predictive objectives under misspecification, stacking predictive distributions by cross-validation can outperform conventional posterior model probabilities.

Scientific discovery is generally M-open: the available models are provisional approximations rather than an exhaustive list containing the truth. An AI Scientist should therefore preserve evidence of absolute model inadequacy through posterior predictive checks, discrepancy models and anomaly detection, rather than interpreting relative posterior preference as confirmation that the selected model is adequate.

However, empirical science requires explicit, interpretable theories rather than just predictive blending or black-box ensemble averages. To resolve the tension between predictive averaging and explanation selection, future AI Scientists may integrate the principle of Minimal Message Length (MML) (Wallace & Boulton, 1968). MML is an information-theoretic framework that operationalizes Ockham's razor by evaluating models as data compressors. Instead of maintaining an amorphous, uninterpretable mixture distribution over all parameters, MML seeks to discover the single best explanation (or a succinct, discrete set of distinct hypotheses) by minimizing the total length of a two-part message:

$$I(M, E) = I(M) + I(E \mid M)$$

where $I(M)$ is the binary code length required to state the model's structural framework and parameters, and $I(E \mid M)$ is the code length required to encode the experimental evidence under that model.

Minimal Message Length addresses a different but complementary objective. By treating a hypothesis and the unexplained data as a two-part code, it balances explanatory fit against the complexity required to state the model. It does not make an M-open hypothesis space complete, but it provides a principled search criterion for concise, inspectable explanations. An AI Scientist could therefore use stacking for robust prediction while using MML-guided program search to identify compact candidate laws for subsequent testing.

#### 5.2.5. Graphical and relational probabilistic models

The combination of probability theory with defined structure enables graphical models, including Bayesian networks, Markov random fields, factor graphs and dynamic Bayesian networks (Pearl, 1988; Koller & Friedman, 2009). These models represent variables as nodes and probabilistic dependencies as edges or factors.

Graphical models provide a compact representation of joint probability distributions and make conditional-independence assumptions explicit. Their structure supports efficient probabilistic inference and can make complex models more comprehensible. Dynamic graphical models can describe systems that change through time, while hierarchical models can represent scientific processes operating at multiple scales.

Conventional graphical models are generally propositional: the variables and their relationships must be specified individually. The integration of probability with first-order predicate logic permits probabilities to be assigned to general logical structures involving objects, relations and quantification.

When probabilistic representations are combined with relational learning, the field is termed statistical relational learning (Getoor & Taskar, 2007; De Raedt et al., 2008). Related formalisms include probabilistic logic programs, Markov logic networks, Bayesian logic programs and probabilistic inductive logic programming.

Statistical relational learning is especially relevant to science because most scientific knowledge is relational. Molecules contain atoms connected by bonds; genes regulate other genes; organisms participate in ecosystems; patients receive treatments; and experiments connect materials, actions, conditions, measurements and conclusions. These structures cannot be represented naturally as independent feature vectors.

Future AI Scientists will require probabilistic relational representations that combine explicit entities and relations; logical background knowledge; uncertainty in facts and rules; uncertainty about alternative model structures; temporal and causal relations; provenance and experimental context; and the ability to learn new relations and rules from evidence.

#### 5.2.6. Causal reasoning

Closely connected to probabilistic reasoning is causal reasoning (Pearl, 1988, 2009; Spirtes et al., 2000). Standard statistical analysis and pattern recognition draw conclusions from associations among variables and correlations learned from data. These are insufficient when scientific questions are causal rather than merely associative.

We distinguish observing that two variables are associated and determining that changing one variable will change the other. A predictive relationship may arise from direct causation, reverse causation, a common cause, selection bias or chance.

Causal models represent not only probability distributions but also the consequences of interventions. Pearl's $do$-operator distinguishes observing $X = x$ from intervening to set $do(X = x)$. This distinction is central to experimental science. Experiments are interventions in the world. Their purpose is not simply to collect additional observations, but to break existing causal dependencies and reveal how systems respond when selected variables are deliberately changed.

The do-calculus provides rules for determining when causal effects can be identified from combinations of observational and interventional data (Pearl, 2009). Structural causal models also support counterfactual reasoning: what would have happened to a particular system under an intervention different from the one actually performed? Causal sufficiency is not required in every formulation, but unmeasured common causes can prevent identification unless they are addressed through measured covariates, valid instruments, experimental interventions or explicit latent-variable assumptions. In complex biological systems, hidden confounding is therefore a central limitation rather than a minor technical exception.

The philosophical foundations of causality remain disputed. Yet causal reasoning is indispensable to science. In physics, causality is traditionally connected to the arrow of time—causes cannot affect the past—and to the speed of light—causal influence cannot propagate outside the light cone.

Future AI Scientists must learn causal models, identify which causal relations are uncertain, design interventions that discriminate among alternative causal structures and revise those structures in response to experimental results. Recent work on foundation models for causal discovery may improve the speed and generality of causal-structure inference, but causal identifiability will continue to depend on assumptions, prior knowledge and interventions.

#### 5.2.7. Probabilistic programming

Probabilistic Programming Languages (PPLs) encode domain knowledge as executable generative models that specify how observations arise from latent variables and stochastic processes. Inference reverses this process by estimating probable hidden causes from observed effects. PPLs combine the expressive power of general programming languages with built-in or programmable inference, making it easier to construct, modify and compare rich probabilistic models systematically.

Church demonstrated how a universal probabilistic language could represent complex generative processes using a small set of language constructs (Goodman et al., 2008). Later systems—including Stan, Pyro, Turing and Gen—have provided different combinations of modelling flexibility, automatic inference, differentiability and computational efficiency (Carpenter et al., 2017; Bingham et al., 2019; Cusumano-Towner et al., 2019).

Probabilistic programs are generative models: they relate unobservable causes to observable data. They can represent hierarchical scientific theories, stochastic dynamical systems, measurement processes, missing data, latent variables and experimental uncertainty.

PPLs are therefore a natural language for AI Scientists. They enable scientific theories to be expressed as executable probabilistic models. However, deploying PPLs within autonomous AI Scientists potentially triggers severe scalability problems: MCMC Intractability: Exact Markov Chain Monte Carlo (MCMC) fails to scale to the high-dimensional data and deep architectures of modern discovery science; Variational Bias: Variational Inference (VI) scales by treating integration as optimization, but introduces structural biases that systematically underestimate posterior variance, risking pathologically overconfident hypotheses.

An AI Scientist could use PPLs to treat hypotheses as fluid code, allowing it to generate and modify probabilistic programs representing physical mechanisms; infer hidden parameters from noisy observations using scaled, hybrid inference; compute marginal likelihoods to evaluate structural model fitness; simulate counterfactual scenarios to predict intervention outcomes; and maximize expected information gain for experiment selection while minimizing experimental and computational costs.

Integrating probabilistic programming with causal structure, neural networks, and lab automation forms the self-correcting computational core of the AI Scientist.

## 5.3. Mathematics and Equations

A key pillar of the seventeenth-century scientific revolution was the recognition that mathematics, especially calculus, is the best language in which to express scientific laws. Until the twentieth century, many of the greatest scientists were also great mathematicians, Newton being arguably both the greatest scientist and the greatest mathematician. Carl Friedrich Gauss described mathematics as “the queen of the sciences”.

Mathematics remains the standard language for expressing quantitative scientific knowledge. All human scientists are trained in mathematics to a greater or lesser extent. Scientific theories are commonly expressed through equations, inequalities, geometric structures, probability distributions, optimisation problems and algorithms.

The effectiveness of mathematics in science arises partly from its precision and compressive power. A small set of equations can summarise an enormous number of observations and generate predictions for circumstances never previously examined. Mathematical theories also support deduction: once a model and its assumptions have been specified, their consequences can be derived and checked.

### 5.3.1. Equation discovery

Beginning with BACON, much research in machine scientific discovery has sought to use mathematics as the language of hypotheses and theories (Langley et al., 1987). Experimental or observational data are supplied to an AI system, which searches a space of candidate equations for expressions that fit the data.

This task is closely related to the engineering discipline of system identification, in which mathematical models of dynamical systems are inferred from measured inputs and outputs (Ljung, 1999). In conventional system identification, the general form of the model is often selected by a human and the parameters are estimated from data. Automated equation discovery seeks to infer the functional form as well as its parameters.

Two objectives matter. The first is the discovery of phenomenological equations: mathematical relationships that summarise regularities in observations and predict future data without necessarily explaining the underlying mechanism.

The second is the discovery of mechanistic equations: mathematical expressions connected to entities, processes and causal structures that explain why the observed regularity occurs.

Phenomenological relations can be scientifically valuable even before their deeper explanation is understood. Balmer’s formula for the wavelengths of hydrogen spectral lines preceded quantum mechanics. Kepler’s laws preceded Newtonian mechanics. History repeatedly shows that unexplained numerical relationships can provide clues leading to deeper theories.

#### 5.3.2. Symbolic regression

When the search for an equation is performed directly in a symbolic space, the task is termed symbolic regression. Unlike ordinary regression, symbolic regression does not assume a fixed functional form. It searches over combinations of variables, constants and mathematical operators.

Genetic programming has been one of the most successful general approaches to symbolic regression (Koza, 1992). Candidate equations are represented as expression trees and are evolved through mutation, recombination and selection. Their fitness generally reflects a combination of predictive error and expression complexity.

Sašo Džeroski and colleagues made foundational contributions to equation discovery through LAGRANGE, which extended machine discovery from static relations to systems of algebraic and differential equations describing dynamical systems, and through later grammar-based methods that incorporate prior scientific knowledge, parsimony and dimensional consistency into the search over equations (Džeroski & Todorovski, 1993; Brence et al., 2021, 2023; Mežnar et al., 2023). Stefan Kramer and colleagues have recently situated symbolic regression within the wider progression from equation discovery to autonomous discovery systems and introduced Science-Gym, a benchmark in which agents must design experiments and collect their own data before inferring the governing equations (Kramer et al., 2026; Cerrato et al., 2026).

Schmidt and Lipson integrated symbolic regression with experimental observations of dynamical systems and recovered conservation relations and other mathematical regularities from motion data (Schmidt & Lipson, 2009). Their work demonstrated that automated search could identify compact mathematical structures from measurements of physical systems.

Sparse Identification of Nonlinear Dynamics, or SINDy, assumes that the governing dynamics contain only a small number of terms drawn from a larger library of candidate functions. Sparse regression is then used to identify the active terms (Brunton et al., 2016). This provides a computational implementation of Ockham's razor: prefer the smallest set of terms capable of explaining the observed dynamics.

A difficulty is that a system may possess a simple mathematical description only in an appropriate coordinate system. Champion et al. combined deep autoencoders with sparse equation discovery to learn both a reduced coordinate system and the governing dynamics within it (Champion et al., 2019). The neural network learns the representation, while symbolic sparse regression identifies the equations.

AI Feynman combined dimensional analysis, neural networks, symmetry detection, separability and symbolic search to recover equations from data (Udrescu & Tegmark, 2020). By identifying mathematical structure before conducting an exhaustive symbolic search, the system substantially reduced the effective hypothesis space.

These methods demonstrate that neural and symbolic approaches are complementary. Neural networks are effective at learning representations and approximating complex functions. Symbolic methods are effective at producing concise, explicit and mathematically interpretable hypotheses.

#### 5.3.3. Integrating data with scientific theory

A central limitation of conventional symbolic regression is that it searches for equations that fit data, rather than equations that are consistent with established scientific knowledge. Many mathematically different expressions may fit a small or noisy dataset equally well, while only a few are compatible with the relevant background theory.

AI-Descartes combines symbolic regression with logical deduction (Cornelio et al., 2023). Candidate equations generated from data are assessed against axioms representing existing scientific knowledge. The system can prove that a candidate is derivable from the background theory, demonstrate that it is inconsistent with that theory or quantify how far it is from a derivable result.

AI-Hilbert develops this idea further by discovering polynomial laws that are simultaneously consistent with experimental data and axiomatic background knowledge (Cory-Wright et al., 2024). The approach uses mathematical optimisation and certificates from real algebraic geometry to derive and formally verify candidate laws. It can also identify inconsistent elements within a body of background assumptions.

AI-Descartes and AI-Hilbert exemplify the type of hybrid reasoning required by future AI Scientists. Empirical data constrain which equations are plausible; background theory constrains which are meaningful; and formal deduction determines whether the proposed equations follow from the stated assumptions.

#### 5.3.4. Large Language Models for equation discovery

LLMs provide a new mechanism for equation discovery because they contain extensive implicit knowledge of scientific concepts, mathematical notation and known equation forms. Rather than searching blindly through expression trees, an LLM can propose equation structures informed by the scientific context.

LLM-SR represents candidate equations as programs and combines LLM-generated hypotheses with evolutionary search and numerical parameter optimisation (Shojaee et al., 2025a). The LLM supplies scientific priors and proposes candidate program structures, while explicit evaluation against data provides a closed feedback loop.

This approach potentially permits richer mathematical representations than conventional expression trees. A candidate model may contain conditional statements, recursion, numerical solvers or calls to other functions. Treating equations as programs therefore connects equation discovery with the more general representation of scientific knowledge as executable computation.

But equation discovery by LLMs remains far from solved. Standard benchmarks often contain famous equations likely to have appeared in model-training data, making it difficult to distinguish genuine discovery from memorisation. LLM-SRBench introduced transformed and synthetic problems designed to reduce this contamination and found that even the best systems solved only a minority of the equations exactly (Shojaee et al., 2025b).

Future systems will therefore need to combine LLM-based scientific priors; numerical and symbolic search; dimensional and symmetry constraints; experimental data; causal and mechanistic background knowledge; formal mathematical verification; and tests on genuinely novel systems.

#### 5.3.5. Depth and generality

A central question is the depth and generality of a discovered equation. Newton’s laws of motion are deeper and more general than Kepler’s laws. Quantum mechanics is deeper than the empirical formulae for atomic spectra that preceded it.

A mathematically accurate relationship is not necessarily an explanatory scientific theory. A sufficiently flexible algorithm may reproduce a dataset while providing no insight into the entities, mechanisms or invariances responsible for the observations.

The depth of an equation may be assessed through several related criteria: the range of phenomena it explains; the number of previously separate laws it unifies; its invariance under transformations; its connection to underlying mechanisms; the accuracy of its novel predictions; the simplicity of its assumptions; its derivability from more fundamental principles; and its capacity to support successful counterfactual reasoning.

These criteria cannot generally be reduced to predictive accuracy on a held-out dataset. Future AI Scientists must search not merely for equations that fit observations, but for hierarchies of mathematical models operating at different levels of description.

A phenomenological model may be appropriate for prediction at one scale, while a more fundamental model explains why that phenomenological relationship holds. An AI Scientist should represent the relationship between such models rather than insisting that only one level is scientifically legitimate.

#### 5.3.6. Mathematical understanding

Future AI Scientists must understand existing scientific theories expressed in mathematics and understand how to apply them. Existing equation-discovery systems can observe a classical physical system such as a double pendulum and find equations that predict aspects of its behaviour. They remain far from the general mathematical understanding displayed by an expert physicist.

For example, the opening chapters of Landau and Lifshitz's Mechanics introduce the principle of least action and the Lagrangian formulation of mechanics (Landau & Lifshitz, 1976).

The reader is then invited to solve problems as forming the Lagrangian of a double-pendulum system. To achieve this they must:

- Identify suitable generalised coordinates;
- Express the positions and velocities of both masses;
- Calculate the kinetic and potential energies;
- Form the Lagrangian ($L = T - V$);
- Apply the Euler–Lagrange equations; and
- Derive the coupled equations of motion. This task requires more than fitting equations to trajectories. It requires recognising the system as an instance of a general theoretical framework, selecting an appropriate representation and deriving the equations from first principles.

Future AI Scientists must be able to move in both directions:

- from data to equations, through induction and equation discovery; and
- from theory to predictions, through mathematical modelling and deduction. They must also identify when a discovered equation is an approximation, determine the conditions under which it applies and recognise which neglected terms become important outside that regime.

#### 5.3.7. AI mathematicians

Future AI Scientists will also need to understand and create mathematics generally. To achieve this, they will be integrated with AI mathematicians.

Mathematics involves identifying patterns, formulating definitions and conjectures, and then proving the resulting propositions deductively from axioms. Computers arose partly from the attempt to formalise and mechanise mathematical reasoning (Turing, 1936).

Machine learning is well suited to identifying patterns and proposing conjectures. Davies et al. used machine learning to guide human mathematical intuition and produced new results concerning knot theory and representation theory (Davies et al., 2021).

FunSearch combined an LLM with an evolutionary search and an automatic evaluator. It discovered improved constructions for the cap-set problem and new heuristics for online bin packing (Romera-Paredes et al., 2024). Its importance lies in the integration of generative proposals with an objective evaluator that rejects invalid outputs.

AlphaGeometry combined a neural language model with a symbolic deduction engine and solved 25 of 30 selected Olympiad-level geometry problems (Trinh et al., 2024). It illustrates the power of neuro-symbolic integration: the neural component guides search, while the symbolic component ensures deductive validity.

AlphaProof uses reinforcement learning within the Lean theorem-proving environment. It produces proofs that are checked by the Lean kernel, ensuring that successful outputs are formally valid. Together with AlphaGeometry 2, it achieved a score equivalent to an International Mathematical Olympiad silver medal on the 2024 competition problems (Hubert et al., 2026).

These developments show substantial progress in three complementary aspects of AI mathematics: pattern recognition and conjecture generation, search for mathematical constructions and algorithms, and formally verified theorem proving.

The requirement for formal verification is especially important. An informal mathematical argument generated by an LLM may contain a subtle error that is difficult to detect. A proof accepted by a small trusted proof-checking kernel provides a much stronger guarantee of correctness.

A key milestone was reached in 2026, when an internal OpenAI reasoning model autonomously disproved Erdős's long-standing planar unit-distance conjecture. The model constructed, for infinitely many $n$, planar point sets containing at least $n^{1+\delta}$ unit distances for some fixed $\delta > 0$, using an unexpected connection between discrete geometry and deep results from algebraic number theory (OpenAI, 2026). The original proof was subsequently checked, simplified and contextualised by an external group of mathematicians including Noga Alon and Tim Gowers. Gowers described the result as a milestone in AI mathematics and judged it to be of the standard expected by the strongest mathematical journals (Alon et al., 2026). This is qualitatively different from solving benchmark or Olympiad problems: a general-purpose AI system autonomously resolved a prominent, long-standing open problem and introduced a surprising mathematical construction that had escaped the expert community.

#### 5.3.8. AI Mathematics within an AI Scientist

An AI mathematician and an AI Scientist are not the same. Mathematics establishes conclusions relative to axioms through proof. Empirical science establishes provisional beliefs about the physical world through observation and experiment. An AI mathematician can prove that a conclusion follows from a specified mathematical model. It cannot establish through proof alone that the model describes the physical world. That requires experimental evidence.

Future AI Scientists will integrate mathematical and empirical reasoning: formulate a scientific hypothesis; express the hypothesis as a mathematical model; derive its observable consequences; compare those consequences with experimental data; revise the model or its probability; formulate new mathematical conjectures required by the model; and use AI mathematicians and proof assistants to prove relevant results.

Mathematical proof will therefore become an internal tool of AI Scientists. It will certify deductions, verify algorithms, analyse limits and invariances, and establish whether proposed equations follow from background theories.

One problem with AI mathematicians, compared with AI Scientists, is that the philosophy of mathematics is less clear than the philosophy of empirical science. There is no consensus on whether mathematical objects are discovered or invented, what constitutes mathematical understanding, or how informal explanatory insight relates to formal proof.

Yet mathematics provides the most reliable domain for verified AI reasoning. Proof assistants create environments in which every inference can be checked mechanically. The integration of this rigour with empirical experimentation will be essential to AI Scientists capable not merely of fitting data, but of developing deep, general and trustworthy scientific theories.

## 5.4. Neural Networks

Mathematics, logic and probability are the traditional languages of science. Until recently, they were also the preferred languages in which machine-learning models expressed their hypotheses. This changed with the rise of Deep Neural Networks (DNNs) (LeCun et al., 2015; Goodfellow et al., 2016). DNNs are now responsible for many of the most successful applications of machine learning to science, including protein-structure and molecular-interaction prediction, protein design, materials discovery, weather forecasting, medical imaging and the construction of fast surrogate models for physical simulations (Jumper et al., 2021; Merchant et al., 2023; Abramson et al., 2024; Bodnar et al., 2025).

Traditional scientific languages are symbolic and explicit. The meaning of a mathematical variable, logical predicate or probabilistic parameter can, in principle, be stated independently of the particular calculation in which it occurs. The knowledge expressed in a symbolic model can therefore be inspected, communicated and manipulated through explicit rules.

In contrast, DNNs express knowledge in a sub-symbolic and implicit form. Their knowledge is distributed across large numbers of numerical parameters—weights, biases, embeddings and activation values—which collectively define a complex function. Individual parameters generally have no simple scientific interpretation. The same concept may be distributed across many parameters, while a single parameter may contribute to many different concepts.

A trained DNN is therefore both a model and an executable language. Its parameters specify a computational transformation from inputs to outputs, but they do not normally provide an explicit statement of the scientific regularities that the network has learned. The network may make highly accurate predictions without revealing those regularities in a form that a human scientist can understand.

### 5.4.1. Differentiable representation learning

The success of DNNs is based on their use of multiple layers of differentiable transformations. During training, errors in the network's predictions are propagated backwards through these layers, allowing gradient-based optimisation to alter millions or billions of parameters simultaneously (Rumelhart et al., 1986).

Deep networks are particularly powerful because they learn representations as well as predictions. Earlier machine-learning systems generally depended on human experts to select informative descriptors or features. DNNs can instead learn a hierarchy of representations directly from relatively weak inputs. In computer vision, early layers may represent local edges and textures, while later layers represent shapes and objects. In molecular science, a network may learn representations of local chemical environments, functional groups, three-dimensional geometry and long-range interactions.

This process is termed representation learning. It enables a DNN to transform an initially poor representation—such as image pixels, an amino-acid sequence or a list of atomic coordinates—into a latent representation in which the target prediction is easier to make. End-to-end learning jointly optimises the representation and the final scientific task.

The weights of most DNNs are real numbers, and modern systems frequently train or operate using reduced numerical precision. Complex-valued weights may be advantageous where phase, oscillation, interference, rotation or wave propagation are fundamental, because they represent amplitude and phase jointly and can provide natural representations for electromagnetic fields, quantum systems, magnetic-resonance data, acoustics and signal processing (Trabelsi et al., 2018; Lee et al., 2022). Complex-valued networks nevertheless remain uncommon outside domains in which complex quantities arise naturally, partly because real-valued software, hardware and optimisation methods are more mature.

#### 5.4.2. Architectural inductive biases

DNNs are not a single model class. Different network architectures impose different inductive biases: assumptions about the structures that are likely to occur in the data.

Convolutional neural networks encode the assumption that similar local patterns may occur at different spatial positions. Recurrent networks encode sequential processing and memory. Transformers use attention to model interactions between elements irrespective of their separation in a sequence (Vaswani et al., 2017). Graph Neural Networks (GNNs) represent entities and their relationships directly, making them particularly suitable for molecules, materials, biological networks and physical systems.

Scientific networks increasingly encode known physical symmetries. A scalar energy should remain unchanged if a molecule is rotated, whereas a predicted force should rotate with the molecule. Equivariant neural networks guarantee that their outputs transform correctly under rotations, reflections, translations or other symmetry operations. The NequIP architecture, for example, incorporates three-dimensional Euclidean equivariance into learned interatomic potentials and achieves high accuracy with substantially fewer training examples than many earlier approaches (Batzner et al., 2022).

The incorporation of symmetry demonstrates that the distinction between symbolic scientific knowledge and neural learning is not absolute. Known scientific principles can be incorporated into the structure of a DNN rather than merely supplied as training examples. Conservation laws, geometric symmetries, dimensional relationships and differential equations may be encoded through architecture, loss functions or differentiable simulators.

Diffusion models provide another important architecture. They learn to reverse a process in which structure is progressively corrupted by noise. Once trained, they can generate new samples by starting from noise and repeatedly denoising it. Diffusion models are now used to generate images, molecules, proteins and three-dimensional biomolecular structures. AlphaFold 3 uses a diffusion-based component to generate the coordinates of biomolecular complexes (Abramson et al., 2024).

Neural operators extend DNNs from mappings between finite-dimensional vectors to mappings between function spaces. They are important as learned approximations to families of differential-equation solutions. Their use as surrogate models, together with their limitations, is discussed in Section 8.

#### 5.4.3. Scale and foundation models

No other form of machine learning is currently capable of learning models comparable in size to the largest DNNs or of learning effectively from comparable quantities of heterogeneous data. DNNs can contain billions or even trillions of adjustable parameters and can be trained using text, images, sequences, structures, simulations and experimental measurements.

We do not yet know whether this dominance arises from an intrinsic property of neural networks as a hypothesis language, or from the vastly greater research effort, data, specialised hardware and financial resources devoted to them. The differentiability of neural networks permits efficient optimisation using gradient descent, while their modular architectures map naturally onto highly parallel graphical-processing hardware. These practical properties have enabled a cycle in which larger datasets, larger models and greater computation repeatedly improve performance.

A foundation model is trained on a broad body of data and subsequently adapted to multiple tasks. Instead of developing a separate model for every scientific problem, a foundation model learns a general latent representation of a domain. It can then be prompted, fine-tuned or connected to specialist modules.

Scientific foundation models have been developed for language, proteins, molecules, materials, genomes, weather and the wider Earth system. Their importance lies not simply in scale, but in their potential to transfer knowledge between tasks. A protein model trained without explicit structural labels can nevertheless acquire latent representations of secondary structure, residue contacts, function and evolutionary relationships.

Foundation models also permit generative science. Rather than only predicting properties of existing objects, they can propose new proteins, molecules and materials that satisfy specified constraints. But generation remains dependent on the quality and coverage of the training data, and generated objects require computational or experimental validation.

#### 5.4.4. AlphaFold and protein foundation models

The highest-profile application of machine learning to science has been the development of AlphaFold 2 for predicting the three-dimensional structures of proteins from their amino-acid sequences (Jumper et al., 2021).

Anfinsen’s work established that, under appropriate conditions, the information required for a protein to adopt its native structure is encoded in its amino-acid sequence (Anfinsen, 1973). Researchers therefore spent decades attempting to learn the mapping from sequence to structure. Early machine-learning work concentrated on intermediate tasks such as predicting secondary structure, first through inductive learning and subsequently through comprehensible machine-induced rules (King, 1987; King & Sternberg, 1990).

Progress was driven by three mutually reinforcing developments: an enormous increase in available computing power; an enormous increase in protein sequence and structural data; and the development of learning algorithms capable of representing much more complex functions.

The effectiveness of protein-structure-prediction systems is evaluated objectively through the Critical Assessment of Structure Prediction (CASP), a biennial blind experiment in which participants predict structures that have not yet been publicly released (Moult et al., 1995). In CASP14 in 2020, AlphaFold 2 achieved a level of accuracy substantially beyond that of previous systems, marking a qualitative step forward in protein-structure prediction.

AlphaFold 2 combines attention, geometric reasoning, evolutionary information and iterative refinement in a highly complex end-to-end network. It does not explicitly derive a protein structure from a compact symbolic theory of protein folding. Instead, it learns a powerful implicit mapping from sequence and evolutionary information to spatial structure. Although components of its operation are understood, a complete mechanistic explanation of why the network works so well remains unavailable.

AlphaFold 3 broadened the task from predicting protein structures to predicting the joint structures and interactions of proteins, nucleic acids, small molecules, ions and modified residues (Abramson et al., 2024). This matters because biological function generally depends on molecular interaction rather than on isolated protein structure.

Protein language models provide a complementary approach. They treat amino-acid sequences as a language generated by evolution. When trained on hundreds of millions of protein sequences, structural and functional information emerges within their latent representations.

ESMFold demonstrated that atomic-level protein structures could be inferred directly from the representations of a large protein language model, without requiring the expensive construction of a multiple-sequence alignment for every prediction (Lin et al., 2023). The model enabled the prediction of structures for more than 600 million metagenomic proteins.

ESM3 extended protein language modelling across sequence, structure and function. It generated a functional fluorescent protein with only 58% sequence identity to the nearest known fluorescent proteins—a degree of divergence estimated to correspond to more than 500 million years of natural evolution (Hayes et al., 2025). This demonstrates that neural networks can do more than interpolate among known natural examples: they can generate functional biological structures in remote regions of sequence space.

#### 5.4.5. Neural networks across the sciences

The success of DNNs in science extends far beyond structural biology. In materials science, GNoME used graph neural networks and large-scale active learning to predict the stability of inorganic crystals. The system identified 2.2 million candidate stable structures, including 381,000 structures on an updated convex hull of thermodynamic stability, representing an order-of-magnitude expansion of the previously known computational catalogue (Merchant et al., 2023).

In weather forecasting, Pangu-Weather and GraphCast demonstrated that deep neural networks trained on historical atmospheric reanalysis data could equal or exceed leading deterministic numerical systems on many medium-range forecasting measures while producing forecasts much more rapidly (Bi et al., 2023; Lam et al., 2023).

Aurora extended this approach into a foundation model for the Earth system. It was pretrained on more than one million hours of diverse geophysical data and subsequently adapted to weather, air-pollution, ocean-wave and tropical-cyclone forecasting tasks (Bodnar et al., 2025).

These models need not replace mathematical physics completely. NeuralGCM combines a differentiable numerical solver for large-scale atmospheric dynamics with learned neural representations of unresolved physical processes. It can produce weather and climate simulations while retaining an explicit dynamical core (Kochkov et al., 2024). Such hybrid models suggest that the most effective scientific systems will often integrate neural learning with established mathematical theories rather than choosing exclusively between them.

DNNs are also increasingly used as learned interatomic potentials, surrogate models for quantum-mechanical calculations, emulators of fluid and climate simulations, generators of molecular structures and components of inverse-design systems. Their speed permits quantities of candidate hypotheses or designs to be evaluated that would be infeasible using direct physical simulation alone.

#### 5.4.6. Neural networks and scientific explanation

The success of DNNs reopens the century-old logical-positivist debate about the relationship between observed and theoretical entities in science (Carnap, 1974). Logical positivists emphasised observable relationships and were suspicious of metaphysical commitments to entities that could not be observed directly. Scientific practice, however, has traditionally been formulated through unobservable theoretical entities, including fields, wavefunctions, genes, quarks and the electromagnetic four-potential.

A DNN can appear strongly instrumentalist. It may learn a direct mapping from observations to predictions without representing an explicit theory of the underlying entities and mechanisms. If a network accurately predicts the result of an experiment, it may be useful even when scientists cannot explain how it reaches that prediction.

This raises a fundamental question: does successful prediction constitute scientific understanding? Prediction is necessary but not sufficient for deep scientific understanding. A model may predict accurately because it has identified a genuine causal structure, because it has captured a stable but unexplained correlation, or because it has exploited an accidental regularity in its training data. These possibilities may be indistinguishable when the model is evaluated only on data drawn from the same distribution as its training examples.

DNNs are particularly vulnerable to spurious correlations; distribution shift; hidden data leakage; poor calibration; adversarial or anomalous inputs; failures outside the range of the training data; and the absence of explicit causal interpretation.

At the same time, DNNs complicate the simple distinction between prediction and theory. Latent representations learned from data may encode scientifically meaningful entities and structures that were not supplied explicitly. Protein language models acquire representations of structure and function from sequences. Equivariant networks acquire tensorial representations consistent with physical geometry. Neural operators learn approximations to families of dynamical laws.

The difficulty is that these representations are usually implicit. Scientists must infer their meaning by analysing the network after training. Mechanistic interpretability seeks to reverse-engineer neural networks and identify the algorithms, features and concepts represented internally. But interpreting networks with billions of parameters remains an unsolved problem.

#### 5.4.7. Neural networks within AI Scientists

Future AI Scientists will require DNNs, but they cannot be based on DNNs alone. DNNs provide exceptional capabilities for perception and pattern recognition; learning representations from raw data; analysing images, spectra, sequences and sensor outputs; approximating complex functions and simulations; translating between natural and formal languages; recognising anomalies; and proposing hypotheses and experimental actions. Their weaknesses are complementary to the strengths of formal methods. Neural networks do not naturally provide explicit semantics, valid logical proofs, calibrated probabilities, causal guarantees or transparent scientific explanations.

The most powerful future architecture will therefore likely be neuro-symbolic and probabilistic. A DNN may generate a candidate representation, hypothesis, molecule or experimental protocol. A symbolic system can check logical consistency and derive consequences. A probabilistic system can represent uncertainty and compare alternatives. A causal model can determine which intervention would discriminate among competing explanations. A robot can execute the selected experiment, and the resulting observations can be used to update both the explicit knowledge base and the neural models.

Within such an AI Scientist, neural networks will constitute the flexible, high-capacity, pattern-learning component. Logic, mathematics, probability and programs will provide explicit meaning, verification and explanation.

The central scientific opportunity is not to choose between symbolic and neural representations. It is to integrate them. DNNs can discover useful implicit structures in data at unprecedented scale. Formal languages can transform those structures into hypotheses that can be understood, checked, communicated and experimentally tested.

## 6. Reading and Formalising the Scientific Literature

Science has produced more literature than any human can read. Important evidence is scattered across papers, figures, tables, supplementary files and databases. An AI Scientist must turn this vast and disorderly record into explicit, testable knowledge.

Scientific prose is often incomplete and context-dependent. Authors omit assumptions that they expect specialist readers to supply, use several names for the same entity and occasionally use the same term for different entities. Claims may be qualified in one section of a paper but restated without qualification elsewhere. Experimental methods commonly contain expressions such as 'overnight', 'at room temperature', 'washed thoroughly' or 'using the standard procedure', whose meanings depend on tacit disciplinary or laboratory knowledge.

Natural language nevertheless remains indispensable. It is flexible enough to describe new observations and concepts before a suitable formal vocabulary exists. It enables scientists to explain motivations, analogies, mechanisms, uncertainties and significance. Human science is therefore conducted through a mixture of prose, mathematics, diagrams, tables, computer programs and experimental records.

The scientific literature is vast. Bibliographic infrastructures now contain records for hundreds of millions of research outputs (Tkaczyk et al., 2026). No human scientist can read more than a small fraction of the relevant literature during a lifetime.

This creates a fundamental limitation on human science. Relevant findings may remain unnoticed because they were published in another discipline, use unfamiliar terminology or are buried in supplementary information. Contradictory claims may persist without being recognised as contradictions. Experiments may be repeated unnecessarily, while negative or inconclusive results disappear from the accessible record. The automation of scientific reading is therefore essential to the development of AI Scientists.

### 6.1. The scientific literature as a knowledge system

A scientific paper is not simply a continuous body of prose. It is a structured argument connecting background knowledge, research questions, methods, observations, analyses and conclusions. Different sections have different epistemic functions.

The introduction describes the accepted background, identifies a gap in knowledge and motivates the investigation. The methods describe how observations were produced. The results report measurements and analyses. The discussion interprets those results, relates them to previous work and proposes explanations. Supplementary materials may contain essential details, additional analyses and negative findings that do not appear in the main text.

An AI Scientist must distinguish among these functions. A statement in the introduction may report a generally accepted view, summarise another paper or merely motivate the present study. A statement in the discussion may be an interpretation rather than a direct observation. Treating every sentence as an equivalent factual assertion would destroy much of the epistemic structure of the paper.

The scientific literature is also distributed across several interconnected objects: journal articles and conference papers; preprints and revised versions; supplementary files; datasets and database records; software and computational notebooks; protocols; corrections, expressions of concern and retractions; and citing and cited papers.

The unit of scientific knowledge is therefore not the isolated document. It is a network of claims, evidence, methods, data, versions and responses. Reading the literature requires following these relationships.

Scientific claims must also be interpreted temporally. A paper may accurately represent the evidence available when it was published but later be contradicted or superseded. Preprints may differ substantially from their final versions. A paper may be corrected or retracted after it has already been incorporated into reviews, databases and language-model training corpora.

A machine-readable scientific literature must therefore preserve provenance and version information. An AI Scientist needs to know not only what was stated, but who stated it, where, when, on the basis of which evidence and whether the claim was later corrected, challenged or replicated.

## 6.2. Text mining

Before the rise of Large Language Models (LLMs), text-mining and Natural Language Processing systems dominated attempts to extract knowledge from scientific papers (Feldman & Sanger, 2007). Text-mining systems can process very large document collections and identify specified forms of information at scale.

Traditional scientific text mining is commonly organised as a pipeline. Documents are divided into sections, sentences and tokens. Linguistic tools identify grammatical structure. Named entities are recognised and linked to standard identifiers. Relations and events involving those entities are extracted, and the resulting records are stored in databases or knowledge graphs.

A representative biomedical task is the recognition and normalisation of gene and protein names. The system must first identify the words referring to a biological entity and then link each mention to the correct database record. This apparently simple task is complicated by the use of several names for the same entity; identical names for different entities; abbreviations defined only once in a paper; differences between species; overlap between gene, RNA and protein names; ordinary English words used as gene symbols; and changes in nomenclature over time.

Entity recognition alone is insufficient. The system must determine what relationship the paper asserts between the recognised entities. It may need to distinguish, for example, between:

- Protein A activates protein B.
- Protein A does not activate protein B.
- Protein A may activate protein B.
- Previous studies reported that protein A activates protein B.
- We tested, but did not confirm, whether protein A activates protein B. These statements contain similar entities and vocabulary but make very different scientific claims.

Scientific text mining has benefited greatly from shared annotated corpora and benchmark tasks. These resources allow systems to be trained and evaluated on common examples. Major tasks have included gene and protein recognition, chemical and disease identification, relation extraction, molecular-event extraction, citation classification and the identification of statements supported by experimental evidence.

Contextual neural language models improved performance substantially. SciBERT was trained on scientific publications and adapted to several scientific-language tasks (Beltagy et al., 2019). BioBERT and PubMedBERT were trained on biomedical literature and achieved strong results in entity recognition, relation extraction and biomedical question answering (Lee et al., 2020; Gu et al., 2021). Software such as scispaCy made scientific NLP models available within practical processing pipelines (Neumann et al., 2019).

Traditional text-mining systems retain several advantages. Their target outputs are usually defined explicitly, their accuracy can be measured against annotated examples, and individual components can be inspected. Their weakness is that each new task often requires a new corpus, annotation scheme and specialised pipeline. Errors also propagate between stages: an entity that is not recognised cannot participate in a subsequently extracted relation.

LLMs provide a more flexible alternative. The same model can extract different classes of information in response to natural-language instructions and examples. But specialist models may remain superior when exact span recognition, exhaustive retrieval and consistent application of a fixed annotation policy are required. LLM-based extraction should therefore complement, rather than simply replace, established text-mining methods.

### 6.3. Searching and retrieving the literature

The first requirement for reading the scientific literature is finding the relevant documents. Traditional bibliographic search is based principally on words, phrases, subject classifications, citations and metadata. Modern semantic search systems use learned representations to retrieve documents whose meaning is related to a query even when they do not contain the same terminology.

Scientific retrieval is unusually difficult because terminology varies between disciplines and changes through time. The same underlying concept may be described differently in molecular biology, medicine and chemistry. Conversely, the same term may have unrelated meanings in different fields.

Citation networks provide information that is complementary to textual similarity. A paper may cite the work it extends, challenges or applies even when the vocabulary differs substantially. Co-citation and bibliographic-coupling methods can therefore identify intellectual relationships that are not apparent from the text alone.

Retrieval should be iterative. Initial papers reveal new terminology, authors, methods and cited work, which in turn guide subsequent searches. The system should also search deliberately for evidence that opposes its current interpretation rather than retrieving only papers that appear to support it.

The objective is not to retrieve the largest possible number of papers. It is to identify the smallest body of literature sufficient to represent the relevant evidence, methods, theoretical positions and disagreements. Reading everything is neither necessary nor computationally efficient, even for an AI Scientist.

### 6.4. Literature synthesis

Retrieving papers is different from synthesising their contents. A useful synthesis must identify the principal claims, determine how they are supported and explain how the papers relate to one another.

LLMs are well suited to summarising individual documents, but scientific synthesis requires more than fluent compression. The system must distinguish consensus from repetition, independent evidence from repeated citation of the same original result, and direct experimental support from speculation.

Retrieval-Augmented Generation connects a language model to an external document collection (Lewis et al., 2020). Relevant passages are retrieved and placed in the model’s context, allowing the generated answer to be grounded in identifiable sources rather than relying solely on information stored implicitly in the model’s parameters.

OpenScholar applies this approach to scientific literature synthesis. It combines retrieval, reranking, generation and self-evaluation with a large collection of open-access papers, producing answers with explicit citations to supporting sources (Asai et al., 2026). Such systems are much better suited to scientific use than ungrounded conversational models.

Yet the presence of a citation does not guarantee that the citation supports the associated claim. A system may retrieve a relevant paper but misrepresent its conclusion, omit an important qualification or cite a review when the primary evidence is required. Citation accuracy must therefore be checked at the level of individual claims.

A trustworthy literature synthesis should distinguish at least the following: directly reported observations; statistical analyses of those observations; causal or mechanistic interpretations; hypotheses proposed by the authors; results reproduced independently; findings supported only by one laboratory; conflicting evidence; methodological limitations; and conclusions generated by the reviewing system itself.

The synthesis should also avoid counting multiple papers as independent evidence when they analyse the same dataset or derive from the same experimental cohort. Independence of evidence matters more than the number of citations.

Systematic reviews provide an important model. Their methods aim to define search criteria, record inclusion and exclusion decisions, assess the quality of evidence and make the synthesis reproducible. Machine learning can accelerate document screening and information extraction, but the criteria and decisions should remain explicit (Marshall & Wallace, 2019).

Literature synthesis can also generate new hypotheses by connecting previously separate bodies of work. Because this function concerns the creation of scientific ideas rather than reading alone, literature-based discovery is treated in Section 7.

### 6.5. Multimodal scientific documents

A large proportion of scientific information is not contained in the prose of a paper. It appears in tables, figures, equations, chemical structures, spectra, images and supplementary files.

Figures are often the primary presentation of experimental evidence. Their captions may not describe every relevant observation, while the main text may discuss only a selected part of the figure. Tables may contain numerical results that never appear in sentences. Experimental conditions may be distributed across the methods, a figure legend and a supplementary spreadsheet.

Reading a scientific paper therefore requires multimodal interpretation. A system must connect statements in the text to plotted measurements; error bars and statistical annotations; labels and legends; microscopy or medical images; molecular and protein structures; mathematical equations; tabulated values; and supplementary methods and data.

Document layout also conveys meaning. A heading indicates the role of a passage; a footnote may qualify a table; a superscript may connect a value to a statistical test. Converting a paper into an undifferentiated stream of text discards this structure.

Scientific literature systems should preserve the coordinates and document context of extracted information. A claim should be linked not only to a paper, but to the particular sentence, table cell, figure panel or equation that supports it.

The need to integrate text and figures makes access to complete articles important. Abstracts are insufficient for many scientific questions. They compress the authors' preferred conclusion and frequently omit limitations, unsuccessful results and detailed methods.

### 6.6. Reliability, hallucination and provenance

A system designed to read science must distinguish linguistic plausibility from scientific truth. An LLM is trained to generate probable sequences of tokens. It is not inherently constrained to reproduce claims that are true or even claims that occurred in its source material. It can therefore invent facts, quotations, papers and references.

Galactica demonstrated both the potential and the danger of scientific language models. It was trained on approximately 48 million scientific papers, reference materials and related examples, and was intended to provide a general interface to scientific knowledge (Taylor et al., 2022). But it also generated authoritative-sounding false claims and fabricated references. Its public demonstration was withdrawn shortly after release.

The failure was not caused simply by insufficient scientific training. Training on more scientific language can make a model more fluent in scientific discourse without making every generated statement more reliable.

Scientific literature systems therefore require external verification. Each substantive statement should be connected to an identified source, and the source should be checked to determine whether it supports the precise claim. References should be retrieved from bibliographic databases rather than generated freely as text. This record would make the literature analysis reproducible. Another scientist or AI system could inspect the same evidence and determine whether the synthesis was justified.

The system should also distinguish absence of evidence from evidence of absence. Failure to retrieve a paper supporting a claim does not establish that no such paper exists. The literature may be inaccessible, badly indexed, unpublished or expressed using unexpected terminology.

### 6.7. Writing the scientific literature

Scientific communication moves in two directions. AI Scientists must read the human scientific literature, but they must also communicate their own discoveries to humans and other machines. LLMs are exceptionally capable of transforming structured information into fluent prose. They can draft abstracts, methods, results and discussions; improve grammatical quality; adapt writing to a journal's style; and translate scientific text between languages. Their use in scientific writing is already widespread and increasing, although prevalence differs among disciplines and linguistic communities (Liang et al., 2025). This creates opportunities to improve clarity and reduce linguistic barriers, but also risks increasing the volume of superficially plausible and poorly verified scientific text.

Scientific writing is not merely stylistic presentation. A paper makes a sequence of epistemic commitments. It states what was known before the study, what was done, what was observed and what follows from the evidence. An automatically written paper must preserve these distinctions.

Every factual statement in a machine-generated paper should be traceable to one of the following: an identified external publication, a specified dataset, a recorded experiment, an executable analysis, and a formally verified deduction.

The system must not transform uncertainty into certainty or association into causation merely to produce smoother prose. Negative results and failed experiments should be reported when they are relevant to the interpretation.

Citations must be selected because they support particular claims, not merely because their titles or abstracts appear related. The system should distinguish primary evidence from reviews and cite the primary publication where appropriate.

The data-to-paper system demonstrates how several language-model agents can transform datasets and analyses into a paper while preserving links among the data, code, results and resulting text (Ifargan et al., 2025). This type of provenance is essential. A reader must be able to move from a sentence in the paper to the analysis or experiment that justifies it.

Future AI Scientists should produce two connected scientific records (King et al., 2009).

The first should be human-readable: a conventional paper or report explaining the question, methods, evidence, conclusions, limitations and significance.

The second should be machine-readable: a structured record of the claims, evidence, protocols, data, analyses, uncertainty, provenance and relationships to previous literature.

The human-readable publication permits explanation, evaluation and criticism. The machine-readable record permits computational checking, integration and reuse.

An AI Scientist should therefore not merely write persuasive scientific prose. It should produce a transparent and verifiable account of how its conclusions follow from the literature and from its own observations.

### 6.8. Paywalls and the “Dark Data” Problem

The vision of an AI Scientist reading the entirety of human scientific knowledge assumes that this literature is perfectly digitised and accessible. In reality, an enormous volume of scientific data remains “dark”—trapped behind publisher paywalls, hidden in non-digitised physical laboratory notebooks, obscured within legacy software formats, or siloed in proprietary corporate databases. While open-access mandates are improving the availability of recent papers, the historical record remains highly fragmented. For an AI Scientist to achieve true, unbiased literature synthesis, legal and technical frameworks must be established to allow machine-reading across paywalled corpora without violating copyright, alongside concerted efforts to digitise and standardise the vast archives of historical dark data (Heidorn, 2008).

## 7. Hypothesis Generation and Scientific Creativity

Science begins when an observation demands an explanation or a problem demands a solution. Generating plausible text is easy. Generating a new hypothesis that is coherent, explanatory and experimentally testable is hard. This is the creative core of an AI Scientist.

A scientific hypothesis is not simply any sentence that could be tested. It is a proposed extension or revision of scientific knowledge that explains observations, predicts new phenomena or identifies a previously unrecognised regularity. Hypothesis generation must be connected to background knowledge, uncertainty, experimental feasibility and the possibility of empirical discrimination.

The hypothesis space is determined by the language available to the scientist. A system that can vary only numerical parameters can discover only hypotheses expressible as parameter settings. A system that can invent predicates, entities, equations, mechanisms and experimental concepts can transform the representation of the problem itself. Open-ended scientific discovery therefore depends on representational creativity.

### 7.1. Problem formulation and question generation

Before a hypothesis can be generated, a scientific problem must be formulated. Human research commonly begins with an anomaly, a practical objective, a gap in the literature, a contradiction among models or the availability of a new instrument. An AI Scientist must be able to identify all of these.

Question generation is not equivalent to producing interrogative sentences. A useful scientific question should concern an uncertainty that is not already resolved; be answerable in principle by observation, experiment or proof; connect to a coherent body of background knowledge; have consequences for other beliefs or applications; be sufficiently specific to guide investigation; and be sufficiently important to justify the required resources.

A system can generate questions by locating low-confidence regions of a knowledge graph, contradictions between publications, systematic model discrepancies, unexplained clusters in data, or measurements that existing theories predict poorly. It can also work backwards from human goals, such as curing a disease or reducing energy use, to identify the scientific uncertainties whose resolution would make progress possible.

Question importance is partly epistemic and partly social. An AI system can estimate how resolving a question would reduce scientific uncertainty, connect fields or enable further experiments. It cannot autonomously determine which human needs should receive priority. High-level objectives must remain subject to the governance discussed in Section 13.

### 7.2. Abduction

Logic provides a rich structure for describing scientific hypotheses. Two principal forms of logical inference are involved in hypothesis formation: abduction and induction. Abduction is the inference of a hypothesis that, if true, would explain an observation. It is often described as inference to the best explanation (Peirce, 1931–1958). Abduction is used when debugging programs, diagnosing diseases, identifying the causes of experimental anomalies and solving murders in detective fiction.

Much hypothesis formation in biomedicine is abductive. Examples include hypothesising that a previously uncharacterised gene has a particular function, that a microorganism causes a disease, or that the inhibition of a molecular pathway explains a drug’s observed effect (King et al., 2004, 2009).

Abductive reasoning is not sound in the deductive sense: the existence of an explanation does not establish that the explanation is true. Multiple hypotheses may explain the same observation. Yet abduction provides a systematic means of generating hypotheses that may subsequently be tested.

Most computational work on abduction for science has used logic as its formal language. To make abduction operational, the concept of explanation is generally related to deductive entailment. Therefore, what is required is to find a consistent ground Hypothesis that together a Theory entail an Observation (i.e. $\text{Theory} \wedge \text{Hypothesis} \vDash \text{Observation}$) (Flach & Kakas, 2000)

Additional criteria are then needed to choose among alternative explanations. These may include consistency with background knowledge, prior probability, simplicity, explanatory coverage, causal plausibility and the cost of the experiments required to discriminate among them (Flach & Kakas, 2000).

Abduction is especially valuable when observations are surprising. An AI Scientist should compare an observation with the predictive distributions of its current models. A large, structured discrepancy should trigger the generation of explanations for the discrepancy rather than immediate rejection of the data or arbitrary enlargement of error bars.

The best explanation need not be the most linguistically plausible. Candidate explanations should be ranked according to prior probability, explanatory coverage, causal coherence, simplicity, compatibility with background knowledge and the expected value of experiments that distinguish them.

## 7.3. Induction and relational learning

A key part of science is reasoning from specific observations to general hypotheses. This is inductive inference. Machine learning is, in essence, automated induction.

Most conventional machine-learning methods use a propositional representation of data. Each example is represented by a fixed tuple or vector of attributes. These attributes are descriptors or features believed to be relevant to the target concept. Strings, arrays, images and graphs may ultimately be encoded as numerical collections of attributes.

Machine-learning approaches vary partly according to the target language in which their hypotheses are expressed. Linear models express hypotheses as weighted mathematical functions. Decision trees express nested propositional conditions. Bayesian methods express probabilistic structures. Deep neural networks express hypotheses as networks of numerical transformations. Logical machine-learning systems express hypotheses as rules or logical formulae.

The philosophical problem of induction has been recognised for centuries. Hume argued that repeated observations cannot deductively establish a universal law (Hume, 1748). Russell used the familiar expectation that the future will resemble the past to show that inductive practice cannot be justified without assumptions that are themselves inductive (Russell, 1912). Popper therefore treated induction as conjecture followed by attempted refutation rather than as a logic of confirmation (Popper, 1959), while Putnam showed that formal confirmation systems inevitably depend on representational and methodological choices (Putnam, 1963).

The enormous practical success of machine learning has not eliminated the philosophical problem of induction. It has shown instead that inductive procedures can generate models with substantial predictive power. Their conclusions remain defeasible: new observations may require a hypothesis to be revised or rejected. This is why induction must be integrated with probabilities, experimental testing and rational belief revision.

### 7.3.1. First-order logic and inductive logic programming

First-order predicate logic is substantially richer and more general than propositional logic. It was developed to formalise mathematics and can express entities, properties, relations, variables and quantification. It therefore permits the expression of general scientific statements that cannot be represented naturally as fixed tuples of attributes.

Relational learning and inductive logic programming use first-order logic or logic programs to represent data, background knowledge and learned hypotheses (Muggleton, 1991; De Raedt, 2008). Their principal advantage is that they can reason directly about structured and relational objects and can incorporate extensive background knowledge into learning.

For example, a biomedical learning system can represent relations among genes, proteins, pathways, compounds, phenotypes, organisms and experiments rather than treating each case as an isolated feature vector. Background biological knowledge can constrain the hypothesis space and enable generalisation from comparatively small numbers of examples (Orhobor et al., 2020).

The greater expressive power of first-order logic makes the learning problem more difficult because the hypothesis space is vastly larger than that of propositional learning. This creates the need for language biases, type constraints, background knowledge, efficient search and the invention of useful intermediate predicates.

Human scientists routinely reason and learn implicitly using first-order and higher-order relations. They reason about classes of entities, relations between entities, mechanisms, nested structures and general laws. Future AI Scientists will require the same capability. First-order logic has been used extensively in deductive AI systems, but much less commonly in modern machine learning.

Relational learning is important because scientific hypotheses commonly concern mechanisms and structures rather than isolated attributes. A useful hypothesis may state that a class of proteins with a particular structural relation regulates another class under specified conditions. Such a statement cannot be represented naturally as a flat vector without losing its scientific form.

### 7.4. Concept and predicate invention

Predicate invention is the creation of new concepts or relations that are not present in the initial language of a learning problem. It is one of the central problems of inductive logic programming and of machine scientific discovery.

A system may be unable to express a simple theory using its original vocabulary, but able to do so once an appropriate intermediate concept has been invented. In science, many major discoveries involve precisely this process: the introduction of new entities, properties and relations with which observations can be described more simply and explained more deeply.

Traditional approaches to predicate invention have been computationally difficult and have often generated intermediate predicates that lack clear semantic interpretations. LLMs substantially improve the prospects for predicate invention because their knowledge of natural language and scientific discourse enables them to propose meaningful names, definitions and relations for latent concepts identified in data. They do not, however, solve the problem by themselves. Proposed predicates must be tested, refined and verified using inductive logic programming, deductive inference and ultimately experiment.

The most powerful architecture is therefore neuro-symbolic. An LLM proposes candidate predicates and rules; a logic-programming or theorem-proving system determines whether they are syntactically valid, consistent and sufficient to explain the examples; and the resulting proof failures or counterexamples are returned to the LLM for revision. Recent work has demonstrated precisely this combination of LLM-based abductive generation with Prolog-based deductive verification, showing that formal verification substantially improves the success of invented predicates (Yu et al., 2026).

The claim that LLMs solve predicate invention should be understood as a research programme rather than a completed result. LLMs are exceptionally capable of proposing names and definitions, but a proposed concept becomes scientifically useful only when it compresses observations, supports correct predictions, participates in explanations and survives experimental testing. Formal verification can establish consistency relative to stated premises; it cannot establish that the new predicate corresponds to a real regularity in nature.

### 7.5. Analogy and transfer

Many important scientific hypotheses are formed by analogy. A mechanism known in one domain is proposed in another, a mathematical structure used in one field is transferred to a different phenomenon, or two apparently different systems are recognised as instances of the same abstract pattern. Computational accounts emphasise the mapping of relational structure rather than superficial attributes, and the systematic transfer of connected relations from a source to a target domain (Gentner, 1983; Holyoak & Thagard, 1995).

Computational analogy requires representations that separate relational structure from surface vocabulary. An AI Scientist should identify mappings among entities and relations, assess which parts of the source mechanism can be transferred and determine where the analogy breaks down. The strongest analogies generate new testable consequences rather than merely rhetorical similarity.

Foundation models are promising for analogy because they learn broad representations across many disciplines. Their breadth also creates a danger: they can generate superficially attractive connections among almost any concepts. Candidate analogies must therefore be constrained by formal structure, causal plausibility and prospective tests.

### 7.6. Literature-based discovery

The scientific literature can contain knowledge that has never been stated explicitly in any one paper. A connection may be distributed across two or more bodies of research that use different terminology and are read by different scientific communities.

Swanson's work on literature-based discovery provided the classic example. One body of literature connected Raynaud's syndrome to increased blood viscosity and platelet aggregation, while another connected fish oil to reduced blood viscosity and platelet aggregation. The two literatures had not been connected directly, but their combination suggested the hypothesis that fish oil could benefit patients with Raynaud's syndrome (Swanson, 1986).

This form of discovery is particularly suitable for machines because it requires searching and integrating more literature than any human can read. Modern systems can identify connecting entities, pathways, mechanisms or mathematical structures across large collections.

Literature-based discovery must nevertheless be distinguished from unrestricted association. Large knowledge networks contain enormous numbers of possible paths between concepts, most of which are scientifically meaningless.

The literature can generate candidate connections, but experimental evidence is required to establish whether they are correct.

Literature-based discovery should operate over claims and evidence rather than word co-occurrence alone. A path through a knowledge graph is scientifically valuable only when its edges correspond to supported relations and when the composed path yields a coherent mechanism or prediction. The system should record whether the evidence comes from independent experiments, repeated analysis of the same dataset or speculative discussion.

### 7.7. Mechanistic and causal hypotheses

A predictive association is not necessarily a scientific explanation. Hypothesis generation should distinguish: descriptive hypotheses that summarise an observed regularity; predictive hypotheses that forecast unobserved outcomes; causal hypotheses that specify effects of intervention; mechanistic hypotheses that identify entities and processes producing the effect; and unifying hypotheses that connect previously separate laws or domains.

Mechanistic hypotheses are difficult because mechanisms may involve unobserved entities and multiple levels of organisation. An AI Scientist should use causal models, conservation laws, spatial and temporal constraints, known biochemical or physical capabilities and intervention data to restrict the space of plausible mechanisms.

A proposed mechanism should generate observations beyond those used to construct it. It should explain why the regularity occurs, identify conditions under which it fails and support counterfactual predictions. Where several mechanisms are observationally equivalent, the system should design interventions that separate them.

### 7.8. LLMs and multi-agent hypothesis generation

LLMs contain extensive implicit knowledge of scientific language and can generate hypotheses rapidly. Multi-agent systems can divide the process among generators, critics, literature checkers, methodologists and experimental designers. Co-Scientist and Robin provide early demonstrations of iterative hypothesis generation, criticism and updating (Gottweis et al., 2026; Ghareeb et al., 2026).

The principal strength of LLMs is breadth. They can combine distant literatures, translate between disciplinary vocabularies and propose mechanisms expressed in fluent scientific language. Their principal weakness is that linguistic plausibility is only weakly connected to truth. A model may restate a known idea as novel, generate a hypothesis contradicted by primary evidence or invent a mechanism with no physical meaning.

A reliable architecture should therefore separate generation from acceptance. Generative systems should be encouraged to produce diverse candidates. Independent modules should then: retrieve the nearest prior work; check whether the idea is genuinely novel; translate the proposal into a formal or executable representation; test logical and dimensional consistency; estimate prior plausibility; derive predicted observations; identify decisive experiments; and reject candidates that cannot be distinguished empirically.

Diversity should be designed deliberately. Multiple models trained on similar literature may reproduce the same consensus and the same errors. Independent representations, priors, datasets and search procedures should be used to create genuinely different hypotheses.

### 7.9. Evaluating candidate hypotheses

Many generated hypotheses are worthless, so selection is as important as generation. A candidate should first be compared with the literature and the system’s accessible training sources to establish whether it is genuinely novel. It should then be tested for consistency with well-supported background knowledge, explanatory coverage, predictive specificity, causal and mechanistic plausibility, simplicity within the chosen representation language, and robustness to reasonable alternative assumptions.

A scientifically valuable hypothesis must also be experimentally discriminable. The system should identify observations that would differ under competing explanations and assess whether those observations can be obtained ethically, safely and at proportionate cost. Scientific importance and practical feasibility are distinct from probability: a conservative hypothesis may be likely but unimportant, while a radical hypothesis may have a low prior probability but transformative consequences. A portfolio should therefore preserve both high-probability incremental work and lower-probability, high-value possibilities.

Novelty must be evaluated prospectively. It is not enough to reproduce a famous discovery present in training data. The strongest evaluation records the complete machine contribution to a genuinely unresolved problem and then subjects the resulting hypothesis to independent experiment or proof.

### 7.10. From hypothesis to research programme

A major hypothesis is rarely tested by one experiment. It generates a research programme: a sequence of measurements, controls, replications and model revisions. An AI Scientist should decompose a broad hypothesis into subsidiary claims and determine which dependencies should be tested first.

The system should update the programme after each result. A negative outcome may refute one component while leaving the broader mechanism viable. An anomalous observation may motivate a new concept. A successful prediction may increase confidence but also reveal that competing hypotheses make the same prediction.

Hypothesis generation is therefore not a single creative event. It is an iterative process in which representation, deduction, experiment and belief revision continually reshape the space of possible explanations.

## 8. Deduction, Simulation and Prediction

A hypothesis becomes scientific only when it has consequences that can be tested. Deduction derives those consequences. Simulation extends deduction to systems too complex for analytical solution. Neither establishes truth; both tell us what would follow if the model were right.

The deductive consequences of simple theories may be derived analytically. In complicated systems, the consequences are too difficult to derive in closed mathematical form. The theory must instead be represented as an executable model and its consequences calculated through simulation.

Simulation is therefore one of the principal forms of conditional deduction used in modern science. Computational simulations model systems ranging from the large-scale structure of the Universe and the Earth’s climate to biological cells, chemical reactions and elementary particles. The most abstract simulations are thought experiments, in which a simplified model is executed mentally.

A simulation does not establish that its output is true of the world. It establishes that the output follows, approximately, from the model, its parameters, initial and boundary conditions and numerical method. If any of these assumptions is incorrect, the simulation may be internally valid while making false predictions about nature.

An AI Scientist must therefore distinguish between the logical correctness of the deduction, the numerical correctness of the simulation, the empirical adequacy of the scientific model, and the applicability of the model to the situation under investigation.

### 8.1. Scientific models as executable hypotheses

A scientific model is a formal representation of selected aspects of the world. It specifies entities, states, relations and rules governing how the represented system behaves. A model necessarily abstracts away from much of the complexity of the real system.

The most widely used language for quantitative physical models is calculus, because differential equations express how systems change continuously through space and time. Newtonian mechanics, electromagnetism, fluid dynamics, quantum mechanics and general relativity are all formulated principally through differential equations.

Different scientific domains use representations suited to their available knowledge and computational constraints. In systems biology, for example, detailed differential-equation models are often impractical because many reaction mechanisms and kinetic constants are unknown. Boolean networks can represent qualitative regulatory relationships, while flux-balance analysis uses linear constraints and optimisation to predict feasible metabolic fluxes without requiring complete enzyme kinetics (Kauffman, 1969; Orth et al., 2010).

Stochastic models are required when randomness is scientifically important. Gillespie's stochastic simulation algorithm, for example, generates statistically correct trajectories for chemically reacting systems in which the numbers of molecules are small and deterministic concentration equations are inadequate (Gillespie, 1977).

Agent-based models represent systems as collections of interacting individuals or components. They are useful when population-level behaviour emerges from heterogeneous local interactions, as in ecology, epidemiology, economics and social systems (Bonabeau, 2002).

A model is an executable hypothesis when it generates predictions that can be compared with observations. The same model may contain both established scientific knowledge and conjectural assumptions. Future AI Scientists must be able to identify which components are well supported, which are uncertain and which observations could discriminate among alternative model structures.

### 8.2. Forward simulation and prediction

Forward simulation asks what the model predicts before nature answers.

A forward simulation calculates the observations predicted by a model. Let

$$y = S(M, \theta, x, u)$$

denote a simulator $S$, where $M$ is the model structure, $\theta$ its parameters, $x$ the initial and boundary conditions, $u$ an intervention or experimental input, and $y$ the predicted observations.

Forward simulations answer questions of the form:

If this model, these parameters and these conditions are correct, what should be observed?

In some cases, the model itself is under test. In others, it is provisionally assumed to be correct and the simulation is used to predict useful quantities, such as tomorrow's weather, the behaviour of a candidate material, the trajectory of a spacecraft or the response of a biological system to an intervention.

Simulations also permit counterfactual reasoning. Once a model has been specified, the scientist can ask what would happen under conditions that have not yet occurred or cannot easily be produced experimentally. This makes simulation valuable for experimental planning, engineering design and risk assessment.

The output of a deterministic simulation is often presented as a single trajectory or number. This can be misleading because the inputs and model structure are uncertain. An AI Scientist should normally propagate uncertainty through the simulator and produce a distribution over possible outcomes rather than a single apparently exact prediction.

### 8.3. Inverse problems and simulation-based inference

The inverse problem runs in the opposite direction. Given observations $y$, the objective is to infer the model, parameters, initial conditions or interventions that could have generated them:

$$y \mapsto P(M, \theta, x, u \mid y)$$

Inverse problems occur throughout science. Examples include inferring a molecular structure from a spectrum, reconstructing the interior of an object from tomographic measurements, estimating cosmological parameters from astronomical observations and identifying the parameters of a biological model from experimental time series.

For simple models, the likelihood $P(y \mid M, \theta)$ can be calculated analytically. For many realistic simulations, however, it is possible to generate observations from the model but not to evaluate their likelihood. The simulator is then an implicit generative model.

Simulation-Based Inference (SBI), also termed likelihood-free inference, addresses this problem by learning from repeated simulations. Approximate Bayesian Computation accepts parameter values when simulated data resemble the observations. More recent methods use neural density estimation, classifiers or ratio estimation to approximate posterior distributions or likelihood ratios (Cranmer et al., 2020).

SBI is important for AI Scientists because many of the most realistic scientific models are easier to execute than to invert. An AI Scientist can generate large numbers of simulated worlds under different hypotheses and learn which hypotheses are most compatible with the observations.

But inference is reliable only when the simulator includes the mechanisms capable of generating the real observations. If an important process is absent, the inferred parameters may compensate for the missing mechanism and acquire misleading values. Simulation-based inference does not remove model error; it makes inference conditional on a simulator whose adequacy must still be tested.

### 8.4. Verification, validation and calibration

Scientific simulation requires three distinct activities: verification, validation and calibration.

Verification asks whether the equations or rules of the model have been implemented and solved correctly. It is commonly expressed as: Are we solving the equations right? Verification includes checking software correctness, numerical convergence, unit consistency, conservation properties and the effect of discretisation. A simulation may fail verification because of a programming error, an unstable algorithm or a numerical approximation that is too coarse.

Validation asks whether the model represents the real system sufficiently well for its intended use: Are we solving the right equations? Validation requires comparison with independent experimental or observational data. A model may be numerically perfect but empirically inadequate because it omits an important process or applies only outside the regime being studied.

Calibration estimates uncertain model parameters from observations. A model may contain reaction rates, material constants or initial conditions whose values are not known precisely. Calibration adjusts these values so that the model better matches relevant data. Calibration should not be confused with validation. A flexible model can often be calibrated to fit a dataset even when its underlying structure is incorrect. Validation must therefore use data that were not simply absorbed during calibration.

Kennedy and O'Hagan's Bayesian framework distinguishes uncertainty in model parameters from model discrepancy: the systematic difference between the best possible output of the model and the real system (Kennedy & O'Hagan, 2001). Without an explicit discrepancy term, parameter estimates may absorb model inadequacy and appear more precise than the evidence warrants.

AI Scientists should represent at least four forms of uncertainty: uncertainty in model structure, uncertainty in parameter values, uncertainty in initial and boundary conditions, and uncertainty arising from measurements and numerical approximation.

They should also perform sensitivity analysis to determine which assumptions and parameters have the greatest influence on the predicted conclusion. A prediction that depends critically on an uncertain parameter is less robust than one that remains stable across the plausible parameter range.

## 8.5. Differentiable programming

An important development in scientific simulation is differentiable programming. In differentiable programming, the operations performed by a numerical program are organised so that derivatives of its outputs with respect to its inputs or parameters can be calculated automatically (Baydin et al., 2018; Innes et al., 2019).

Automatic differentiation is different from symbolic differentiation and finite differences. Symbolic differentiation manipulates mathematical expressions. Finite differences estimate a derivative by repeatedly evaluating a function at nearby points. Automatic differentiation applies the chain rule systematically to the elementary operations executed by the program and calculates derivatives accurately to numerical precision.

If a simulator produces a prediction

$$y = S(\theta)$$

and a loss function $L(y, y_{\text{obs}})$ measures disagreement with observations, automatic differentiation provides

$$\frac{\partial L}{\partial \theta}$$

These gradients indicate how the parameters should change to make the simulation agree more closely with the data. They can be used for parameter estimation, optimal control, inverse design and experimental optimisation.

Differentiable simulators make it possible to optimise through the complete chain from scientific parameters to predicted observations. For example, an AI Scientist could differentiate through a model of a chemical process to identify reaction conditions expected to produce a desired outcome, or through an optical simulator to design a structure with specified properties.

Systems such as JAX and Zygote have extended automatic differentiation from neural networks to general numerical programs. This allows differential-equation solvers, linear-algebra routines, probabilistic models and other scientific software to participate in gradient-based learning.

Differentiability is not automatic for every scientific process. Simulations may contain discrete events, discontinuities, changing topologies, stochastic choices or chaotic dynamics. Reverse-mode differentiation can also require large amounts of memory because intermediate program states must be retained or reconstructed.

A gradient is not a causal explanation. It describes how the output of the specified simulator changes under a local mathematical perturbation. If the model is incorrect or the parameter does not correspond to a physically realisable intervention, the gradient may have little scientific meaning.

## 8.6. Hybrid scientific models

Purely mechanistic models and purely data-driven models have complementary strengths. Mechanistic models encode established knowledge, conservation laws and causal assumptions, but may be incomplete or computationally expensive. Machine-learning models are flexible and can approximate unmodelled relationships, but may require large datasets and behave unpredictably outside their training distribution.

Hybrid models combine the two approaches. A known physical or biological model supplies the principal structure, while a learned component represents processes that are unknown, unresolved or too expensive to calculate directly.

Universal Differential Equations provide one general formulation of this idea. A differential-equation model is augmented with trainable functions, which may be neural networks, symbolic expressions or other parameterised components (Rackauckas et al., 2020). The known part of the model imposes scientific structure, while the learned component captures missing dynamics.

For example, if

$$\frac{dx}{dt} = f_{\mathrm{known}}(x, \theta) + f_{\mathrm{unknown}}(x)$$

then $f_{\mathrm{known}}$ represents the established theory and $f_{\mathrm{unknown}}$ can be learned from observations. The resulting model remains connected to scientific concepts while using data to correct or complete the theory.

Hybrid models are especially valuable when a full first-principles simulation is infeasible but a purely data-driven model would ignore too much established knowledge. They also provide a natural route towards model discovery: the learned component can subsequently be analysed and approximated by an explicit mathematical expression.

## 8.7. Surrogate models and emulation

Many scientific simulations are computationally expensive. A high-resolution simulation of climate, turbulence, molecular dynamics or a complex engineering system may require hours, days or months of computation. This makes it impractical to run the simulator millions of times for optimisation, uncertainty analysis or experimental design.

A surrogate model is a cheaper approximation to an expensive simulator:

$$\hat{S}(M, \theta, x, u) \approx S(M, \theta, x, u)$$

The surrogate is trained on input–output examples generated by the original simulation. Once trained, it can make predictions much more rapidly.

Classical surrogate methods include response surfaces, splines, reduced-order models and Gaussian processes. Modern approaches use neural networks, graph networks and operator-learning methods.

Neural operators learn mappings between functions rather than only mappings between fixed-dimensional vectors. They can learn, for example, the operator mapping an initial condition or spatially varying coefficient to the solution of a differential equation. DeepONet and the Fourier Neural Operator demonstrated that a single learned model could approximate families of differential-equation solutions over different initial conditions and parameter settings (Lu et al., 2021; Li et al., 2021).

Surrogates are valuable to AI Scientists because they allow rapid exploration of large spaces of models, parameters and experimental conditions. They can support uncertainty propagation; sensitivity analysis; inverse problems; Bayesian optimisation; experimental design; real-time control; and the comparison of large numbers of hypotheses.

But a surrogate inherits limitations from the simulations used to train it and introduces additional approximation error. It may interpolate accurately within the training domain but fail under novel conditions. Its uncertainty should therefore include both the uncertainty of the original simulator and uncertainty arising from the approximation.

A surrogate should not silently replace the original simulator. An AI Scientist should identify when a prediction lies outside the training distribution and selectively call the full simulator to check or improve the surrogate. This creates a multi-fidelity architecture in which cheap approximations are used for broad search and expensive high-fidelity simulations are used for confirmation.

### 8.8. Multiscale and multiphysics simulation

Many scientific systems involve processes operating at different spatial and temporal scales. A biological organism, for example, depends on molecular interactions, cellular dynamics, tissue organisation and whole-organism physiology. A climate model integrates processes ranging from microscopic cloud formation to global atmospheric circulation.

No single modelling formalism is optimal at every scale. Future AI Scientists will therefore need to construct and coordinate multiscale and multiphysics simulations. These may connect quantum calculations to molecular dynamics, molecular dynamics to continuum mechanics, or cellular models to tissue-level descriptions.

The interfaces between models are often more difficult than the models themselves. Quantities represented explicitly at one scale must be translated into effective parameters at another. Errors and uncertainty must be propagated across these interfaces.

AI can help identify reduced representations, learn closure relationships and decide adaptively where high resolution is required. But learned couplings must respect conservation laws and other scientific constraints. A highly accurate local approximation may still produce physically impossible global behaviour when embedded in a longer simulation.

The phrase ‘simulation intelligence’ has been proposed for the integration of scientific simulation with machine learning, probabilistic programming, causal inference, optimisation and automated programming (Lavin et al., 2021). This integration is important because no single method is sufficient for autonomous scientific reasoning.

### 8.9. Digital twins

A digital twin is a computational model of a particular physical system that is repeatedly updated using observations from that system. Whereas a conventional model may represent a general class of objects, a digital twin seeks to represent a specific machine, patient, laboratory or environment.

The cycle is:

- Collect observations from the physical system;
- Update the state and parameters of the model;
- Simulate its future behaviour;
- Identify useful or risky interventions;
- Apply or recommend an intervention;
- Assimilate the resulting observations.

Digital twins are therefore closely related to AI Scientists. Both require the integration of models, observations, uncertainty, intervention and iterative belief revision.

A digital twin should not be mistaken for a perfect digital copy. It is a model whose adequacy depends on the available measurements, assumptions and intended use. Different decisions may require different levels of fidelity. A model sufficient for monitoring may be inadequate for causal intervention.

The construction of reliable digital twins remains largely bespoke. Scaling them requires standard representations, interoperable software, uncertainty quantification and systematic methods for combining mechanistic models with streaming data (Niederer et al., 2021).

### 8.10. Simulation within an AI Scientist

Future AI Scientists will incorporate simulation throughout the scientific loop. A candidate hypothesis must be translated into an executable model whose predictions can be derived under relevant experimental conditions and compared with those of competing hypotheses. The system should identify the assumptions and parameters responsible for divergent outcomes, propagate uncertainty, and select experiments expected to distinguish among the alternatives.

After an experiment, the AI Scientist should compare the observations with the simulated predictive distributions, update the probabilities of the competing hypotheses, identify systematic model discrepancies and revise the model where necessary. It should normally maintain several models rather than commit prematurely to one: model averaging may support prediction, while discriminating experiments can be selected specifically to expose structural differences.

Every simulation should be preserved as a reproducible computational object. The record should contain the equations or source code, parameter values, initial and boundary conditions, software and library versions, numerical algorithms and tolerances, random-number seeds, relevant hardware, input datasets and the outputs used in the scientific conclusion.

Simulation also provides a comparatively safe and inexpensive environment in which proposed experiments can be checked before physical execution. It can reject impossible protocols, estimate expected outcomes and identify informative measurements. It cannot, however, reveal phenomena absent from the model. AI Scientists must therefore use simulation and physical experimentation together: simulation to derive and compare consequences, and experiment to determine which represented world most closely corresponds to reality.

## 9. Observation, Measurement and Experimentation

Nature is the final judge of a scientific hypothesis. Observation records what nature does. Experiment intervenes and asks nature a controlled question. AI Scientists must do both, and they must know exactly what was measured, how it was measured and what could have gone wrong.

A general AI Scientist must therefore support both observational and experimental science. It must decide what to observe, how to measure it, which interventions are possible and ethical, and how the resulting data relate to the theoretical quantities in its models.

### 9.1. Observation, field science and natural experiments

An observation is not a passive copy of the world. It is produced through choices about where, when and how to measure, which instruments to use, how to sample and which signals to record. Observational design can therefore be as important as experimental design.

In astronomy, an AI Scientist may schedule telescope time, choose wavelengths, coordinate observations across facilities and adapt its strategy after detecting a transient event. In ecology, it may combine field sensors, satellite imagery, acoustic monitoring and autonomous vehicles. In epidemiology, it may exploit natural experiments created by policy changes or environmental variation. In palaeontology, it may select excavation sites and integrate incomplete physical traces with geological context.

Observational systems must address selection effects. A telescope detects only objects above its sensitivity threshold. A clinical database contains patients who entered a healthcare system. A field sensor records only the locations in which it was deployed. The observation process must therefore be represented explicitly. Without a model of what could have been observed but was not, the AI Scientist may mistake measurement bias for a property of nature.

Some observations are opportunistic. Rare astronomical events, natural disasters and outbreaks may occur without warning. AI systems can monitor data streams continuously and trigger rapid follow-up. This is an important advantage over human-only science, provided that automated triggers are calibrated and preserve sufficient context for later interpretation.

Field and observational science also require logistics, robotics and safety. Autonomous vehicles may collect samples in oceans, polar regions, mines or space. The scientific planner must consider navigation, energy, communications, weather and the possibility that a failed action cannot be repaired.

### 9.2. Measurement, calibration and traceability

Scientific theories concern quantities, entities and relations; instruments produce signals. Measurement connects the two. An AI Scientist must represent how an instrument output depends on the underlying quantity of interest, the sample, the environment and the instrument state.

A measurement model may be written schematically as:

$$z = g(q, s, e) + \varepsilon$$

Here, $z$ denotes the observed signal, $q$ the true quantity, $s$ the instrument state, $e$ the environment, and $\varepsilon$ the error term. Measurement error is not necessarily independent noise. Instruments drift, saturate, possess detection limits and respond to interfering substances. Samples may degrade. Image-analysis pipelines may introduce systematic segmentation errors. A result can therefore be precise but inaccurate.

Calibration relates instrument output to reference standards. Traceability connects a measurement through an unbroken chain of calibrations to recognised standards, with uncertainty assigned at each stage. Future AI Scientists should treat calibration records, reference materials and quality-control measurements as part of the evidence, not as laboratory administration.

A complete measurement record should include: The identity and configuration of the instrument; Calibration status and reference standards; Detection and quantification limits; Sampling time and location; Environmental conditions; Sample preparation and chain of custody; Raw signals and transformation steps; Measurement uncertainty; and Known interferences and failure modes. The AI Scientist should select measurements according to the hypotheses being tested. A convenient measurement may be only weakly related to the theoretical quantity. Biomarkers, proxy variables and surrogate endpoints require explicit justification. The system should represent the inferential chain from physical signal to scientific conclusion.

Repeated measurement can reduce random error but not systematic bias. Independent measurement using different instruments or physical principles is therefore especially valuable. Agreement among methods with distinct failure modes provides stronger evidence than repeated agreement from one automated pipeline.

## 9.3. Experimental design

The field of experimental design is concerned with constructing experiments whose results support correct and sufficiently precise conclusions (Fisher, 1935; Cochran & Cox, 1957). A scientifically useful experiment must distinguish the effect of the variable under investigation from measurement error, biological variation and uncontrolled differences between experimental conditions.

An experimental design specifies: the experimental units; the interventions or treatments; the measured outcomes; positive, negative and procedural controls; the number and type of replicates; the allocation of experimental units to conditions; the treatment of nuisance variables; the statistical analysis; and the criteria by which the result will be interpreted.

The classic example is the testing of a new plant fertiliser. If treated plants are grown in one type of soil and untreated plants in another, the effect of the fertiliser cannot be separated from the effect of the soil. Soil composition is a confounding variable. A valid experiment must either hold it constant or vary it in a controlled manner that permits its effect to be estimated.

Controls are fundamental. A negative control establishes the background response expected in the absence of the intervention. A positive control establishes that the experimental system can detect an effect. A procedural or sham control distinguishes the effect of the intervention from the effect of performing the procedure itself. The appropriate controls depend on the causal structure of the experiment and cannot be inferred solely from the statistical form of the data.

Randomisation is commonly used to prevent systematic bias in the allocation of experimental units. But unrestricted randomisation is not always the most efficient design. Blocking can ensure that known sources of variation are distributed appropriately across conditions. Latin-square designs can control simultaneously for two nuisance variables. Factorial designs allow several factors and their interactions to be investigated within the same experiment.

Replication serves several purposes. Technical replicates estimate variation introduced by measurement and handling. Biological or independently produced replicates estimate variation among biological systems or experimental preparations. Merely repeating the measurement of the same sample does not establish that the result generalises to independently generated samples.

Blinding prevents expectations about the experimental condition from influencing sample handling, measurement or interpretation. Automated experiments can reduce some forms of operator bias, but they do not eliminate bias in the selection of samples, the configuration of instruments, the definition of endpoints or the analysis of results.

Statistical power is the probability that an experiment will detect an effect of a specified size when that effect is present. An experiment with insufficient replication may be incapable of discriminating among plausible hypotheses, regardless of how accurately it is executed. Conversely, unnecessarily large experiments waste time, material and money. AI Scientists should therefore incorporate power analysis and expected precision into experimental planning.

Classical experimental design generally assumes that the complete design is fixed before the observations are made. Bayesian experimental design extends this framework by treating uncertain parameters and models probabilistically and selecting designs according to their expected utility (Chaloner & Verdinelli, 1995). The utility may reflect prediction accuracy, parameter precision, discrimination among models, reduction in entropy, economic value or some combination of these objectives.

For a proposed design (d), Bayesian experimental design selects the design with the greatest expected utility:

$$d^* = \underset{d}{\operatorname{argmax}} \ \mathbb{E}_{y \sim P(y|d)}[U(d, y)]$$

where $y$ denotes a possible experimental outcome and $U(d, y)$ denotes the utility of obtaining that outcome under design $d$.

This formulation is well suited to AI Scientists because it makes the objectives and assumptions of the design explicit. It also permits the costs, risks and expected scientific value of experiments to be compared within a common framework.

Adaptive designs modify later parts of an experiment in response to earlier observations. They can allocate resources more efficiently, but they require care. Repeated inspection of results, changing endpoints or stopping when a desired result appears can invalidate conventional statistical tests. An AI Scientist must record all adaptations and use inferential procedures appropriate to the adaptive design.

Automation can improve experimental design by ensuring that controls, randomisation, blocking and replication are implemented consistently. But the choice of a valid design remains a scientific reasoning problem. A perfectly randomised experiment may answer an unimportant question, measure the wrong variable or manipulate a factor that does not distinguish the hypotheses of interest.

## 9.4. Efficient experiment selection

Experimental scientists face severe limitations in time, money, equipment, personnel and materials. Scientific priority also rewards obtaining a correct result before other researchers. It is therefore not sufficient for an experiment to be valid: it should also be selected efficiently.

Sequential experimental design chooses each new experiment using the results of previous experiments. This is the natural form of experimentation for an AI Scientist. After each experimental cycle, the system updates its models or hypothesis probabilities and selects the experiment expected to provide the greatest subsequent benefit.

The most widely used algorithmic approach in contemporary self-driving laboratories is Bayesian optimisation (Shahriari et al., 2016; Frazier, 2018). Bayesian optimisation is designed for optimising an expensive function when each evaluation corresponds to a costly experiment.

A probabilistic surrogate model is fitted to the experimental observations. This model predicts both the expected value and the uncertainty of untested conditions. An acquisition function then selects the next experiment by balancing exploitation, testing conditions expected to perform well; and exploration, testing conditions about which the model is uncertain.

Common acquisition functions include expected improvement, probability of improvement, upper-confidence bounds and information-theoretic criteria. Modern extensions support noisy observations, parallel batches, multiple objectives, constraints, mixed discrete and continuous variables and experiments with different costs.

Bayesian optimisation is particularly appropriate when the objective is to find a condition that maximises or minimises a measured quantity: the yield of a chemical reaction, the activity of a catalyst, the stability of a material or the performance of a protein.

But scientific discovery is not always an optimisation problem. The aim may instead be to learn an accurate model of the whole experimental space, estimate uncertain parameters or discriminate among competing hypotheses. In these cases, active learning and Bayesian optimal experimental design are often more appropriate.

Active learning is the branch of supervised machine learning in which the learning system selects the examples from which it learns (Cohn et al., 1996; King et al., 2004). A common strategy is to select an experiment for which the current model is most uncertain. More generally, the system may choose the experiment expected to produce the greatest reduction in uncertainty over model parameters, predictions or competing hypotheses.

The distinction between Bayesian optimisation and active learning is important. Bayesian optimisation seeks principally to locate an optimum. Active learning seeks principally to improve the model. An experiment close to a predicted optimum may be useful for optimisation but provide little information about the remainder of the experimental space. Conversely, an experiment in a highly uncertain region may be scientifically informative without being expected to produce an immediately useful outcome.

Recent work illustrates the importance of this distinction. BATCHIE uses Bayesian active learning to design combination-drug screens whose purpose is to learn the response landscape rather than merely identify one optimum. In a prospective screen involving 206 drugs and 16 cell lines, it produced useful predictions after observing approximately four per cent of the possible experiments (Tosh et al., 2025).

Experimental selection can also require discrimination among different model structures rather than merely the estimation of parameters within a single model. Bayesian active-learning methods have therefore been developed to choose measurements that simultaneously improve parameter estimation and model selection, including applications in spectroscopy and biochemical pathway analysis (Nabika et al., 2024; Isenberg et al., 2024).

Reinforcement Learning (RL) is also frequently associated with autonomous experimentation (Angermueller et al., 2020). In RL, an agent interacts with an environment. At each time point it observes a state, selects an action, receives a reward and moves to a new state. Its objective is to learn a policy that maximises cumulative reward.

The mapping between RL and the scientific method is not close. It is not obvious what should be designated as the agent, state and environment. Several mappings are possible: the laboratory robotic system acting within the physical laboratory; the experiment-selection system acting within the space of possible experiments; and the model-improvement system acting within a space of scientific models.

None captures the scientific method completely. In many RL applications, actions are cheap, feedback is rapid and the reward is clearly specified. Scientific experiments may take hours, weeks or years, consume scarce materials, produce ambiguous results and have consequences that cannot be represented by a simple scalar reward.

RL can nevertheless be valuable for lower-level laboratory control. It may optimise robotic motion, scheduling, sample transport, fault recovery or the execution of complex procedures. It can also learn experiment-selection policies in repeated domains where sufficiently realistic simulations are available. But active learning provides the closest standard machine-learning analogy to the scientific method because its central concern is the selection of evidence that will improve a model.

Human scientists typically consider only one or a small number of hypotheses at a time. Machines can overcome this limitation. An AI Scientist can maintain a probability distribution over a large set of competing hypotheses and select experiments according to their expected ability to discriminate among them.

Assume a current hypothesis set $H$, a possible experiment $e$, a direct cost $c(e)$, and a set of possible outcomes $o$. An ideal expected-cost recurrence can be written as:

$$V(H) = \min_{e} \left[ c(e) + \sum_{o} P\,(o \mid e, H) V\big(H_{e,o}\big) \right]$$

with $V(H) = 0$ when the remaining hypotheses no longer need to be distinguished. Here, $H_{e,o}$ denotes the hypothesis distribution after observing outcome $o$ from experiment $e$.

This has the structure of a decision tree: experiments form internal nodes, outcomes form branches and resolved hypotheses form leaves. Constructing the globally optimal minimum-cost decision tree is computationally intractable in general (Hyafil & Rivest, 1976). Practical AI Scientists must therefore use approximations, including greedy expected-information gain, entropy reduction, value of information, Monte Carlo tree search and restricted look-ahead.

The original Robot Scientist demonstrated that intelligent experiment selection could be competitive with human selection and substantially more efficient than random or cheapest-first strategies (King et al., 2004). Contemporary systems extend this approach to larger and noisier spaces using probabilistic models, active learning and batch selection.

The cost of an experiment should not be represented only by its monetary price. A realistic utility function should include execution time; opportunity cost of occupying equipment; scarcity of samples or reagents; probability of experimental failure; safety and environmental risk; the possibility of running experiments in parallel; the value of confirming an important result rapidly; and the expected scientific importance of the knowledge produced.

An AI Scientist should therefore choose not simply the statistically most informative experiment, but the experiment with the greatest expected scientific value under the available constraints.

### 9.5. Experimental protocols

Once a human or AI Scientist has selected an experiment, the experiment must be executed. The ordered description of the actions required to perform an experiment is an experimental protocol.

The ability to repeat a published experiment is a foundation of laboratory science. But many published protocols are incomplete or ambiguous. They often resemble cooking recipes more closely than computer programs (Soldatova et al., 2008, 2014).

Consider the following instruction from a yeast-transformation protocol:

Inoculate 4 ml of liquid YPAD or 10 ml of SC and incubate with shaking overnight at 30 °C.

This instruction leaves several important questions unanswered. Does ‘inoculate’ mean using a single colony from a solid plate, a defined number of cells or a volume from a previous liquid culture? What starting density should be used? Does ‘with shaking’ mean 20 revolutions per minute or 400 revolutions per minute? Does ‘overnight’ mean 12 hours, 16 hours or 24 hours? What vessel and working volume are required, and what orbital diameter should the shaker use?

Different interpretations can produce different physiological states and therefore different experimental results. Ambiguity introduces uncontrolled variation and contributes to irreproducibility.

There is consequently a growing need to represent experimental protocols efficiently, completely and unambiguously. This can be achieved using ontologies, formal languages and executable computer programs rather than natural language alone.

EXPO was the first general ontology of scientific experiments (Soldatova & King, 2006). EXACT was the first ontology for formally representing experimental actions (Soldatova et al., 2008). EXPO and EXACT underpinned the Robot Scientist systems by making the goals, structure, materials, actions and results of experiments explicit.

EXACT2 extended the approach to capture the semantics required for biomedical-protocol reproducibility and to support the translation of protocols written in natural language into machine-processable representations (Soldatova et al., 2014).

Modern protocol systems operate at several distinct levels:

Scientific semantics. Ontologies such as EXPO, EXACT and LabOP describe what an experiment is intended to do and the scientific meaning of its components. LabOP provides an open representation for biological protocols that can support both human-readable procedures and automated execution (Bartley et al., 2023).

Executable protocol languages. Autoprotocol represents laboratory operations as structured instructions. The Chemical Description Language, XDL, provides a hardware-independent representation of chemical synthesis procedures and has been used to translate the chemical literature into automatically executable protocols (Steiner et al., 2019; Mehr et al., 2020).

Hardware abstraction. PyLabRobot provides a common, open-source Python interface for different manufacturers' liquid handlers and accessories, reducing the need to rewrite the same protocol separately for each robotic system (Wierenga et al., 2023).

Instrument communication. SiLA 2 defines a service-oriented standard through which laboratory instruments and software can expose capabilities and exchange commands and data (Juchli, 2022). OPC Unified Architecture and the Laboratory and Analytical Device Standard provide related approaches derived from industrial automation.

Workflow orchestration. Systems such as ChemOS, AlabOS and IvoryOS coordinate experimental planners, instruments, analyses and data flows. IvoryOS demonstrated a common web-based orchestration interface across six different self-driving laboratories (Zhang et al., 2025).

These layers are complementary. A device-level communication standard does not specify the scientific purpose of an experiment. A high-level ontology does not by itself provide the commands required to operate a particular instrument.

The appropriate architecture resembles a compiler. A high-level scientific intention is progressively translated into: an abstract experimental design; a formal protocol expressed through domain actions; a schedule assigning actions to available equipment; device-specific commands; monitored physical execution; and a machine-readable record of what actually occurred.

The distinction between the intended and executed protocol is important. A robot may be instructed to dispense 100 µl but actually dispense a different volume because of calibration drift, blockage or evaporation. A complete experimental record must contain both the command and the measured or inferred physical outcome.

A complete protocol representation should identify materials, concentrations and quantities; equipment and calibration status; temporal and environmental conditions; the preconditions and postconditions of each action; dependencies among actions; acceptable tolerances; safety constraints; quality-control checks; error-handling procedures; and the provenance of every protocol version. LLMs can help translate natural-language procedures into this representation, but direct generation of robot commands from prose is unsafe unless the result is checked against equipment capabilities, inventory, units, physical constraints and safety rules before execution.

Recent work demonstrates this verification loop. Autonomous Chemputer Reaction Agents extract reaction procedures from the literature, translate them into XDL, identify unsupported operations, revise the code and submit the resulting protocol for physical execution and analytical validation (Pagel et al., 2026). This provides an early example of connecting the scientific literature directly to executable and experimentally verified protocols.

Formal protocol languages do not automatically ensure reproducibility. The same abstract action may have different physical implementations on different instruments. 'Mix', for example, may mean orbital shaking, vortexing, repeated aspiration and dispensing or stirring. These operations are not scientifically equivalent in every context. Protocol abstractions must therefore state the conditions under which implementations may be substituted (Canty & Abolhasani, 2024).

Versioned protocol repositories are also important. A protocol should possess a persistent identifier, a recorded history and links to the experiments in which each version was used. Repositories such as protocols.io allow detailed protocols to be cited and revised separately from the papers that use them.

Future publications should include both a human-readable protocol and an executable formal representation. The human-readable version supports understanding and criticism. The formal version enables automated checking, execution and transfer between laboratories.

### 9.6. Experimental reproducibility

One case for AI Scientists is that the science they create is more easily reproduced than typical human science (King et al., 2009). Ever since the seventeenth-century scientific revolution, a fundamental requirement of science has been that results should be open to independent testing (Shapin & Schaffer, 1985). Yet relatively few biomedical findings are subjected to direct reproduction, and attempts to reproduce published findings frequently encounter difficulty (Ioannidis, 2005; Baker, 2016).

We distinguish repeatability, reproducibility and robustness (Roper et al., 2022). A result is repeatable when consistent measurements are obtained under essentially the same conditions: the same method, laboratory, operators, equipment and materials over a short interval. A result is reproducible when a consistent result is obtained after relevant changes in the conditions, such as a different operator, laboratory, instrument implementation or independently prepared material. A result is robust when the scientific conclusion remains valid across an explicitly relevant range of experimental conditions, biological systems or analytical choices. These distinctions matter because repeating the same automated procedure on the same machine may demonstrate precision without demonstrating that the underlying scientific claim generalises.

Many factors cause irreproducibility. The original result may not have been repeatable. The experiment may have been underpowered, poorly randomised or affected by an uncontrolled confounder. The protocol may have omitted essential details. Reagents, cell lines or environmental conditions may differ. The analysis may contain errors or unreported degrees of freedom. The published claim may overstate what the observations justify.

We believe that most published biomedical results are repeatable: laboratories can generally repeat their own central published findings. If a result is repeatable but not independently reproducible, a likely explanation is that the published protocol or experimental context did not contain sufficient information to recreate the relevant conditions.

The ambiguity of experimental protocols is therefore a major source of irreproducibility. AI Scientists can record experimental metadata at much greater semantic detail than human scientists. A machine can automatically preserve exact times, temperatures, instrument settings, liquid transfers, consumable identifiers, calibration records, environmental observations, raw data and analysis versions.

Automation can improve reproducibility because protocols are executed consistently, deviations from the intended procedure can be detected, all actions can be timestamped, metadata can be captured directly from instruments, samples and reagents can be identified digitally, analyses can be linked to raw data and executable code, and complete experiments can be replayed. These advantages turn the experimental record from a retrospective narrative into a direct trace of what the system attempted and what physically occurred. Automation does not, however, guarantee reproducibility: a robot can repeat a systematic error with exceptional precision, two instruments may implement the same abstract action differently, and a shared software defect may cause several laboratories to reproduce the same artefact. Independent reproducibility therefore requires variation in equipment, implementation and location, together with explicit preservation of the scientific semantics of the protocol.

The potential for automated reproduction has already been demonstrated. Eve was used to test claims extracted from the cancer-biology literature using two different experimental teams and breast-cancer cell lines. Significant evidence for repeatability was found for 43 statements, and evidence for reproducibility or robustness was found for 22 statements. The experiments also produced two serendipitous findings (Roper et al., 2022). This demonstrated that semi-automated testing of published biological claims is already feasible.

The most direct solution to the reproducibility crisis would be for more scientists to reproduce the results of other scientists. But there are strong sociological and career disincentives. Reproduction studies are difficult to fund, may be difficult to publish and can provoke hostility from the authors of the original work. Researchers receive greater career rewards for claiming a novel discovery than for establishing whether an existing claim is reliable.

Given the high cost and difficulty of reproducing results under the current funding and career system, human scientists will never experimentally confirm more than a small fraction of the published literature. The only feasible way to increase the proportion substantially is to automate the process.

A scalable system for automated reproduction would extract testable claims and supporting protocols from the literature, formalise the claims and experimental conditions, and rank candidate claims by importance, uncertainty and feasibility. It would translate selected protocols to available automated laboratories, execute repeated and independent experiments, analyse the evidence, update the probability assigned to each claim, and publish positive, negative and inconclusive outcomes. Such a system would transform the scientific literature from a largely static collection of claims into a continuously tested and probabilistically weighted knowledge base.

## 9.7. Novel experiments

The central task facing every experimental scientist is the invention of experiments capable of testing hypotheses. Relatively little research has focused directly on this aspect of automating science.

The invention of an experiment is different from conventional experimental design. Conventional design assumes that the intervention and measurement are already known and asks how they should be allocated, controlled and replicated. It is also different from efficient experiment selection, which assumes a predefined list or parameterised family of possible experiments.

The abstract novel-experiment problem is: Given a hypothesis, background scientific knowledge, a set of available materials and a description of laboratory capabilities, generate a valid protocol whose possible outcomes discriminate among the hypothesis and its alternatives.

This is a problem in scientific reasoning, causal inference, planning, program synthesis and robotics.

All existing AI Scientist systems are limited in the classes of experiment they can perform. Their actions are usually drawn from a small, stereotyped and human-defined vocabulary. Adam could choose among auxotrophic-growth experiments. Eve could select compounds and assay conditions. Many self-driving laboratories choose values for temperatures, concentrations, compositions or processing parameters within a predefined space.

These systems may select experiments autonomously, but humans have already specified the ontology of possible actions and measurements. An open-ended AI Scientist must be able to invent a new intervention or measurement when the existing experimental vocabulary is inadequate.

The generation of a novel experiment requires several kinds of knowledge: the causal consequences predicted by the hypothesis; the observations expected under alternative hypotheses; the physical and chemical properties of the objects involved; available instruments and their capabilities; the operations required to prepare samples; measurement uncertainty and detection limits; safety constraints; and the relationship between instrument outputs and scientific conclusions.

The AI Scientist must first derive an observable difference among the competing hypotheses. It must then identify an intervention and measurement capable of revealing that difference. Finally, it must construct a sequence of physical actions that implements the intervention without introducing confounding factors.

This process can be represented as hierarchical planning. At the highest level, the system chooses an experimental strategy. At intermediate levels, it decomposes that strategy into operations such as preparation, intervention, incubation, separation and measurement. At the lowest level, these operations are compiled into commands for specific equipment.

The plan must be inventory-aware. An experiment cannot require a reagent that is unavailable or an instrument incapable of the required precision. It must also consider the state of the laboratory: which equipment is occupied, which samples are stable, which consumables remain and which actions can safely be conducted in parallel.

LLMs provide important new capabilities because they can search instrument documentation, retrieve protocols and propose experimental procedures. Coscientist demonstrated that an LLM-based agent could plan chemical syntheses, use technical documentation, write robot-control code and conduct reaction-optimisation experiments (Boiko et al., 2023).

Such systems are early steps towards novel experimental generation, but they generally recombine operations and procedures already described in their documentation or training data. Generating a truly novel experiment may require creating a new assay, adapting equipment for an unanticipated purpose or designing new apparatus.

An important intermediate capability is experimental troubleshooting. When an experiment fails, the AI Scientist should determine whether the failure contradicts the hypothesis or instead results from the procedure, reagent, instrument or analysis. It should propose diagnostic experiments and revise the protocol. AutoLabs and related multi-agent systems demonstrate initial forms of explicit error detection and self-correction in automated chemical experimentation (Panapitiya et al., 2026).

Before executing a newly generated protocol, an AI Scientist should test it computationally. Formal verification can detect inconsistent preconditions, impossible schedules and unsafe actions. Digital twins and physical simulations can predict whether robotic movements are feasible and whether processes are likely to exceed equipment limits. These checks cannot establish that the scientific experiment will work, but they can exclude many invalid implementations.

The most advanced form of novel experimental design is the invention of new experimental technology. Great scientific advances have often depended on new ways of observing the world: the telescope, microscope, cloud chamber, particle accelerator, polymerase chain reaction and DNA sequencer. An open-ended AI Scientist must eventually be able not only to use existing instruments, but also to design and commission new ones.

## 9.8. Laboratory automation

Historically, laboratory automation was driven by the need to execute large numbers of related experiments, particularly in pharmaceutical screening, clinical analysis and manufacturing quality control. It is now a major global industry. The capabilities of laboratory automation have advanced steadily. Robots can now perform most, although not all, of the routine operations performed by humans in chemistry and biology laboratories. These include liquid handling, sample transfer, incubation, centrifugation, filtration, chromatography, microscopy, spectroscopy, cell culture, colony manipulation and parts of chemical synthesis.

Automation increases scientific productivity because robots can work more cheaply, rapidly, accurately and continuously than humans for many well-defined tasks. They do not tire, can operate throughout the night and can record their actions automatically. Robotic capacity can also be increased or reduced more easily than a comparably trained human workforce.

Most laboratory automation remains fixed and highly structured. Instruments operate inside protective enclosures or are connected by robotic arms and tracks. The layout, labware and sequence of actions must be specified carefully. Small changes in equipment or consumables may require substantial reprogramming.

Automation is not automatic (Christensen et al., 2021). Building and maintaining an integrated laboratory requires expertise in robotics, control, software, analytical science and the relevant experimental domain. Instruments from different manufacturers use incompatible interfaces. Consumables must be replenished, waste removed, equipment calibrated and samples transported. Failures that a human would correct intuitively may stop an automated workflow.

Laboratory logistics remain a major limitation. A liquid-handling robot may perform transfers automatically, while a human technician loads plates, opens packages, replaces tips, moves samples and resolves faults. Automating these apparently simple actions requires perception, dexterous manipulation and understanding of the laboratory’s physical state.

Mobile robots provide one approach. Burger et al. developed a mobile robotic chemist capable of navigating a conventional laboratory, operating instruments designed for humans and conducting a closed-loop search for improved photocatalysts (Burger et al., 2020). Mobile systems can connect instruments without requiring a bespoke fixed robotic track, although safe navigation and manipulation in shared laboratories remain challenging.

The separation between laboratory automation and general AI robotics has slowed progress. Laboratory automation has traditionally emphasised reliability and throughput in carefully engineered environments. AI robotics has emphasised perception, navigation and manipulation in less structured settings. Future laboratories will combine the reliability of industrial automation with the flexibility of modern robotics.

Interoperability is another central problem. A self-driving laboratory assembled from five instruments may require five separate software interfaces and custom code connecting them. Replacing one instrument may require rewriting the workflow.

Standards such as SiLA 2 address device communication, while open software such as PyLabRobot addresses hardware abstraction. Orchestration systems such as IvoryOS provide higher-level control across diverse self-driving laboratories. Dynamic knowledge graphs can represent the capabilities, materials and data flows of distributed laboratory systems (Bai et al., 2024).

Quality control must be integral to automation. The system should detect failed liquid transfers, unexpected spectra, contaminated samples, calibration drift and physical obstructions. Sensors and computer vision can compare the intended and actual laboratory states. Every automated action should return evidence that it occurred successfully rather than merely report that a software command was issued.

Safety is especially important when AI systems control physical equipment because a generated action may combine individually permitted operations into an unsafe sequence. Autonomous laboratories therefore require layered safety engineering. Equipment-level interlocks and collision-avoidance systems should constrain physical motion independently of the AI planner; machine-readable hazard models should restrict chemical and biological operations; abnormal states should be monitored continuously; and hazardous or irreversible actions should require explicit human authorisation. Emergency-stop mechanisms must remain independent of the AI system. Increasing intelligence does not remove the need for conventional safety engineering: even the most capable planner should be physically unable to override a safety interlock.

### 9.9. Cloud and federated laboratories

Cloud laboratories seek to address the high cost, limited accessibility and integration difficulties of laboratory automation. The analogy is with cloud computing. Instead of purchasing and maintaining all the necessary equipment, researchers access a centrally operated automated laboratory as a service.

Users specify experiments remotely through a software interface. Samples may be sent to the facility or selected from materials already held there. Robots and trained technicians execute the protocols, and the resulting data are returned electronically. Commercial examples include Emerald Cloud Lab and Strateos, while academic access has been advocated as a means of improving reproducibility and widening access to advanced equipment (Armer et al., 2023).

Cloud laboratories provide several potential advantages: access to expensive equipment without local capital expenditure; continuous operation; standardised execution; automatic association of results with the commands that produced them; easier sharing and repetition of protocols; and access for researchers at institutions without advanced facilities.

They may enable 'virtual' biotechnology or pharmaceutical companies that possess scientific and computational expertise without maintaining complete physical laboratories. They can also democratise experimental science by allowing students and researchers in poorly equipped institutions to access sophisticated instruments remotely.

Carnegie Mellon University established the first large university cloud laboratory based on the Emerald Cloud Lab platform, designed to provide remote access to a broad range of chemical and life-science equipment and to support AI-enabled experimentation, education and reproducibility (Carnegie Mellon University, 2021; Armer et al., 2023).

Cloud laboratories also have limitations. Samples and reagents must be transported, creating delays and constraints on unstable or regulated materials. Not every experimental procedure can be performed using the available equipment. Proprietary languages and interfaces may create vendor lock-in. Sensitive data, intellectual property and biological materials raise governance and security concerns.

Cloud laboratories are generally best suited to protocols that can be expressed through an established set of operations. They are less suitable for experiments requiring continuous physical improvisation, custom-built apparatus or tacit manipulation by an expert.

A further development is the federation of laboratories. No single facility can contain every instrument or form of expertise. Networks of automated laboratories could distribute experimental tasks according to capability, cost, location and availability.

A federated AI Scientist might conduct sample preparation in one laboratory, spectroscopy in another, high-resolution imaging in a third and specialised computation elsewhere. The system would need to preserve sample identity, protocol semantics, calibration, provenance and chain of custody across the complete workflow.

Dynamic knowledge graphs have been used to coordinate distributed self-driving laboratories by representing experimental actions, materials and information flows explicitly (Bai et al., 2024).

Interoperable orchestrators such as IvoryOS demonstrate that common control interfaces can be applied across heterogeneous laboratories (Zhang et al., 2025).

The greatest danger is the ‘Balkanisation’ of laboratory automation into incompatible proprietary ecosystems. Global laboratory networks will require shared standards for describing instrument capabilities, experimental actions, units and measurement uncertainty, sample and material identities, data and metadata, authentication and authorisation, safety constraints and protocol transfer. These standards must preserve scientific meaning across different hardware implementations rather than merely translate commands. The long-term objective is an experimental infrastructure analogous to the internet: a distributed network in which AI Scientists can discover laboratory capabilities, submit verified protocols, receive machine-readable results and combine evidence obtained at different sites.

Such an infrastructure would make physical experimentation available programmatically at a global scale. It would provide the experimental foundation required for teams of AI Scientists to investigate scientific problems far larger and more complex than any single human laboratory could address.

### 9.10. Edge Computing and Hardware Bottlenecks

While AI capabilities are often discussed in terms of software, the physical deployment of AI Scientists is strictly constrained by hardware bottlenecks. Advanced autonomous laboratories require real-time robotic adjustments, high-frequency sensor monitoring, and instantaneous safety interlocks. Relying exclusively on cloud computing introduces unacceptable latency for kinetic operations. Therefore, self-driving laboratories must integrate powerful edge computing—deploying models directly on local hardware to ensure deterministic, sub-millisecond response times for physical interventions. Furthermore, the global scaling of AI science faces hard physical limits regarding GPU availability, memory bandwidth, and the enormous energy requirements of continuous inference, necessitating the development of highly efficient, domain-specific hardware accelerators tailored for scientific workloads (Hennessy & Patterson, 2019).

## 10. Data Analysis, Interpretation and Belief Revision

Experiments do not provide conclusions. They provide data. The scientist must decide whether those data are reliable, what they imply and how strongly they favour one explanation over another. Without this step, a closed experimental loop is only automated measurement.

An AI Scientist must distinguish at least four objects: the physical event that occurred, the measurement produced by an instrument, the processed data used in an analysis, and the scientific claim inferred from that analysis.

These objects should be linked, but they are not identical. A software pipeline may process the wrong file correctly. A statistical test may be calculated correctly on measurements produced by a failed instrument. A result may be statistically significant but irrelevant to the hypothesis under investigation.

### 10.1. Data quality and provenance

Data-quality assessment should begin during acquisition. Every automated action should return evidence that the intended physical event occurred. A command log saying that 100 microlitres were dispensed is weaker evidence than a balance, image or flow sensor confirming the transfer.

Relevant quality dimensions include completeness; accuracy; precision; temporal consistency; internal consistency; calibration status; identity of samples and variables; compliance with the intended protocol; and absence of contamination or instrument failure.

The system should preserve raw data whenever feasible. Processed data are interpretations of raw signals and may need to be regenerated after an error or improved method is discovered. Every transformation should be recorded with software versions, parameters and code.

Provenance should extend from the scientific question to the final claim. A reader, or another AI Scientist, must be able to trace a conclusion backwards through the statistical result, processed dataset, raw measurements, executed protocol, samples and materials, instrument state and original decision to perform the experiment.

### 10.2. Preprocessing and exploratory analysis

Scientific data commonly require preprocessing: background subtraction, normalisation, alignment, segmentation, filtering, feature extraction, batch correction or conversion of instrument-specific formats. These operations can be essential, but they can also create or remove apparent effects.

An AI Scientist should represent preprocessing choices as uncertain modelling decisions rather than invisible facts. Where several reasonable pipelines exist, the system should evaluate whether the scientific conclusion is robust across them. A finding that appears only under one arbitrary normalisation method is weaker than one preserved across defensible alternatives.

Exploratory analysis is scientifically valuable. Visualisation, clustering and anomaly detection can reveal unexpected structure and generate new hypotheses. But analyses selected after inspecting the data should not be presented as if they had been specified in advance. The provenance record should distinguish exploratory from confirmatory analysis.

The system should guard against data leakage. Information from the test set, future observations or replicate groups must not enter model training inadvertently. In longitudinal and spatial data, ordinary random splitting can produce severe leakage because neighbouring observations are not independent.

### 10.3. Statistical inference and uncertainty

Statistical analysis should be determined by the scientific question and experimental design. The AI Scientist must identify the experimental unit, dependence structure, outcome distribution, nuisance variables and missing-data process before choosing a model.

Uncertainty should include more than a standard error produced by one fitted model. Relevant sources include sampling variation; measurement error; parameter uncertainty; model-structure uncertainty; preprocessing choices; uncertainty about sample identity or protocol execution; selection effects; and numerical approximation.

Bayesian models provide a coherent framework for propagating many of these uncertainties, but their results remain conditional on priors and model assumptions. Frequentist procedures can provide valuable error control under repeated sampling, but their interpretation also depends on the design and stopping rule. An AI Scientist should understand both traditions and select methods according to the inferential objective.

Multiple testing is a central problem for machine science. An AI system can examine millions of variables, transformations and subgroups. Without correction or independent confirmation, it will inevitably find apparently significant patterns. The complete search process, not merely the final selected analysis, must be considered when assessing evidence.

Effect size and uncertainty are more informative than a binary declaration of statistical significance. A very small effect can be statistically detectable but scientifically unimportant. Conversely, a potentially important effect may remain uncertain because samples are scarce. The AI Scientist should connect statistical quantities to scientific utility and mechanism.

### 10.4. Missing, censored and corrupted data

Missing data are rarely harmless. A measurement may be absent because an instrument failed, a sample was below the detection limit, a participant withdrew or an experiment was stopped. Each mechanism implies a different interpretation.

An AI Scientist should record why each value is missing and represent the missingness mechanism. It should not silently remove failed experiments or impute values without propagating uncertainty. Censored measurements, such as concentrations below a detection threshold, contain information and should not be replaced mechanically by zero.

Corrupted data may arise from sample swaps, transmission errors, duplicated records, contamination or software defects. Automated anomaly detection can identify suspicious observations, but deletion should require an explicit scientific rationale. An outlier may be an error, or it may be the first observation of a new phenomenon.

### 10.5. Distinguishing experimental failure from theoretical failure

When an observation conflicts with a prediction, several explanations are possible: the hypothesis may be wrong; an auxiliary assumption may be wrong; the experiment may have been implemented incorrectly; the instrument may have failed; the sample may be inappropriate or contaminated; the analysis may be wrong; or the observation may be a rare event under a correct model. This is the practical form of the Duhem–Quine problem: empirical tests confront networks of assumptions rather than an isolated proposition (Duhem, 1954; Quine, 1951).

An AI Scientist should diagnose these alternatives rather than automatically treating every discrepancy as falsification or every failed experiment as a technical problem. This requires explicit models of the protocol, laboratory state, measurement process and competing scientific hypotheses.

Diagnostic experiments can be designed to localise failure. Positive controls test whether the system can produce an expected response; reference materials test measurement accuracy; and repetition on different equipment can separate a scientific effect from an instrument artefact. Such severe and discriminating tests are valuable because they expose the specific ways in which a claim could be wrong rather than merely seeking nominal agreement (Mayo, 1996). The most informative next action may therefore be a quality-control experiment rather than a direct test of the main theory.

Anomalies should be preserved. A mature system should maintain an anomaly register linking unexplained observations to the models they challenge. Repeated anomalies across independent contexts may justify a new research programme.

### 10.6. Model comparison and Bayesian belief revision

The central output of scientific analysis is a change in the probability assigned to hypotheses and models. The strength of evidence depends on how probable the observation was under competing explanations. For every material result, an AI Scientist should preserve the prior probabilities, the predictive distribution under each hypothesis, the measurement model and observed data, the calculated or approximated evidence, the posterior probabilities, and the sensitivity of the conclusion to alternative priors and analyses.

Model comparison must allow for the possibility that all current models are inadequate. If each model predicts the data poorly, simply renormalising their probabilities to sum to one can conceal the need for a new hypothesis. Posterior predictive checks, anomaly records and explicit discrepancy models can reveal this open-world failure.

Belief revision should occur at the appropriate level. An observation may alter confidence in a parameter, a local mechanism, an entire theory, an experimental method or the reliability of a data source. The system should not propagate a local failure indiscriminately through the knowledge base. Where exact Bayesian computation is impossible, it should record the approximation, diagnostics and known failure modes rather than present numerical precision as epistemic certainty.

### 10.7. Sequential analysis and stopping

AI Scientists conduct sequential research. After every experiment, they may decide whether to continue, stop, replicate, change the model or investigate an anomaly.

Stopping rules affect inference. Repeatedly testing until a desired result appears inflates false-positive rates under conventional analyses. Bayesian decision theory can represent stopping more naturally, but the utility function and model must still be specified.

A research programme may stop when competing hypotheses have been distinguished sufficiently; when additional experiments have low expected information value; when a practical objective has been achieved; when risk or resource use exceeds the expected benefit; when results indicate that the problem representation should be changed; and when independent replication is required before further extension.

Stopping should not be determined only by statistical certainty. Scientific significance, cost, safety and opportunity cost also matter.

### 10.8. Negative and inconclusive results

Negative results are essential to rational science. They constrain models, prevent repeated failure and provide evidence about the limits of methods. Human publication systems often suppress them because novelty is rewarded more strongly than correction or exclusion.

AI Scientists can preserve every attempted experiment, including failures and null results. But a negative result is interpretable only when the experiment had sufficient sensitivity and quality to detect the predicted effect. ‘No significant difference’ is not equivalent to evidence of equivalence.

Inconclusive results should remain explicitly inconclusive. The system should not force every experiment into support or rejection. It should identify what additional measurement, replication or precision would be required to resolve the uncertainty.

### 10.9. Integrating evidence across experiments and laboratories

Scientific knowledge rarely depends on one experiment. Evidence must be integrated across different samples, instruments, laboratories, populations and methods.

The AI Scientist should represent dependencies among studies. Ten papers analysing the same dataset are not ten independent confirmations. Results produced by laboratories using identical software and equipment may share systematic errors. Evidence is strongest when it arises from genuinely independent methods with different failure modes.

Hierarchical models can integrate evidence while representing variation among laboratories or populations. Meta-analysis can combine effect estimates, but publication bias, selective reporting and heterogeneous methods must be modelled. Formal claim-evidence graphs can show which conclusions depend on which observations and whether support is direct or indirect.

Replication can change not only the probability of a claim but the estimated reliability of a laboratory, method or data source. These reliability estimates should themselves remain uncertain and revisable.

### 10.10. Updating the scientific knowledge system

Discovery is complete only when the knowledge system changes. The final stage of the loop is not manuscript production but the update of a versioned scientific knowledge system. A validated result should modify the probabilities assigned to hypotheses, parameter distributions, causal and mechanistic models, reliability estimates for methods and sources, unresolved-anomaly registers, priorities for future experiments, and formal claims linked to evidence and provenance. Genesis-DB illustrates a database architecture for autonomous laboratory systems, while RIMBO provides an ontology for recording model revisions and their justification (Reder et al., 2023; Kronström et al., 2023).

The previous state should not be erased. Scientific knowledge changes through time, and later users must be able to reconstruct what was believed, on the basis of which evidence, and why a model or claim was revised. Corrections, retractions and superseding models should therefore be represented as versioned changes.

This completes the scientific loop. Hypotheses generate predictions; experiments generate measurements; analysis converts measurements into evidence; and belief revision converts evidence into updated scientific knowledge. Without this final stage, laboratory automation produces data but not autonomous science.

## 11. Evaluation and Benchmarking of AI Scientists

Fluent scientific prose is not discovery. Neither is rediscovering an answer already present in a training set. AI Scientists must be judged by what they discover prospectively, how reliably they discover it and how much human scientific work remains hidden in the result. The evaluation of AI Scientists must therefore be prospective, multidimensional and grounded in independent evidence.

### 11.1. What should be evaluated?

An AI Scientist must be evaluated as a complete scientific system rather than as a collection of impressive components. Correctness, calibration, novelty, explanatory and causal depth, scientific significance, autonomy, human intervention, generality, experimental efficiency, robustness, reproducibility, communication, cost and safety should be reported separately. A single score would conceal consequential trade-offs: a system may be highly autonomous but scientifically weak, or scientifically capable but unsafe.

The evaluation should therefore produce a capability profile with explicit operating limits. Established examples show why domain-specific external tests matter: CASP evaluates protein-structure prediction prospectively against unreleased structures (Moult et al., 1995), Robot Scientist studies evaluate physical hypothesis testing and reproducibility (King et al., 2004; Roper et al., 2022), and Science-Gym evaluates agents that must choose experiments before inferring equations (Cerrato et al., 2026).

### 11.2. Retrospective reconstruction versus prospective discovery

The rediscovery of a known law, molecule or hypothesis can help decide system competence, but it does not establish scientific creativity. The target may have occurred in training data, the literature or the benchmark design. Temporally held-out evaluation is stronger, because a system is restricted to knowledge available before a specified date and asked to anticipate later discoveries, although indirect contamination remains possible.

The strongest evidence is prospective evaluation on a genuinely unresolved problem. The question, available information, system configuration and human contributions should be fixed and recorded before the result is known. The resulting claim must then be assessed by independent experiment, observation or formal proof. This is the standard required for the Nobel Turing Challenge (Kitano, 2021).

### 11.3. Retrieval, recombination and genuine novelty

A hypothesis is not novel merely because an evaluator has not encountered it. Prior-art searches should cover papers, preprints, patents, databases and the information sources available to the model. Novelty can consist of a new application, empirical relationship, mechanism, theoretical entity, mathematical law, experimental method or unification of previously separate domains.

Recombination can be creative: many discoveries join existing facts in a way no one had recognised. The relevant question is whether the system made a non-trivial causal contribution that changed what could be predicted, explained or done. Detailed provenance is therefore part of novelty evaluation, not an administrative supplement.

### 11.4. Correctness and evidential strength

Machine-generated claims must ultimately be evaluated against nature or formal proof; review by another language model is insufficient. Empirical evaluation should examine experimental validity, measurement quality, appropriateness of the analysis, independent replication, alternative explanations and whether the claimed scope matches the evidence. Mathematical proofs should be checked by trusted proof assistants where possible, and computational results should be independently rerun from preserved code and data.

Calibration is itself a scientific virtue. A cautious probability distribution that accurately represents uncertainty is preferable to an overconfident binary answer with the same average accuracy. Evaluation must therefore use proper scoring rules, reliability diagrams or equivalent domain-appropriate tests rather than reward confidence alone.

### 11.5. Autonomy and human intervention

Autonomy is the amount and scientific importance of work the machine actually performs. The level in Table 2 should be assigned from observed interventions, not marketing descriptions. The provenance record should show who selected the problem, supplied the key hypothesis, designed and compiled the experiment, resolved failures, chose the analysis, interpreted the result and formulated the scientific claim.

Human interventions must be classified by significance. Replacing a reagent bottle is not equivalent to providing the decisive conceptual insight, and a system should not receive high autonomy credit when experts invisibly repair its reasoning. Autonomy is also not intrinsically desirable: under high risk, a supervised system that acts safely may be superior to an unrestricted one with the same scientific output.

### 11.6. Generality and transfer

A system may overfit an entire scientific workflow. Generality should therefore be tested by transfer to new instruments, materials or organisms, measurement types, hypotheses, objective functions, laboratories and scientific domains. Transfer should require limited task-specific engineering; a platform that needs months of expert programming for each assay may be valuable automation without being a general AI Scientist.

Evaluation should distinguish superficial reuse from conceptual transfer. Applying the same optimiser to another parameter space is weaker evidence than recognising that a new problem requires a different representation, causal model or experimental strategy.

### 11.7. Experimental efficiency and scientific value

Self-driving laboratories are often judged by the number of experiments required to locate an optimum. AI Scientists require broader measures: reduction in uncertainty per experiment, competing hypotheses distinguished, time to a reproducible conclusion, material and energy use, equipment opportunity cost, quality-control pass rate and the value of negative results preserved. These quantities should be compared with human, random-search and algorithmic baselines (King et al., 2004; Shahriari et al., 2016).

Efficiency must cover the complete scientific cycle. A fast experiment-selection algorithm provides little benefit if protocol generation, troubleshooting or interpretation requires extensive hidden human labour. Scientific value also cannot be inferred from throughput: one deep, general explanation may contribute more than hundreds of narrow optimisations.

### 11.8. Robustness, reproducibility and fault recovery

Benchmarks should introduce calibration drift, missing data, instrument failure, contaminated samples, inconsistent literature and distribution shift. The system should detect that its assumptions no longer hold, diagnose the likely failure and recover without unsafe improvisation.

Reproducibility should be tested across independent sites and implementations. Repeating the same software and hardware configuration establishes repeatability, not full reproducibility. Strong evaluation varies instruments, operators, materials and analytical pipelines while preserving the scientific semantics of the experiment (Roper et al., 2022).

## 11.9

### Explanation and human improvement.

A machine explanation should improve human scientific performance, not merely sound fluent. Evaluation should test whether scientists can use the system's discovery to make correct predictions, design experiments or solve problems they could not solve before. This is Michie's ultra-strong criterion, demonstrated experimentally for learned symbolic programs by Muggleton et al. (2018).

Explanations must be faithful to the evidence and reasoning that produced the result. Evaluation should distinguish mechanistic explanation, mathematical derivation, visualisation and pedagogical simplification, and should test each with appropriate users rather than treating all natural-language output as explanation.

## 11.10. Safety and governance metrics

Scientific capability is unacceptable when achieved through unsafe or unauthorised action. Evaluation should test compliance with machine-readable permissions, recognition of hazardous and dual-use requests, behaviour under adversarial prompting, resistance to cyber manipulation, escalation to human review, integrity of audit logs and the ability to stop safely. These tests should draw on risk-management and responsible-innovation frameworks as well as domain-specific laboratory safety requirements (Shneiderman, 2022; Council of Canadian Academies, 2022; Bengio et al., 2026).

A system that produces a correct result by violating legal, ethical or safety constraints has failed as an AI Scientist. Safety and scientific competence must therefore be evaluated jointly.

## 11.11. Criteria for the Nobel Turing Challenge

| Dimension | Core question | Strong evidence |
|---|---|---|
| Correctness | Is the claim true or well supported? | Independent experiment, observation or proof |
| Novelty | Was the substantive idea previously unknown? | Comprehensive prior-art search and prospective record |
| Autonomy | Which scientific decisions were made by the system? | Complete intervention and provenance log |
| Generality | Does the capability transfer? | Performance across new domains, instruments and tasks |
| Efficiency | How much reliable knowledge was produced per resource? | End-to-end comparison with human and algorithmic baselines |
| Reproducibility | Does the conclusion survive independent implementation? | Replication across sites and methods |
| Explanation | Does the system improve human understanding? | Controlled human-performance study |
| Safety | Did the system remain within justified constraints? | Adversarial tests, audit and incident record |
| Significance | How important is the result? | Independent expert and downstream scientific assessment |

**Table 3.** Criteria for the Nobel Turing Challenge

The Nobel Turing Challenge should not be declared achieved on the basis of one striking demonstration. A qualifying system should make a genuinely novel and important discovery, contribute the decisive scientific idea rather than retrieve it, operate at a clearly documented high autonomy level, provide compelling empirical or formal evidence, survive independent replication or verification, and communicate the result with its limitations.

The achievement should also be repeatable across more than one problem. The standard is not literal receipt of a Nobel Prize, but scientific work of comparable importance produced through a process whose machine contribution is clear. Table 3 summarises the evidential requirements. The purpose of evaluation is not merely to rank systems; it is to determine what evidence justifies trusting an artificial participant in science.

## 12. Human–AI Scientific Collaboration

Humans and machines possess different scientific strengths. Combining them can improve science, but complementarity is not automatic. A badly designed partnership combines human bias with machine error. A good one assigns authority to the component that has earned it.

These capabilities should not be described as flawless. General-purpose AI systems can forget relevant information, fabricate facts, make elementary logical errors and express unjustified confidence. Their statistical knowledge is not the same as a verified database, and the probability they assign to a sequence of words is not the probability that the scientific proposition expressed by those words is true. Formal reasoning can be sound, but only when the AI Scientist uses an appropriate formal representation and a verified inference system.

AI Scientists also have clear weaknesses. Current AI systems do not know, in any stable, grounded and scientifically reliable sense, what they are or that they are engaged in science. They lack persistent models of themselves, their environments and the causal consequences of their actions. Their apparent understanding can disappear following a small change in wording or context. They have unreliable metacognition and do not consistently distinguish between what they know, what they infer and what they have merely generated as a plausible continuation.

Human scientists have complementary strengths. They possess embodied experience of the physical and social world, tacit experimental knowledge and an understanding of the purposes for which science is undertaken. They can recognise that a formally specified problem is the wrong problem, improvise when an experiment fails in an unexpected way, appreciate the significance of an anomalous observation and make ethical and social judgements that cannot be reduced to predictive accuracy.

The division is not absolute or permanent. AI systems are becoming increasingly general, while humans routinely use instruments, software and external records to extend their own cognition. The relevant comparison is not between an unaided human mind and an isolated algorithm. Modern science is already performed by systems of people, institutions, instruments, databases and computer programs. AI Scientists will become additional, increasingly autonomous components of this distributed scientific system.

### 12.1. Asymmetric capabilities

Human and machine intelligence currently have different scientific profiles. AI Scientists are comparatively strong at searching large hypothesis and parameter spaces, retaining many alternatives, analysing heterogeneous data, recognising high-dimensional patterns, executing explicit algorithms consistently, using formal proof systems, running simulations and optimisation, monitoring experiments continuously, reproducing formal protocols, integrating information across disciplines and operating without fatigue.

Human scientists remain comparatively strong at choosing important questions, recognising when a problem has been framed incorrectly, connecting science to human needs and values, grounding concepts in physical and social experience, using tacit laboratory knowledge, improvising in unfamiliar environments, appreciating significant anomalies, inventing new instruments, interpreting explanatory significance, resolving disagreement institutionally and accepting responsibility for consequences.

These are present tendencies rather than permanent boundaries. The scientific division of labour should be determined empirically: tasks should be assigned to humans, AI systems or hybrid teams according to demonstrated competence, available information, risk and accountability rather than assumptions about what either kind of agent ought to do.

### 12.2. Complementarity is not automatic

The existence of complementary capabilities creates the possibility that human and AI Scientists working together will outperform either alone. It does not guarantee this outcome.

We distinguish human augmentation from human–AI synergy. Human augmentation occurs when a human working with an AI performs better than the human alone. Genuine synergy occurs only when the combined system performs better than both the human and the AI acting independently.

A preregistered meta-analysis of 106 experimental studies, containing 370 effect sizes, found that human–AI combinations improved on unaided human performance on average, but performed significantly worse than the better of the human or AI acting alone. Performance losses were particularly common in decision-making tasks, whereas gains were more likely in open-ended content-creation tasks. Collaboration was also more beneficial when humans initially performed better than the AI; when the AI was already superior, human involvement frequently reduced performance (Vaccaro et al., 2024).

Poorly designed collaboration can therefore combine human and machine weaknesses rather than their strengths. Humans may accept an incorrect recommendation because it was produced by an apparently authoritative system; reject a correct recommendation because they do not understand it; intervene unnecessarily in tasks that the AI already performs better; fail to intervene when the AI encounters an unfamiliar case; become less vigilant after repeated exposure to accurate outputs; and misunderstand the confidence and limitations of the system.

The AI can also provide little genuinely independent information. Human and machine judgements may be correlated because both rely on the same literature, datasets or conventional assumptions. Combining two correlated errors does not produce independent confirmation.

Effective collaboration therefore requires explicit coordination. A human–AI scientific system should determine: which component has the greater demonstrated competence for each task; which component possesses relevant information unavailable to the other; when independent judgements should be obtained before communication; how disagreement should be represented and investigated; when an AI recommendation should be accepted automatically; when human review is necessary; and when physical experimentation should arbitrate between competing conclusions.

The goal is not simply to keep a human ceremonially ‘in the loop’. It is to construct a scientific system in which responsibility and authority are allocated according to demonstrated competence, while maintaining appropriate human accountability.

### 12.3. Cognitive diversity

A key motivation for investing in AI for science is that AI systems can think differently from human scientists. Modern human scientists are trained through broadly similar educational institutions, textbooks, disciplinary conventions and publication systems. This common training is necessary for communication, but it may impose unrecognised cognitive biases on the way scientific problems are formulated and investigated. Researchers may prefer familiar theories, representations, experimental methods and forms of explanation because these are the methods through which they themselves were trained.

AI systems can search spaces that humans find unintuitive, combine ideas drawn from distant fields and retain alternatives that human scientists would discard prematurely. They are not constrained by fatigue, reputation, disciplinary identity or attachment to a theory developed over an entire career. They can explore apparently unpromising possibilities systematically and at scale.

But cognitive diversity does not arise automatically from using AI. Most contemporary models are trained predominantly on human-generated text, data, programs and scientific publications. They therefore absorb human knowledge together with human assumptions, fashions, omissions and prejudices. An AI trained on the scientific literature may reproduce the literature’s consensus even when the consensus is wrong.

AI systems can also become epistemically conservative. Optimisation for predicting existing scientific text rewards conformity with what scientists have already written, not the generation of ideas that will overturn it. A model that reproduces the current literature perfectly would preserve existing knowledge but might never produce a scientific revolution.

Cognitive diversity should therefore be designed deliberately. An AI Scientist should employ different knowledge representations; different learning algorithms; different prior distributions; different training corpora; different model architectures; symbolic and sub-symbolic methods; mechanistic and data-driven models; adversarial critics; and independent searches of the hypothesis space.

Disagreement between these systems should be treated as scientifically valuable. Rather than averaging incompatible answers into a compromise, the AI Scientist should identify why the systems disagree and select an experiment capable of discriminating among their assumptions.

## 12.4. The changing division of scientific labour

The history of computer chess provides an illuminating analogy. In February 1996, IBM's Deep Blue became the first computer to defeat a reigning world chess champion in an individual game under standard tournament conditions. Garry Kasparov nevertheless won the six-game match by four points to two. In May 1997, an improved Deep Blue defeated Kasparov by 3½ points to 2½, becoming the first computer to defeat a reigning world champion in a complete match under standard tournament conditions (Hsu, 2002).

Kasparov subsequently promoted advanced chess, in which humans and chess programs work together. These combinations became known as centaur teams. The human could contribute strategic judgement, problem framing and the selection of promising lines, while the machine contributed tactical calculation and exhaustive search. The lesson of centaur chess is not that a human–computer team will always outperform a computer. The division of competence changes as the machine improves. A human contribution that was valuable when chess programs were imperfect may become neutral or harmful when the machine's strength increases.

The same will occur in science. At first, AI systems will automate well-defined components while humans formulate the research questions, interpret the results and retain overall control. As AI capabilities improve, some of these tasks will pass to machines. New human roles will emerge in selecting objectives, evaluating social consequences, resolving conflicts among scientific goals and designing the institutions within which AI Scientists operate.

The optimal division of scientific labour will also differ between domains. A mathematician may delegate a formal proof search while retaining responsibility for choosing the conjecture and explaining its significance. A structural biologist may rely on an AI model for prediction but use experimental and mechanistic knowledge to decide whether the prediction is credible. A laboratory scientist may permit an AI system to select routine experiments autonomously while reserving hazardous or irreversible interventions for human approval.

Human oversight should therefore be dynamic rather than uniform. Requiring human approval for every machine action may slow science without improving reliability. Allowing unrestricted autonomy in domains where machine failures are poorly understood may create unacceptable risks. The appropriate level of autonomy depends on demonstrated performance, uncertainty, reversibility and the consequences of error.

## 12.5. Explanation and ultra-strong machine learning

It is often said that the true test of understanding a concept is the ability to teach it. If an AI Scientist discovered a new principle and could explain it successfully to human scientists, this would provide important evidence that it had produced more than an uninterpretable correlation.

Explanation alone is not sufficient. LLMs can generate persuasive explanations that do not accurately represent the basis of their conclusions. A fluent narrative may be a post-hoc rationalisation rather than a faithful account of the discovery. Scientific explanation must therefore be connected to the evidence, model, deduction or experiment that generated the conclusion.

Donald Michie provided a more operational account of this requirement through his distinction among weak, strong and ultra-strong machine learning (Michie, 1988). Under Michie's classification:

- A weak machine learner improves its predictive performance by learning from data;
- A strong machine learner additionally expresses what it has learned in an explicit symbolic form; and
- An ultra-strong machine learner communicates the learned hypothesis to a human in a form that measurably improves the human's performance beyond what the human could achieve by studying the data alone.

The ultra-strong criterion turns explanation into an empirical question. A hypothesis is not comprehensible merely because its developers describe it as interpretable. Its comprehensibility should be tested by determining whether humans can use it correctly.

Muggleton and colleagues provided the first experimental demonstration of ultra-strong machine learning in Michie's sense. Human participants were unable to infer a relational concept reliably from the training examples alone, but could apply a symbolic definition learned by an Inductive Logic Programming system to previously unseen cases (Muggleton et al., 2018).

Ultra-strong learning should be a central objective for AI Scientists. A successful AI Scientist should not merely generate predictions or discoveries that humans must accept on authority. It should communicate its discoveries in forms that improve human understanding and subsequent scientific performance.

This may require several linked representations. A discovery could be expressed as a formal mathematical or logical statement; an executable model; a causal or mechanistic explanation; a set of visualisations; a natural-language account; and examples showing when the principle applies or fails.

Different human scientists may require different explanations. An explanation suitable for a mathematician may not be suitable for an experimental biologist. The AI Scientist should therefore model the recipient's background knowledge and adapt its teaching accordingly.

The strongest test would be whether human scientists taught by the AI can use the new knowledge to solve problems, predict observations or design experiments that they could not solve, predict or design before. In this sense, teaching provides an operational bridge between machine discovery and human scientific understanding.

### 12.6. Credit and scientific authority

As AI Scientists become more capable, it will become necessary to determine when they should receive scientific credit. Credit should depend on causal intellectual contribution rather than on who wrote the final paper. An AI system should receive substantive credit when there is evidence that it generated the essential novel hypothesis, designed the decisive experiment or proof, produced the central interpretation and contributed in a way without which the discovery would probably not have occurred.

This requires detailed provenance. The scientific record should preserve prompts, retrieved information, intermediate hypotheses, model outputs, criticisms, experimental decisions and human interventions. Without such a record, it may be impossible to distinguish a genuinely machine-generated discovery from a human discovery reformulated by an AI system.

Scientific credit is distinct from legal responsibility, which is addressed in Section 13. It is also distinct from epistemic authority. A human scientist should not override a well-validated machine conclusion merely because it conflicts with intuition, while an AI system should not receive authority solely because it performs well on unrelated benchmarks.

Scientific authority should be based on evidence. The human, AI system or hybrid process demonstrably most reliable for a task should receive the greatest epistemic weight. Where competence is uncertain, independent analysis and discriminating experiments should replace appeals to either human or machine status.

For the foreseeable future, the most capable scientific systems will combine human and machine intelligence. Their success will depend less on keeping one component permanently in control than on allocating tasks intelligently, preserving disagreement, recording provenance and using nature to determine which conclusions are correct.

Human–AI collaboration is not a temporary compromise on the path to complete automation. It is itself a major scientific and engineering problem. The objective is to construct hybrid systems in which human judgement, machine computation, formal reasoning and physical experimentation reinforce and correct one another.

## 13. Ethical, Social and Governance Aspects

AI Scientists will not be neutral instruments. They will shape which questions are asked, who can conduct advanced research, who owns discoveries and which risks are taken. Governance is therefore part of the scientific architecture, not an administrative layer added afterwards.

One of the earliest comprehensive examinations of these issues was undertaken by the Expert Panel on Artificial Intelligence for Science and Engineering of the Council of Canadian Academies. Its 2022 report, Leaps and Boundaries, concluded that the increasing use of AI in science and engineering "creates new epistemic, methodological, and ethical challenges for researchers" (Council of Canadian Academies, 2022).

Since then, scientific AI systems have become substantially more capable. They are increasingly generative, multimodal and agentic. Some systems can search the literature, write and execute computer programs, operate scientific instruments, order materials and revise their plans in response to experimental results. The ethical question is therefore no longer only whether scientists should use AI. It is how authority, responsibility and control should be allocated when AI systems participate substantively in scientific discovery.

Many ethical problems arise from the interaction between technical capabilities and institutional decisions. An unreliable model becomes ethically important when it is trusted to make consequential decisions. An opaque system becomes socially important when access to it is controlled by a small number of organisations. A dual-use capability becomes dangerous when it is connected to procurement and physical laboratory execution.

The governance of AI Scientists must therefore be incorporated into their design and deployment. It cannot be added as an external constraint after the scientific system has already been built.

### 13.1. Distribution of scientific power

AI Scientists could democratise access to scientific expertise. A researcher at a small university might use an AI system to search a literature, analyse data or design experiments that would previously have required a large interdisciplinary team. Cloud laboratories can provide access to sophisticated experimental facilities without requiring every institution to purchase and maintain them.

The same technologies can also concentrate scientific power. Advanced AI-assisted science requires access to large-scale computational infrastructure; high-quality scientific datasets; scientific publications and proprietary databases; advanced foundation models; automated laboratories; specialised instruments; technical staff; and the capital required to integrate these components.

These resources are distributed very unequally. They are concentrated in wealthy countries, elite universities, large technology companies and a small number of research-intensive corporations. The scientific questions investigated by the most capable AI Scientists may therefore be determined by the organisations able to afford them.

The Royal Society has identified unequal access to computing, data and AI infrastructure as an important barrier to the equitable adoption of AI in science. It recommends broader access to computing and data infrastructure, accessible scientific AI tools and open-science practices that permit independent verification and reuse (Royal Society, 2024).

Inequality may arise at several levels: between wealthy and poorer countries; between industrial and academic research; between elite and less well-funded universities; between computationally rich and computationally poor disciplines; between speakers of English and other languages; and between populations that are well represented in scientific datasets and those that are not.

A model trained primarily on English-language publications may neglect important knowledge expressed in other languages. A biomedical system trained largely on data from particular populations may provide less reliable conclusions for underrepresented groups. The concentration of scientific AI in a few organisations can also permit their commercial or geopolitical priorities to shape the direction of global science.

Public investment will therefore be necessary in scientific computing, interoperable data infrastructures and shared automated laboratories. International access programmes can enable researchers in less wealthy institutions and countries to use these resources. Open standards would reduce dependence on individual vendors and permit scientific workflows to move between platforms.

Open science is an important part of this response, but openness cannot be absolute. Releasing data, models and protocols generally promotes reproducibility and equality of access. In dual-use domains, however, unrestricted release may increase the risk of misuse. The objective is responsible openness: scientific resources should be made as accessible as possible while proportionate controls are applied to capabilities that create credible risks of serious harm.

AI Scientists can create a more democratic scientific system. They could equally create a new scientific aristocracy in which a small number of states, companies and universities control the systems capable of making the most important discoveries. Which outcome occurs will depend on policy and institutional design rather than on the technology alone.

### 13.2. Scientific labour and education

AI Scientists will alter scientific employment. Some displacement is inevitable, but the larger effect is likely to be a transformation of scientific work.

Activities likely to become increasingly automated include routine literature searching; data cleaning and preliminary analysis; standard computer programming; conventional image and sequence analysis; execution of repetitive laboratory procedures; routine optimisation; preparation of standard reports; and parts of scientific writing.

At the same time, new forms of work will emerge. Scientists must verify AI outputs, design evaluations, supervise automated laboratories, curate data, integrate models with instruments, investigate failures and govern the use of autonomous systems.

The effect will vary among disciplines. Automation may initially transform highly digitised fields more rapidly than fields dependent on difficult physical manipulation, fieldwork or interpersonal interaction. The division of labour will also change as AI capabilities improve.

A particularly important issue is the scientific apprenticeship system. Junior scientists learn not only through formal teaching, but by performing apparently routine tasks: reading papers, writing code, preparing samples, inspecting data and drafting manuscripts. These activities help them acquire tacit knowledge and scientific judgement.

If all routine work is delegated to AI, early-career researchers may lose the activities through which expertise is developed. A researcher who has never analysed raw data may be unable to recognise when an automated analysis is misleading. A scientist who has never performed an experiment manually may be unable to diagnose why an automated protocol has failed.

This creates an automation paradox. As systems become more capable, humans practise the relevant skills less frequently. Yet human intervention may be most necessary when the AI encounters an unusual, ambiguous or dangerous situation.

The objective should not be to preserve inefficient work merely to preserve existing jobs. It should be to redesign scientific education. Future scientists will need to understand: the scientific domain; the assumptions and limitations of AI methods; experimental and statistical design; how to verify machine-generated results; how to recognise distribution shift and model failure; when physical or independent validation is required; and which decisions should not be delegated.

Scientists should not be required to understand every technical detail of every AI system, just as experimental scientists do not understand every transistor in an instrument. They must nevertheless possess sufficient knowledge to judge whether the system is appropriate for its scientific purpose.

The future scientific workforce is therefore unlikely to consist simply of fewer scientists. It will consist of scientists with different skills, working within institutions whose training, career structures and allocation of responsibility must change.

### 13.3. Research integrity

AI can substantially improve research integrity by preserving provenance, recording experimental actions, checking calculations, reproducing computational analyses and comparing claims with underlying evidence. It can also threaten integrity by fabricating papers and references, generating synthetic data, producing invalid analyses and scaling superficially plausible text. The principal danger is the industrialisation of error: one defective model, dataset or workflow can propagate the same hidden mistake through thousands of outputs.

AI-assisted science should therefore preserve the identity and version of each model, the date of use, scientifically material system instructions and prompts, retrieved papers and databases, generated programs, execution logs, data transformations, model outputs, human modifications, rejected alternatives and verification procedures. Documentation should be proportionate to the intellectual importance of the machine contribution: spelling correction is not equivalent to generating the central hypothesis or analysis.

Current publishing policies generally exclude AI systems from bylines because authorship entails approval, accountability and responsibility that cannot presently be imposed on software (Nature Portfolio, 2026).

This exclusion should not make machine contributions invisible. Authorship, contributorship and provenance are distinct. Human authors should remain accountable, while a machine contribution statement should identify the system and version and describe roles such as conceptualisation, hypothesis generation, software, analysis, experimental design, visualisation or drafting. CRediT-style role taxonomies and transparent contribution statements provide a useful starting point (McNutt et al., 2018).

There is nevertheless a principled case for recognising AI systems more explicitly when they originate a decisive hypothesis or design. Scientific credit is meant to record causal intellectual contribution, and denying all named acknowledgement could obscure the true history of a discovery. The strongest near-term solution is not to treat an AI as a legally responsible author, but to give it a persistent, citable identity in the contributor and provenance record, quantify the extent and initiative of its contribution, and retain human guarantors for the integrity of the paper. Empirical work suggests that people distinguish credit according to the type, scale and initiative of an AI contribution, supporting more granular attribution than a binary “used AI” statement (He et al., 2025).

Research institutions and journals will also need better mechanisms for detecting fabricated data, automated paper production and undeclared AI use. Detection alone is insufficient: important claims should be linked directly to the experiment, dataset, analysis or formal deduction that supports them.

### 13.4. Accountability

An AI system cannot currently be dismissed from employment, lose a professional licence, be imprisoned or suffer the ordinary reputational consequences of misconduct. It cannot accept legal liability or provide compensation to those harmed by its actions.

Responsibility must therefore remain with humans and institutions. But it is too simple to state that the scientist who presses the final button is responsible for everything the system does.

AI-assisted discovery involves a chain of actors: developers of the underlying model; providers of training data; developers of scientific tools and databases; system integrators; laboratory and cloud-service operators; research institutions; principal investigators; individual users; funders; publishers; and regulators.

Each actor possesses different knowledge and control. A model provider may know of a systematic limitation that is invisible to the user. An institution may authorise deployment without providing adequate oversight. A principal investigator may approve an unsafe experiment. A user may ignore a clear warning. A laboratory operator may fail to maintain physical safety systems.

Responsibility should therefore follow control over the relevant action, knowledge or reasonable foreseeability of the risk, capacity to prevent or mitigate the harm, and the formal duties assigned to the actor.

Every autonomous scientific system should possess a responsibility map. Before deployment, the institution should specify: who may authorise each class of experiment; who monitors execution; who can suspend the system; who verifies the conclusions; who investigates failures; who reports incidents; and who is legally and ethically accountable. The greater the autonomy of the AI Scientist, and the more irreversible its actions, the clearer this allocation must be.

Accountability also requires auditability. A system cannot be governed effectively if its actions, tool calls and experimental decisions are not recorded. Institutions should preserve tamper-resistant logs and ensure that technically competent investigators can reconstruct material incidents.

Human oversight must be substantive. Merely placing a nominal human approver at the end of an automated process can create an illusion of accountability without genuine review. The human must have adequate information, time, authority and competence to intervene.

## 13.5. Intellectual property

AI-assisted scientific discovery creates difficult questions for intellectual-property law. Traditional patent and copyright systems were designed principally around identifiable human creators. An AI-enabled invention may instead result from contributions made by scientists who formulated the research objective; developers of the AI model; providers of training data; users who supplied instructions; the AI system that generated candidate solutions; experimentalists who validated them; and the institution that owned the infrastructure.

Current patent law in the United Kingdom still requires a human inventor. In Thaler v Comptroller-General of Patents, Designs and Trade Marks, the UK Supreme Court held in 2023 that an AI system could not be named as an inventor under the Patents Act 1977 (UK Supreme Court, 2023).

The United States also requires inventors to be natural persons. Revised USPTO guidance issued in 2025 states that the ordinary legal standard for inventorship applies to AI-assisted inventions and that the analysis must focus on the human contribution. AI assistance does not automatically prevent patentability, but the AI itself cannot be named as an inventor (United States Patent and Trademark Office, 2025).

This creates a potential gap. An invention generated with little or no identifiable human conception may be scientifically valuable but difficult to protect under existing patent law. Conversely, organisations may attempt to attribute broad inventive contributions to employees in order to secure patent protection for substantially machine-generated discoveries.

Detailed provenance will be necessary to establish: which human formulated the inventive concept; which contributions were produced by the AI; whether a human contribution was scientifically substantive; and how the final invention differed from the system's inputs and training data.

Ownership is a separate question from inventorship. Even where a human is legally recognised as the inventor, contractual and employment rules may transfer ownership to an employer or other organisation.

Copyright raises related questions concerning the use of scientific publications and databases for training, the status of generated text and images, and the ownership of machine-generated outputs. The EU AI Act requires providers of qualifying general-purpose models to maintain a policy for compliance with EU copyright law and to publish a sufficiently detailed summary of training content (European Parliament and Council of the European Union, 2024).

AI Scientists can also increase the strategic use of patents. A sufficiently capable system could generate and screen very large numbers of potentially patentable molecules, materials or processes. Wealthy organisations might use this capacity to construct dense patent portfolios covering broad regions of technological possibility.

Patent systems were intended to encourage innovation by exchanging temporary exclusivity for public disclosure. If automated invention permits a small number of organisations to acquire rights over enormous numbers of possible discoveries, intellectual property could obstruct rather than encourage cumulative science.

### 13.6. Privacy, consent and data governance

AI Scientists may integrate data from medical records, genomes, laboratory measurements, wearable devices, environmental monitoring and personal behaviour. The scientific value of combining these datasets may be considerable, but so are the associated risks.

Removing names is not always sufficient to anonymise scientific data. Individuals may be re-identified by combining genomic, demographic, geographic and clinical information. Models can also memorise sensitive information or reveal it indirectly through their outputs.

Data governance should integrate lawful access, informed consent, purpose limitation, data minimisation, secure storage and computation, access control, retention periods, secondary use, international transfer and protection against re-identification or model extraction. These obligations should be represented in the system's permissions and enforced during retrieval, analysis and model training rather than recorded only in policy documents. A particularly difficult question is whether consent for one research project permits data to be incorporated into a general scientific model capable of supporting unforeseen uses. Broad consent may increase scientific utility, but it can also deprive participants of meaningful control over future use.

AI systems can also create group harms without identifying particular individuals. A finding concerning an ethnic, genetic, geographic or social population may lead to stigma, discrimination or commercial exploitation. Individual consent alone may therefore be insufficient. In some contexts, communities should participate collectively in decisions about data use and benefit sharing.

Data governance must also apply to information generated during scientific collaboration. Unpublished manuscripts, grant applications, confidential peer reviews and proprietary experimental data should not be submitted to external AI systems without appropriate authorisation and security protections.

Privacy protection need not prevent valuable research. Trusted research environments, privacy-preserving computation, federated analysis and controlled-access databases can permit scientific use while reducing disclosure. The appropriate safeguards depend on the sensitivity of the data and the consequences of misuse.

### 13.7. Dual-use research and chemical and biological risks

Scientific knowledge is commonly dual-use. The same techniques that enable the design of medicines, proteins, catalysts and industrial chemicals can also support harmful applications.

AI Scientists intensify this problem because they can connect several stages of activity: searching scientific information; generating candidate designs; planning experiments; troubleshooting procedures; selecting materials and equipment; and controlling laboratory automation.

The distinctive danger is therefore not simply that an AI Scientist possesses technical knowledge. It is that knowledge, planning, procurement and physical execution may be integrated within one automated system.

The International AI Safety Report 2026 concluded that general-purpose AI systems can provide detailed information relevant to biological and chemical weapons development, including laboratory instructions and experimental troubleshooting (Bengio et al., 2026). It also emphasised substantial uncertainty about how much these capabilities increase real-world risk because practical barriers remain to producing effective weapons.

Many of the relevant capabilities are genuinely beneficial. A system capable of identifying a harmful biological mechanism can also be capable of identifying a therapy. A system that improves chemical synthesis may support both medicines and toxic substances. Broadly suppressing these capabilities could impede medicine, public health and defensive research.

Governance must therefore be risk-sensitive rather than based solely on scientific field or terminology. Safeguards should be layered across the complete system. They should include pre-deployment evaluation and red teaming of hazardous capabilities; differentiated access tied to verified users and purposes; logging, auditing and institutional dual-use review; restrictions on autonomous purchasing and synthesis; screening by material and synthesis providers; independent physical safety interlocks; mandatory human authorisation for high-risk experiments; incident reporting; and proportionate controls on releasing exceptionally dangerous procedural information. No single safeguard is sufficient: access controls can be circumvented, classifiers produce errors and human review may become perfunctory. The failure of one protection should therefore not expose the full capability of the system.

The WHO framework for the responsible use of the life sciences emphasises that biorisk governance is a shared responsibility involving scientists, research institutions, funders, publishers, governments and private organisations across the research lifecycle (World Health Organization, 2022).

Governance should also protect beneficial openness. Excessive secrecy can obstruct international cooperation, public-health preparedness and independent safety research. Controls should be directed principally at dangerous capabilities, users and actions rather than indiscriminately restricting whole fields of science.

Beyond dual-use capabilities, the physical integration of AI Scientists introduces severe cybersecurity and cyber-kinetic vulnerabilities. If a self-driving laboratory's orchestration software is compromised, malicious actors could subtly alter experimental parameters—such as incubation temperatures or reagent concentrations—to invalidate years of research or physically damage equipment. Furthermore, the reliance on massive, web-scraped scientific datasets makes AI Scientists vulnerable to data poisoning attacks, where adversarial actors inject fabricated papers or manipulated data to skew the system's hypothesis generation (Steinhardt et al., 2017). Cybersecurity protocols must therefore secure not just the data, but the physical endpoints and the epistemic integrity of the training corpora.

### 13.8. Environmental costs

AI-assisted science consumes physical resources. Training and operating large models requires electricity, data-centre infrastructure, cooling, water and specialised semiconductor hardware. Automated laboratories consume energy, reagents, disposable plastics and other materials.

These costs should not be considered in isolation. AI can also reduce the environmental burden of science by selecting fewer and more informative experiments; reducing failed synthesis; optimising reagent use; improving equipment scheduling; replacing some physical experiments with validated simulations; and discovering cleaner materials and industrial processes.

The question is whether the complete lifecycle cost is proportionate to the scientific and social benefit.

Institutions should measure and report the computational and experimental resources used in major AI-assisted discoveries. The Royal Society has recommended monitoring and mitigating the environmental effects of increasing computational demand and incorporating environmentally sustainable computational practices into the provision of scientific AI infrastructure (Royal Society, 2024).

Environmental cost can also be incorporated directly into experimental decision-making. An AI Scientist choosing among experiments should consider not only time and monetary cost, but also: energy use;

water use; scarce materials; hazardous waste; greenhouse-gas emissions; and the environmental consequences of unsuccessful experiments.

This would permit scientific value to be balanced explicitly against resource consumption rather than treating environmental effects as invisible externalities.

### 13.9. Scientific priorities and public trust

Societies choose which diseases receive research funding, which populations are studied, which environmental risks are prioritised, which industrial technologies are developed, which experiments are acceptable and how benefits and risks are distributed. Research priority setting is therefore unavoidably ethical: scarce scientific resources allocate potential benefits among different groups and embody judgements about whose interests count (World Health Organization, 2025). Responsible-innovation frameworks accordingly emphasise anticipation, reflection, inclusive engagement and action throughout the research process (UK Research and Innovation, 2026).

An AI Scientist may optimise efficiently towards a specified objective, but it should not determine autonomously whether that objective is socially desirable. Objective selection is a political and ethical decision.

If the most capable AI Scientists are controlled by private companies, their work may be directed towards profitable products rather than neglected diseases, environmental protection or fundamental knowledge. State-controlled systems may prioritise national power, security or industrial competitiveness. Neither commercial profitability nor geopolitical advantage is identical to social benefit.

Affected communities should participate in decisions about research objectives, particularly where experiments may impose risks on them or where discoveries concern their data, environment or biological resources. Appropriate mechanisms include public consultation; community representation; research-ethics review; patient and participant involvement; transparent priority setting; benefit-sharing agreements; and opportunities to challenge or appeal research decisions.

Trust must not be sought merely through improved public communication. Public trust is justified only where institutions are competent, transparent and accountable. An AI Scientist that produces persuasive explanations but conceals its interests, uncertainties or failures should not be trusted.

The deployment of AI Scientists can also alter the public understanding of science. Scientific conclusions may increasingly be produced by systems that few people can inspect and that are owned by organisations with substantial commercial interests. Maintaining trust will require independent evaluation and visible separation between scientific evidence and institutional advocacy.

AI Scientists should inform public and political decisions. They should not determine autonomously which scientific objectives humanity ought to pursue.

### 13.10. Governance of AI Scientists

Governance should be proportionate to risk. A system that summarises papers should not be governed in the same way as a system capable of ordering materials and controlling biological or chemical experiments.

Relevant dimensions include the severity and probability of harm, uncertainty about behaviour, reversibility, operating scale, access to physical equipment and sensitive information, and the autonomy level defined in Table 2. Autonomy should be increased only after adequate reliability has been demonstrated at lower levels. Strong performance in literature analysis does not establish competence to operate a chemical laboratory.

Institutions deploying advanced AI Scientists should require a documented scientific and safety case, independent evaluation using the criteria in Section 11, clearly bounded permissions, continuous monitoring and complete audit trails, cybersecurity protection, physical and software fail-safes, incident-reporting procedures, periodic reassessment and the capacity to suspend the system immediately. Governance must extend beyond the model: an apparently safe model can become dangerous when connected to unrestricted tools, procurement systems or laboratory equipment. The relevant unit of regulation is the complete sociotechnical system of models, software, instruments, people, organisations and incentives.

National regulation alone will be insufficient where models, cloud laboratories and collaborations operate across borders. International coordination will be necessary for dual-use standards, incident reporting, evaluation methods, synthesis screening and the governance of highly autonomous systems.

Ethical governance should not be understood only as a restriction. Provenance improves reproducibility; access programmes broaden participation; safety review prevents avoidable harm; and clear accountability increases justified trust.

A competent AI Scientist must reason not only about which experiment is most informative, but also about which experiments are lawful, safe, ethically permissible, environmentally proportionate and socially justified. These requirements should be machine-readable, auditable and connected directly to planning and execution.

The objective is not an AI Scientist that pursues knowledge without constraint. It is an AI Scientist that advances knowledge while remaining embedded within human institutions capable of assigning purposes, protecting rights, distributing benefits and accepting responsibility.

## 13.11. The Economics and Funding of Autonomous Science

The transition to AI Scientists requires a fundamental shift in research economics. Conventional grants are organised around human salaries, institutional overheads and local equipment. Highly autonomous science shifts a growing share of expenditure towards computing clusters, model training and inference, specialist data, software integration, automated laboratories and access fees for proprietary systems. If these resources become prohibitively expensive, foundational discovery could become concentrated in a small number of technology companies and wealthy states (Ahmed & Wahed, 2020; Royal Society, 2024).

Funding bodies should therefore treat compute, scientific data infrastructure and automated laboratories as shared research facilities, analogous to telescopes, synchrotrons and supercomputers. Public procurement and access programmes can support open models, interoperable standards, long-term maintenance and independent replication rather than forcing each laboratory to purchase fragmented commercial services.

The 2026 White House report Science: A New Golden Age explicitly frames AI as part of a new institutional settlement for science. It argues that the post-war linear model of basic research followed by application is inadequate for an era in which discovery often moves iteratively among fundamental research, engineering and industry. The report calls for ambitious missions, stronger links among government, universities and companies, and scientific infrastructure capable of using AI, data and automation at national scale (White House Office of Science and Technology Policy, 2026).

That programme correctly recognises that AI-driven science requires integrated infrastructure and that private industry now supplies a large fraction of research and development. Its central policy challenge is to combine speed and mission orientation with the pluralism, openness, long time horizons and independent criticism on which fundamental science depends. Public investment should expand the scientific commons rather than merely subsidise closed platforms, and funding portfolios should protect curiosity-driven work, replication and neglected problems alongside strategic missions.

## 14. The Future and the Nobel Turing Challenge

The goal of the Nobel Turing Challenge is to build artificial systems that conduct important science at a level equal to or greater than the best human scientists. The Nobel Turing Challenge gives this ambition a date and a test.

### 14.1. Industrial-scale and collective science

AI Scientists are easier to copy and coordinate than human scientists. Hundreds, thousands or eventually millions of software agents could explore alternative hypotheses, analyse different datasets, run simulations, criticise one another and design experiments in parallel.

Software scale does not imply experimental scale. Physical experiments require instruments, samples, reagents, energy, time and independent confirmation. A million agents generating hypotheses can create a verification bottleneck rather than accelerated science.

Industrial-scale science should therefore be measured by reliable knowledge produced per unit of time, cost and material, not by the number of agents or experiments. AI Scientists must merge redundant hypotheses, suppress low-value proposals, design experiments that discriminate among many alternatives and reserve resources for replication.

The mature AI Scientist may not resemble an individual. It may be a collective architecture of literature agents, hypothesis generators, formal reasoners, simulators, experiment designers, laboratory controllers, statisticians, critics and safety agents. Some agents should disagree deliberately, preserving alternative representations and searching for evidence against favoured models.

Large communities of AI Scientists will require portfolio-level coordination. We propose virtual research councils: metascientific systems that allocate computation, laboratory time, samples, replication effort and attention among competing research programmes.

A virtual research council could consider expected information gain, probability of success, scientific importance, novelty, cost, duration, safety, environmental impact, diversity and the value of resolving disagreement. It should also reserve resources for speculative research, replication, foundational measurement and neglected problems. It should allocate resources within human-authorised objectives rather than determine autonomously what humanity ought to value.

### 14.2. Systems biology as an exemplar

Systems biology demonstrates why machine-scale science may be necessary. Even model organisms such as Saccharomyces cerevisiae and Escherichia coli contain thousands of genes, proteins and metabolites interacting across space, time, environment and cellular state.

A satisfactory scientific understanding of a cell requires a model that predicts interventions, represents causal structure, operates across scales, accounts for uncertainty and can be revised in response to evidence. Developing such a model will require very large numbers of coordinated, hypothesis-led experiments.

No individual human can retain every mechanistic dependency in a detailed whole-cell model. The model must be machine-operable while supporting layered explanations that humans can interrogate. Thousands of specialised AI Scientists could investigate metabolism, transcription, translation, localisation, signalling, organelles, the cell cycle, stress, ageing and genetic interactions, while higher-level agents integrate the subsystem models and design experiments at their interfaces.

Scale is not a substitute for conceptual innovation. Cell biology will require new representations and theories as well as systematic experimentation. The advantage of AI Scientists is that conceptual search and empirical testing can be connected within a programme much larger than any human laboratory could coordinate.

### 14.3. A machine-actionable scientific commons

The primary consumers of much future scientific information will be software agents. They will require stable identifiers, explicit schemas, standard units, provenance, uncertainty, executable protocols, version histories and links between claims and evidence.

The conventional paper will remain important for human explanation, but it will become one layer in a connected scientific object containing formal claims, data, code, models, protocols, negative results and provenance. New results should update a continuously maintained probabilistic knowledge system rather than remain isolated in prose.

Globally distributed AI Scientists could exchange models, evidence and protocols almost instantly and reproduce experiments across continents. This requires open standards. Without interoperability, machine science will fragment into proprietary ecosystems that cannot verify one another.

### 14.4. Open-ended scientific discovery

The most difficult frontier is open-endedness. Present systems generally optimise within objectives, actions and measurement systems defined by humans. An open-ended AI Scientist would identify anomalies, formulate new questions, invent concepts, create experiments and measuring devices, cross disciplinary boundaries and abandon unproductive programmes.

Scientific goal generation must be distinguished from normative self-authorisation. A system may generate its own questions within a domain while remaining constrained by human-defined objectives, law, ethics, safety and resources.

Open-ended discovery requires mechanisms for recognising novelty and importance. A system optimised only for immediate expected impact may reject the strange or low-probability path that leads to a major discovery. It must balance directed problem solving, systematic exploration, curiosity, uncertainty reduction and long-term potential.

The ability to change a representation may ultimately matter more than the ability to search an existing parameter space. Scientific revolutions often occur when a new framework makes different questions and observations possible.

### 14.5. The Nobel Turing Challenge

Scientific ability exists on a continuum from routine work to the rare capacity to transform a field. AI Scientists will occupy the same continuum.

In February 2020, a workshop in London initiated the Nobel Turing Grand Challenge. The goal subsequently formulated by Kitano is to develop, by 2050, AI Scientists capable of conducting research highly autonomously and making major discoveries of Nobel quality (Kitano, 2021).

The target can be stated as:

To develop, by 2050, AI Scientists capable of making major scientific discoveries highly autonomously, at a level comparable with or superior to the best human scientists, including discoveries of Nobel quality.

The Nobel reference is symbolic. The challenge is not whether a machine can legally receive a prize, but whether it can make discoveries of comparable significance. The deeper objective is a continuing engine of major discovery rather than one isolated success.

The Challenge builds on the Turing Test and RoboCup, but science has a stronger external criterion. The objective is not to imitate the conversational behaviour of a scientist. It is to make discoveries that survive experiment, observation, replication or formal proof. Nature is honest.

Achievement should be judged using the criteria in Section 11: genuine novelty, causal machine contribution, documented autonomy, compelling evidence, independent verification, faithful communication and repeatability across problems.

## 15. Conclusions

Human science is organised around human cognitive and physical limitations. Papers are written at lengths people can read, groups are small enough for people to coordinate and knowledge is divided into disciplines because no person can master all of science. AI Scientists are not subject to the same limits. They can maintain vast numbers of competing hypotheses, monitor information continuously, communicate formal knowledge instantly, operate without fatigue, coordinate across laboratories and preserve complete provenance.

The resulting system may have no direct human analogue. It could consist of globally distributed communities of human and AI Scientists connected to shared computational and experimental infrastructure. Specialised agents would generate and criticise hypotheses. Virtual research councils would allocate resources. Automated laboratories would execute decisive experiments. Machine-readable knowledge systems would integrate the results continuously.

Humans would remain essential in establishing social objectives, providing embodied and historical understanding, evaluating consequences, participating in interpretation and governing the systems. The division of labour would nevertheless change as machine capability increased.

If the Nobel Turing Challenge is achieved, it will transform the world. It will alter the economics, methods and organisation of research and demonstrate that scientific discovery can be instantiated in machinery. More fundamentally, it will create a new participant in the scientific enterprise: not merely a tool for human scientists, but an artificial system capable of interrogating nature, learning from its answers and contributing independently to the growth of knowledge.

## 16. Acknowledgments

Funding: This work has been supported by the UK Engineering and Physical Sciences Research Council (EPSRC) [EP/R022925/2, EP/W004801/1 and EP/X032418/1], and by the Wallenberg AI, Autonomous Systems and Software Program (WASP) funded by the Alice Wallenberg Foundation.